\newif\ifanonymous \anonymousfalse
\newif\ifarxiv \arxivtrue
\newif\ifappendix \appendixtrue

\ifarxiv
    \documentclass[acmtog,nonacm]{acmart}
    \setcopyright{none}
\else
    \ifanonymous
    \documentclass[acmtog, review, anonymous, nonacm]{acmart}   
    \else
    \documentclass[acmtog]{acmart}
    \fi
    \acmSubmissionID{381}
\fi

\AtBeginDocument{%
  }

\ifarxiv
  \setcopyright{none}
  \acmDOI{XXXXXXX.XXXXXXX}
\else
  \ifanonymous
    \setcopyright{none}
    \acmDOI{XXXXXXX.XXXXXXX}
  \else
    \copyrightyear{2026}
    \acmYear{2026}
    \setcopyright{cc}
    \setcctype{by}
    \acmConference[SIGGRAPH Conference Papers '26]{Special Interest Group on Computer Graphics and Interactive Techniques Conference Conference Papers}{July 19--23, 2026}{Los Angeles, CA, USA}
    \acmBooktitle{Special Interest Group on Computer Graphics and Interactive Techniques Conference Conference Papers (SIGGRAPH Conference Papers '26), July 19--23, 2026, Los Angeles, CA, USA}
    \acmDOI{10.1145/3799902.3811073}
    \acmISBN{979-8-4007-2554-8/2026/07}
  \fi
\fi

\usepackage{cleveref}
\usepackage{multirow}

\usepackage{amsmath}
\usepackage{comment}
\usepackage{multirow}
\usepackage{array}
\usepackage{graphicx}
\usepackage{booktabs}
\usepackage{makecell}
\usepackage[normalem]{ulem}
\usepackage{xcolor}
\usepackage{soul}
\usepackage{xspace}
\usepackage{cleveref}
\usepackage{amsfonts}
\usepackage{listings}
\usepackage{placeins}
\lstdefinestyle{promptstyle}{
  basicstyle=\ttfamily\footnotesize,
  breaklines=true,
  breakatwhitespace=true,
  frame=single,
  columns=fullflexible,
  keepspaces=true,
  showstringspaces=false
}

\makeatletter
\newcommand{\settitle}{\@maketitle}
\makeatother

\newcolumntype{C}[1]{>{\centering\let\newline\\\arraybackslash\hspace{0pt}}m{#1}}

\newif\ifdraft
\drafttrue
\ifdraft
\definecolor{darkpink}{rgb}{0.561, 0.282, 0.427}

\newcommand{\dcc}[1]{{\color{red}[\textbf{Danny} #1]}}

\newcommand{\drop}[1]{}

\else
\newcommand{\dcc}[1]{}
\newcommand{\rgc}[1]{}
\newcommand{\opc}[1]{}
\newcommand{\gcc}[1]{}
\newcommand{\hmc}[1]{}
\newcommand{\abc}[1]{}

\fi

\makeatletter
\DeclareRobustCommand\onedot{\futurelet\@let@token\@onedot}
\def\@onedot{\ifx\@let@token.\else.\null\fi\xspace}

\makeatother

\usepackage[bottom]{footmisc}
\usepackage{arydshln}

\makeatletter
\def\blfootnote{\xdef\@thefnmark{}\@footnotetext}
\makeatother

\newcommand{\algoname}{ShapeUP\xspace} 
\ifarxiv
  \newcommand{\supp}[1]{\Cref{#1} of the appendix\xspace}
\else
\newcommand{\supp}[1]{the sup. mat\xspace} 
\fi

\begin{document}

\title{ShapeUP: Scalable Image-Conditioned 3D Editing}

\author{Inbar Gat}
\affiliation{%
  \institution{Aigency.ai}
  \country{USA}
}
\affiliation{%
  \institution{Tel Aviv University}
  \country{Israel}
}
\email{gatinbar2344@gmail.com}

\author{Dana Cohen Bar}
\affiliation{%
  \institution{Tel Aviv University}
  \country{Israel}
}

\author{Guy Levy}
\affiliation{%
  \institution{Tel Aviv University}
  \country{Israel}
}

\author{Elad Richardson}
\affiliation{%
  \institution{Runway}
  \country{USA}
}

\author{Daniel Cohen-Or}
\affiliation{%
  \institution{Tel Aviv University}
  \country{Israel}
}

\renewcommand\shortauthors{Gat, I. et al.}


\ifarxiv
\else
  \ifanonymous
  \else
  \fi
\fi

\begin{teaserfigure}
  \includegraphics[width=\textwidth, trim={0 0 0 80}, clip]{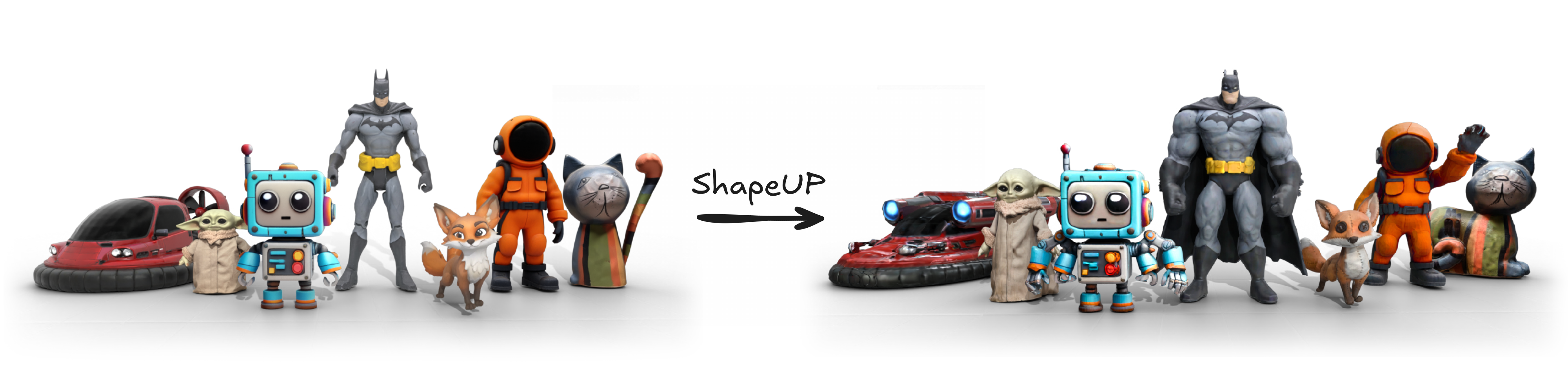}
  \caption{ShapeUP enables diverse 3D edits across a wide range of object categories, supporting both global transformations and localized modifications, including part addition, articulation, and shape deformation.}
  \Description{Description of the teaser figure.}
  \label{fig:teaser}
\end{teaserfigure}

\begin{abstract}
Recent advancements in 3D foundation models have enabled the generation of high-fidelity assets,
yet precise 3D manipulation remains a significant challenge. Existing 3D editing frameworks often face a
difficult trade-off between visual controllability, geometric consistency, and scalability.
Specifically, optimization-based methods are prohibitively slow, multi-view 2D propagation techniques suffer from visual drift, and training-free latent manipulation methods are inherently bound by frozen priors and cannot directly benefit from scaling.
In this work, we present \textbf{ShapeUP}, a scalable, image-conditioned 3D editing framework that
formulates editing as a supervised latent-to-latent translation within a native 3D representation. This formulation allows ShapeUP to build on a pretrained 3D foundation model, leveraging its strong generative prior while adapting it to editing through supervised training. 
In practice, ShapeUP is trained on triplets consisting of a source 3D shape, an edited 2D image, and the corresponding edited 3D shape, and learns a direct mapping using a 3D Diffusion Transformer (DiT).
This image-as-prompt approach enables fine-grained visual control over both local and global edits and achieves implicit, mask-free
localization, while maintaining strict structural consistency with the
original asset. Our extensive evaluations demonstrate that ShapeUP consistently outperforms
current trained and training-free baselines in both identity preservation and edit fidelity, offering a robust and scalable paradigm for native 3D content creation.
\ifanonymous
\else
Our project page, including video results, code, and benchmark, is available at \url{https://inbar-2344.github.io/ShapeUp-page/}.
\fi
\end{abstract}

\maketitle

\section{Introduction}
\label{sec:intro}
In recent years, the field of 3D computer graphics has undergone a paradigm shift, driven by the
emergence of powerful foundation models. Inherent 3D architectures such as TRELLIS
\cite{xiang2024trellis}, Hunyuan3D 2.0 \cite{zhao2025hunyuan3d20scalingdiffusion} and Step1X-3D
\cite{li2025step1x} have demonstrated an unprecedented ability to synthesize high-fidelity geometry
and intricate textures from sparse inputs. These foundational models for 3D content represent a
significant leap toward democratizing 3D asset creation. However, as the focus of the community
matures from generation to practical content creation, the challenge has shifted toward
\textit{3D editing} and manipulating existing assets while preserving their underlying 3D identity.

Despite the strength of current generative backbones, effective 3D editing remains a formidable
challenge plagued by several trade-offs. Ideally, a 3D editing framework should satisfy four key
criteria: 
(i)~\textbf{Inherent 3D consistency},
ensuring that edits are holistically integrated into the 3D representation, preserving geometric coherence and avoiding view-dependent artifacts;
(ii)~\textbf{Implicit localization}, inferring edit regions directly from the conditioning signal, unifying localized refinements and global transformations, without relying on explicit spatial masks that require manual annotation;
(iii)~\textbf{Fine-grained control}, enabling detailed control over edits through visual conditioning, beyond the ambiguity of natural language;
(iv)~\textbf{Inherent scalability}, ensuring that the editing framework benefits from increased data and model capacity, in line with ``The Bitter Lesson'' \cite{sutton2019bitter}.
Current state-of-the-art methods
typically sacrifice one or more of these properties. For instance, text-driven models like
Steer3D \cite{ma2025steer3d} lack visual precision, while training-free methods like EditP23
\cite{baron2025editp23} and Nano3D \cite{nano3d_2025} are restricted by the static priors of
frozen weights, preventing them from benefiting from the scaling laws that defined the success
of large-scale 2D models.

In this work, we present \algoname, a scalable, image-conditioned 3D editing framework designed
to satisfy all the aforementioned requirements.
Rather than formulating 3D editing as an optimization process or a multi-view propagation task,
we cast it as a supervised latent-to-latent translation within a native 3D representation. Our edits are specified via an image showing a single view of the edited target. This image-as-prompt formulation provides a unified interface that naturally supports a wide variety of edits, where the model learns to propagate the visual instruction across the 3D shape while preserving its underlying structure.

To train our model we design a synthetic dataset that spans both global edits, such as pose changes and shape deformations, and fine-grained local edits. First, we create part-based edits that simulate component addition and removal, capturing localized structural changes. To extend beyond these local edits, we introduce edit pairs originating from temporally distant frames in animation sequences. We observe that such pairs naturally capture coherent pose and deformation changes while preserving object identity, making them well suited for supervising global 3D edits. We refer to these pairs as Distant Frames in Motion (DFM) and show that they play a critical role in enabling identity-preserving global edits and improving overall generalization.

Current 3D foundation models typically decompose the generation process into two stages, first generating the geometry and then applying texture to the resulting shape. We adopt the same decomposition and fine-tune each stage to support shape and texture editing, respectively.
For geometry, we build on a pretrained 3D Diffusion Transformer (DiT) and
fine-tune it with a LoRA~\cite{hu2022lora} to accommodate both the original 3D shape and the editing image. By encoding the original asset into the structured
latent space of a native-3D foundation model and conditioning the diffusion
process on both the edited image and the encoded source shape, \algoname
learns to ``shape up'' the existing geometry into the target state.
For texture, we adapt a pretrained image-to-multiview model to
synthesize appearance-consistent renderings of the edited shape, explicitly
conditioning on the source texture to preserve fine-grained details that
may be ambiguous in a single input view.
In both cases, the core method remains backbone-agnostic and relies on lightweight adaptation of existing foundation models to achieve strong condition alignment and identity preservation across both local and global edits, see~\Cref{fig:teaser}.

Extensive qualitative and quantitative evaluations show that \algoname consistently outperforms both trained and training-free baselines in identity preservation and edit fidelity. These results highlight the effectiveness of formulating 3D editing as supervised translation within a native 3D latent space, enabling high-bandwidth visual control via image prompts while preserving the identity of the original shape. Additionally, we show that by operating directly on native 3D representations, \algoname naturally supports mask-free localization and global edits, such as pose changes and shape deformations, and avoids the view inconsistency and registration artifacts common in multi-view propagation approaches. Together, these capabilities establish \algoname as a robust and scalable framework for 3D asset editing.
\begin{figure*}

    \centering
    
    \includegraphics[width=\textwidth]{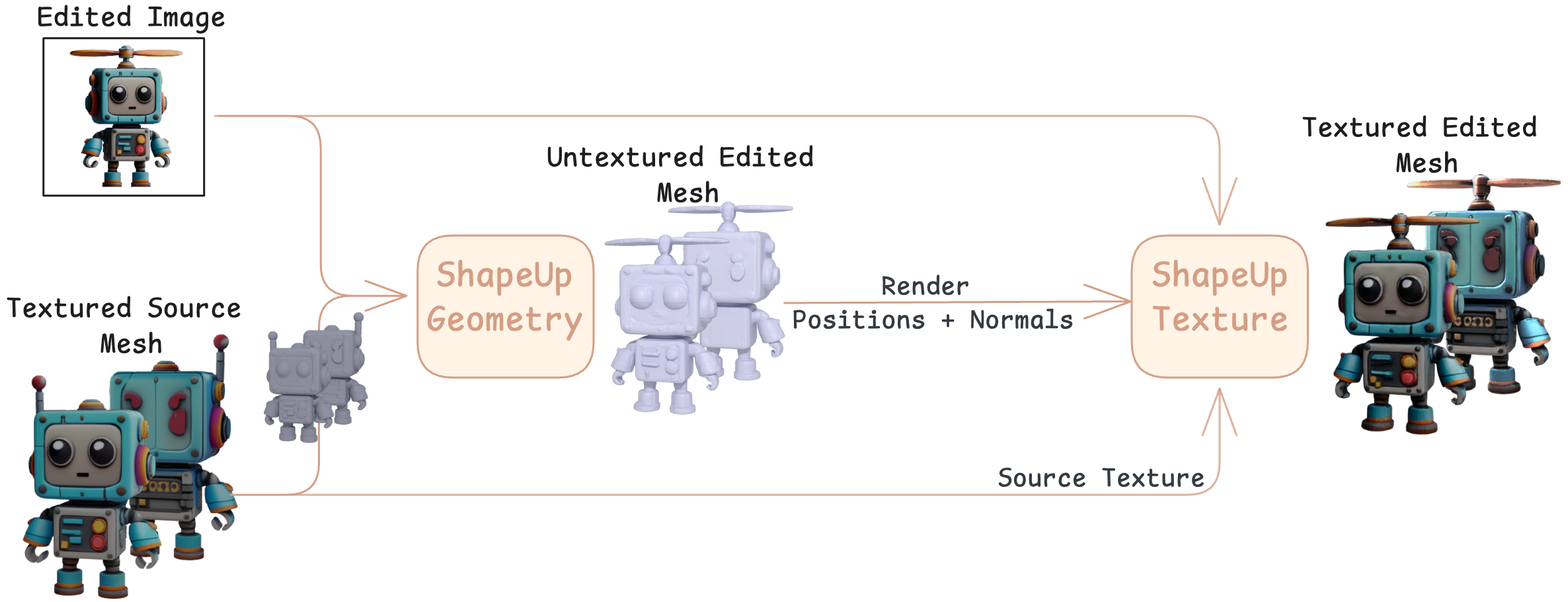}
    
    \caption{
        \textbf{Overview.}
ShapeUP takes a Textured Source Mesh together with a single Edited Image (left). The ShapeUP Geometry module produces an Untextured Edited Mesh by editing the source shape directly in a native 3D latent space, preserving identity and enabling implicit localization. The edited geometry is rendered to obtain Positions + Normals, which guide the ShapeUP Texture module (right) to generate the final Textured Edited Mesh while retaining details from the Source Texture.
    }
    \label{fig:arch}
    \Description[]{}  %
\end{figure*}

\section{Related Work}
\label{sec:related}

\subsection{3D Diffusion Models}
Early 3D generation methods leveraged the abundance of 2D image data by adopting
a two-stage approach: first synthesizing multiple consistent views of an object
\cite{shi2023zero123plus, shi2023mvdream, liu2023syncdreamer},
then lifting these to 3D via reconstruction
\cite{xu2024instantmesh, tang2024lgm, wang2024crm, long2024wonder3d}.
However, inconsistencies across synthesized views can degrade the final output.
More recently, native 3D generative models have emerged that operate directly
in 3D latent space
\cite{xiang2024trellis, zhang2024clay, li2024craftsman, wu2024direct3d, zhao2024michelangelo, li2025triposg}.
These models train variational autoencoders paired with diffusion transformers
for end-to-end 3D generation.
By bypassing the multi-view bottleneck, they achieve higher fidelity and consistency,
providing the foundation for native 3D editing.
Building on these advances, Spice-E \cite{sella2024spicee} enables controllable
generation via cross-entity attention, and Sharp-It \cite{edelstein2025sharpit} refines the original shape multi-view using text prompts to produce 3D-aware edits.

\subsection{3D Editing via 2D Lifting}
Many existing 3D editing approaches perform edits in 2D and subsequently reconstruct the edited object in 3D.
\paragraph{Optimization-based methods.}
Score Distillation Sampling (SDS) lifts 2D diffusion priors into 3D by
optimizing a representation to match text instructions.
Methods such as Instruct-NeRF2NeRF \cite{haque2023instructnerf2nerf},
DreamEditor \cite{zhuang2023dreameditor}, Vox-E \cite{sella2023voxe},
TIP-Editor \cite{zhuang2024tipeditor}, QNeRF \cite{qnerf2024},
and GaussianEditor \cite{chen2024gaussianeditor} follow this paradigm.
While these achieve high-quality results, they require minutes to hours per edit
and are prone to the Janus problem or over-saturation.
\paragraph{Multi-view propagation methods.}
To enable faster editing, several works propagate 2D edits across multiple views.
EditP23 \cite{baron2025editp23} and CMD \cite{cmd2025} leverage
image-conditioned diffusion to propagate edits across views.
MVEdit \cite{chen2024mvedit} introduces a training-free multi-view adapter,
and Instant3Dit \cite{barda2024instant3dit}, PrEditor3D \cite{erkoc2024preditor3d},
NeRFiller \cite{weber2024nerfiller}, and Sked \cite{mikaeili2023sked}
perform inpainting or sketch-guided editing within spatial masks.
Other works in this space include Tailor3D \cite{qi2024tailor3d},
MVInpainter \cite{cao2024mvinpainter}, and DGE \cite{chen2024dge}.
Although faster than optimization, these methods translate 3D to 2D and back,
often introducing registration artifacts or identity drift during reconstruction.

\subsection{Native 3D Editing}
With the rise of native 3D generative models, researchers have begun exploring
editing directly in 3D latent space, bypassing the 2D lifting bottleneck.
\paragraph{Training-free methods.}
VoxHammer \cite{voxhammer2025} employs inversion and contextual feature replacement
to preserve the original shape and apply masks to preserve unedited regions. Despite their strong performance, masking unedited regions can limit expressiveness and suppress natural interactions.  
Nano3D \cite{nano3d_2025} introduces mask-free voxel-slat merging for efficient editing.
While these methods provide 3D-consistent results,
they are fundamentally limited by the fixed priors of their frozen weights.
\paragraph{Learned methods.} InstructPix2Pix \cite{brooks2023instructpix2pix} showed that supervised training on edited triplets enables robust, scalable image editing. In the 3D domain, 3DEditFormer~\cite{xia20253deditformer} fine-tunes TRELLIS to be conditioned on deep-features extracted from the source mesh via a frozen TRELLIS model. In contrast, we condition the model directly on a compact shape latent code, providing a structured, geometry-faithful signal, enabling more accurate and controllable edits.

\textbf{ShapeUp} combines the architectural strengths
of a native 3D diffusion transformer with image-conditioning, enabling high-bandwidth visual control. Our learned approach scales with data, and operates directly in the native 3D latent space, avoiding reconstruction drift from 2D lifting.

\section{Method}\label{sec:method}
\algoname follows the common two-stage structure of modern 3D foundation models, decomposing editing into geometry and texture stages. Given a source 3D asset and an image prompt specifying the desired edit, the geometry stage first modifies the underlying shape, after which the texture stage synthesizes appearance consistent with the edited geometry.

Our implementation builds on the Step1X-3D~\cite{li2025step1x} image-to-3D foundation model, using its pretrained geometry and texture backbones as the basis for image-conditioned 3D editing. In the following sections, we describe the geometry and texture editing stages in detail.

\subsection{Geometry Editing}
The geometry editing stage modifies the underlying 3D shape to match the image-specified edit while preserving the identity and structural consistency of the source asset.

\begin{figure}
    \centering
    \includegraphics[width=\columnwidth]{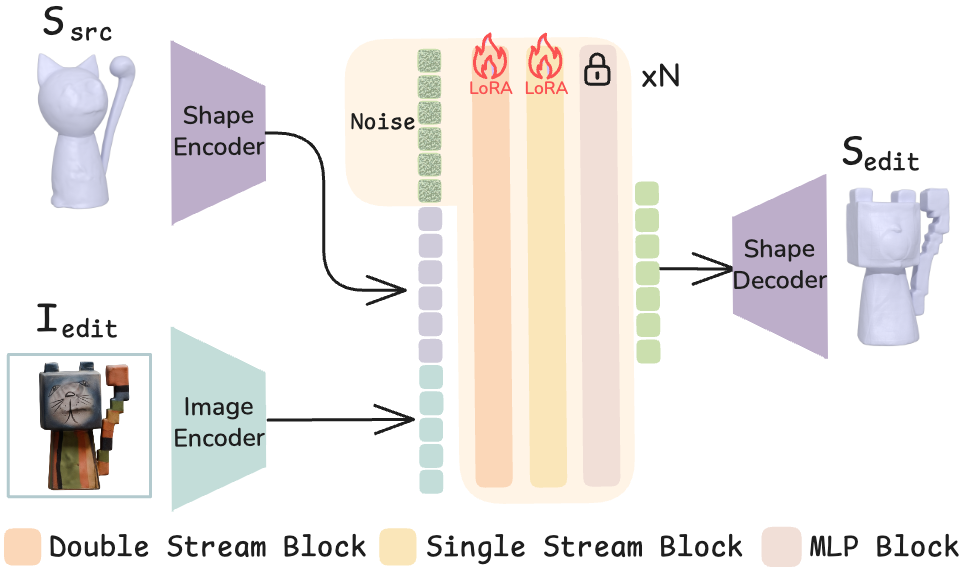}
    
    \caption{
        \textbf{\algoname Geometry Editing.} During inference, the source shape $S_{src}$ is encoded by the Shape Encoder, from which $K$ latent vectors are sampled to form the \it{source geometry conditioning signal}. The edited image $I_{edit}$ is encoded by the Image Encoder to provide the target edit conditioning signal. Both signals are concatenated and processed by $N$ identical layers comprising Double Stream, Single Stream, and MLP blocks, with $LoRA$ trained on the Double and Single Stream blocks. The geometry pipeline produces a latent representation of the edited mesh, which is decoded into 3D by the Shape Decoder to obtain the target shape $S_{edit}$.
       }
    \label{fig:geometry_cond}
    \Description[]{}  %
\end{figure}

\paragraph{Geometry Generation Backbone}
The Step1X-3D geometry generation pipeline follows the common paradigm of first compressing 3D
shapes into a compact latent representation using a shape VAE
\cite{xiang2024trellis, zhao2025hunyuan3d20scalingdiffusion}, and then training a DiT-style
transformer to perform image-to-3D lifting within that latent space. The diffusion backbone adopts
a FLUX-inspired MMDiT design \cite{flux2024}, alternating between double-stream blocks, in which
latent tokens and conditioning tokens are processed in separate branches with cross-attention
interactions, and single-stream blocks that merge both token types and update them jointly. See \supp{appendix:method} for more details.

\paragraph{Geometry Editing Pipeline}
To incorporate the source shape as a conditioning signal, we first encode the source shape,
$S_{src}$, into the same latent space the DiT backbone was trained on, using the pretrained shape VAE encoder, which maps each shape to 2048 latent vectors. Our experiments show that this
representation is highly redundant, and we therefore subsample the encoded representation to
$K = 1024$ latent tokens, which serve as the source shape condition.

Geometry editing is supervised using paired source and target shapes, $(S_{src}, S_{edit})$,
together with an image prompt, $I_{edit}$, depicting a single view of the edited target shape. The
model is trained to translate the latent representation of $S_{src}$ into that of $S_{edit}$
while conditioning on $I_{edit}$.
To inject the source shape condition into the diffusion model, we repurpose the LoRA-based
conditioning mechanism introduced in Step1X-3D for 3D shape conditioning. Specifically, the
subsampled shape latent tokens are concatenated with the image-derived representation of
$I_{edit}$ produced by the pretrained backbone's image encoder, and LoRA adapters are trained on both the double-stream
and single-stream blocks of the MMDiT backbone. This conditioning scheme enables the model to
jointly reason about the original geometry and the target edit within a unified latent space.
The incorporation of the source shape condition is illustrated in
\cref{fig:geometry_cond}.

\subsection{Texture Editing}
The texture editing stage synthesizes appearance consistent with the edited geometry while
preserving fine-grained texture details from the source asset.
\begin{figure}
    \centering
    \includegraphics[width=\columnwidth]{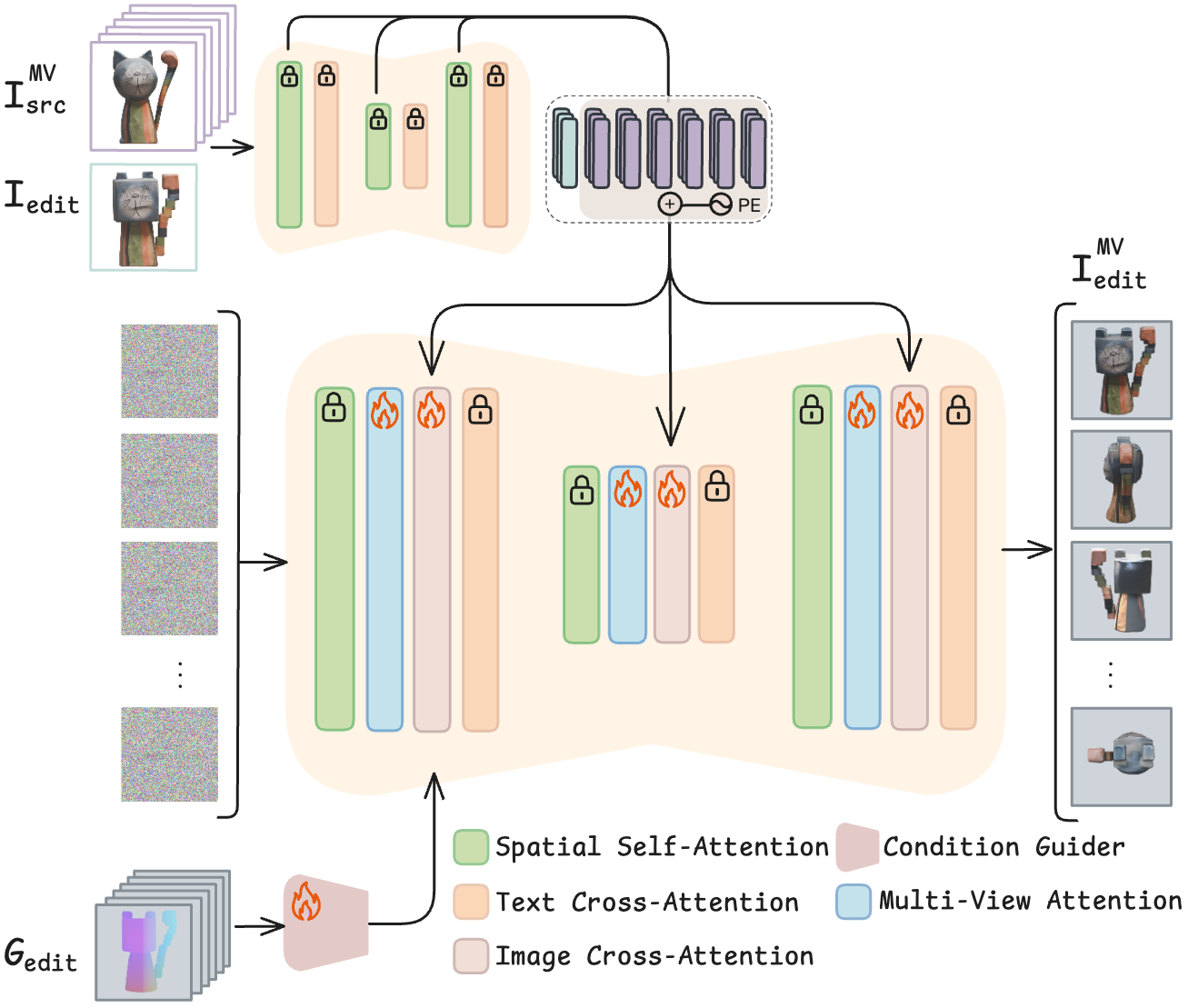}
    
    \caption{
        \textbf{\algoname Texture Editing.}
Texture editing is conditioned on the edited image, $I_{edit}$, multi-view renders of the source textured mesh , $i^{MV}_{src}$, and multi-view normal and position renders of the edited shape, $G_{edit}$. Deep features are extracted from both $I_{edit}$ and $i^{MV}_{src}$ and fused through cross-attention, while $G_{edit}$ features are incorporated as additive-residuals. The model outputs a set of consistent edited multi-view images, which are subsequently baked onto the edited geometry.
       }
    \label{fig:texture_cond}
    \Description[]{}  %
\end{figure}

\paragraph{Texture Generation Backbone}
The Step1X-3D texture generation pipeline builds textures through a geometry-guided
single-view to multi-view diffusion stage followed by texture baking. The diffusion
model takes as input the reference image together with multi-view geometric cues, including
normal maps and position maps, rendered from the untextured mesh. The backbone follows the
MV-Adapter design~\cite{huang2024mvadaptermultiviewconsistentimage}. Geometric conditions
(normals and positions) are encoded and injected as residual signals after the adapter layers,
while image consistency with the reference view is enforced by feeding image-derived condition
tokens through cross-attention within the diffusion network.

\paragraph{Texture Editing Pipeline}
To incorporate the original texture as a conditioning signal, we inject multi-view renders of
the source shape, $I^{MV}_{src}$, through the model’s existing image cross-attention mechanism.
We extract conditioning tokens from each source view using the same feature extraction procedure
applied to the target image $I_{edit}$.

In the cross-attention layers, each generated view attends to both the editing cues derived
from $I_{edit}$ and the corresponding source-view tokens from $I^{MV}_{src}$. To enable the
model to distinguish between target edit cues and view-aligned source texture features, we add a
view-axis positional encoding to the tokens originating from $I^{MV}_{src}$. We fine-tune the
adapter layers to support this additional conditioning signal. This design exposes the generator
to both the edited appearance attributes and fine-grained texture details from the original
asset that may be missing or occluded in the single reference view.
In \cref{sec:experiments}, we present an ablation study of this pipeline and compare alternative
strategies for injecting source multi-view information into the texture generation process.
\subsection{Training and Implementation Details}
\begin{figure}[b]
  \centering
  \setlength{\tabcolsep}{0pt} %
  \renewcommand{\arraystretch}{1.2} %

  \newcommand{\rowlabel}[1]{%
    \raisebox{1.5em}{\textbf{#1}}%
  }

  \begin{tabular*}{\columnwidth}{@{\extracolsep{\fill}} c c c c @{}}
    \toprule
    & \small\textbf{Variant \#1} & \small\textbf{Variant \#2} & \small\textbf{Variant \#3} \\
    \midrule \\[-5ex]
    
    \textbf{DFM} & 
    \raisebox{-0.5\height}{\includegraphics[width=0.18\columnwidth, trim={71.43 35.71 71.43 35.71}, clip]{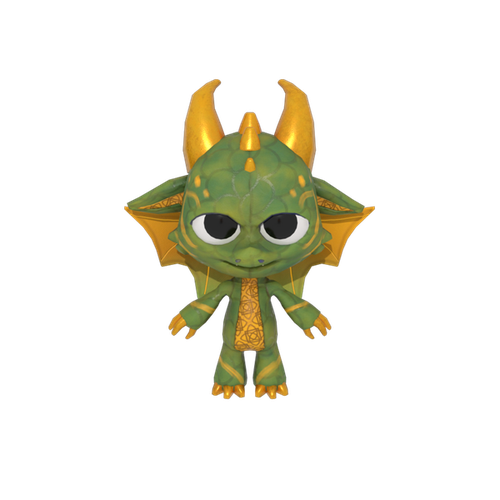}} & 
    \raisebox{-0.5\height}{\includegraphics[width=0.18\columnwidth, trim={71.43 35.71 71.43 35.71}, clip]{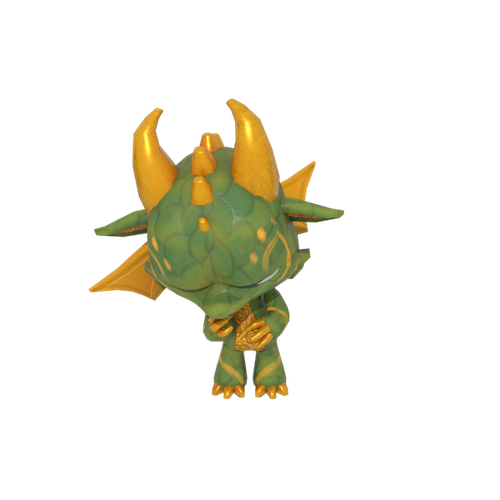}} & 
    \raisebox{-0.5\height}{\includegraphics[width=0.18\columnwidth, trim={71.43 35.71 71.43 35.71}, clip]{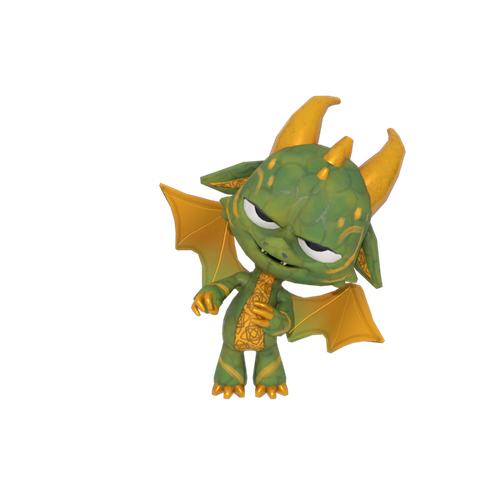}} \\[-6pt]

    \textbf{Parts} & 
    \raisebox{-0.5\height}{\includegraphics[width=0.18\columnwidth, trim={71.43 35.71 71.43 35.71}, clip]{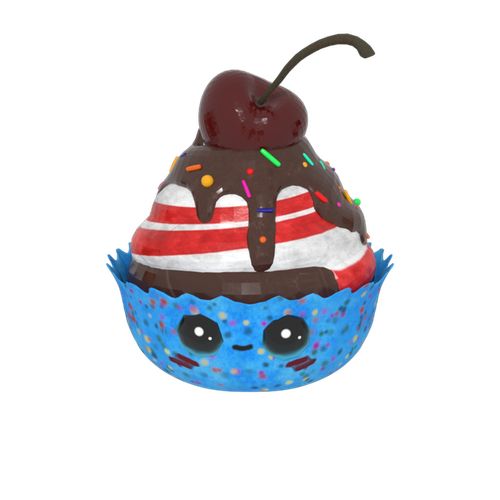}} & 
    \raisebox{-0.5\height}{\includegraphics[width=0.18\columnwidth, trim={71.43 35.71 71.43 35.71}, clip]{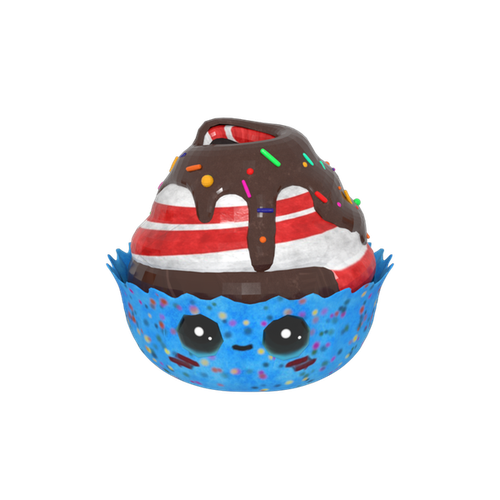}} & 
    \raisebox{-0.5\height}{\includegraphics[width=0.18\columnwidth, trim={71.43 35.71 71.43 35.71}, clip]{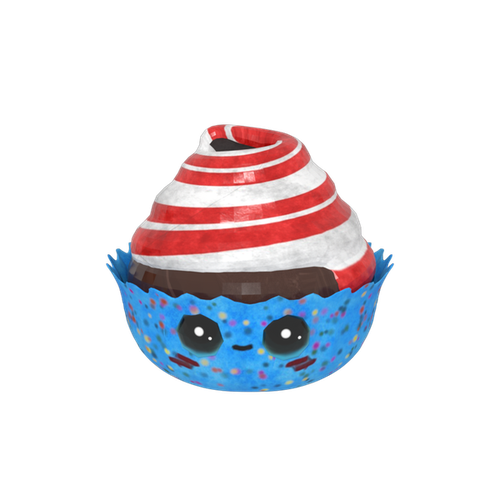}} \\[-6pt]
    
    \bottomrule
  \end{tabular*}

  \caption{\textbf{Dataset.} The first row shows a DFM sample, comprising three temporally distant keyframes. The second row shows a Parts sample, consisting of three variants of the same mesh with progressively removed components.}
  \label{fig:dfm_parts_grid}
\end{figure}

\paragraph{Data}
The dataset comprises 7,430 textured meshes from Objaverse~\cite{deitke2022objaverseuniverseannotated3d}, of which 560 correspond to DFM samples and the remainder are Parts samples. Each Parts sample includes 2--4 variants derived from the original mesh through progressive component removal. Specifically, the first variant represents the complete mesh, the second excludes part 1, the third excludes parts 1 and 2, and so on. For DFM, each sample contains three distinct key frames extracted from the original animation sequence. Examples of DFM and Parts samples are shown in \cref{fig:dfm_parts_grid}. The data generation pipeline is detailed in \supp{appendix:data}. Our ablation in~\cref{sec:experiments} highlights the importance of using the DFM data for identity-preserving global edits.

\paragraph{Training}
For the \algoname geometry pipeline, LoRA adapters of rank 128 were trained on top of the Step1X-3D DiT using a single NVIDIA L40S GPU, with a batch size of 8 for 500K iterations. An input image dropout rate of 0.2 and a shape dropout rate of 0.1 were applied. Due to class imbalance between Parts and DFM samples, DFM samples were upsampled during training with a sampling probability three times higher than that of Parts samples, resulting in an effective DFM sampling probability of 22\%.

For the \algoname texture pipeline, the Step1X-3D UNet adapter layers were fine-tuned on an NVIDIA A100 PCI GPU with a batch size of 4 for 27K iterations. The same DFM upsampling strategy used for geometry training was applied during texture model training.
Both pipelines were trained using the same objectives as their respective base models.

\paragraph{Sampling}
For geometry editing, $k=1024$ latent vectors are sampled from the compact shape representation produced by the shape encoder. During sampling, the following classifier-free guidance formulation is used:
\begin{equation*}
\tilde{\epsilon}_{\theta}^{\mathrm{G}}
= \epsilon^{\mathrm{G}}(\emptyset,\emptyset)
+ s^{\mathrm{G}}_{i}\bigl(\epsilon^{\mathrm{G}}(c_{i},\emptyset) - \epsilon^{\mathrm{G}}(\emptyset,\emptyset)\bigr)
+ s^{\mathrm{G}}_{s}\bigl(\epsilon^{\mathrm{G}}(c_{i},c_{s}) - \epsilon^{\mathrm{G}}(\emptyset,\emptyset)\bigr)
\end{equation*}

Here, $\epsilon^{\mathrm{G}}(\emptyset,\emptyset)$ denotes the unconditioned prediction, $\epsilon^{\mathrm{G}}(c_{i},\emptyset)$ is conditioned on the edited image only, and $\epsilon^{\mathrm{G}}(c_{i},c_{s})$ is conditioned on both the edited image and the source shape. The parameters $s^{\mathrm{G}}_i$ and $s^{\mathrm{G}}_s$ denote the image and shape guidance scales, respectively. At inference time, $s^{\mathrm{G}}_i=2.5$ and $s^{\mathrm{G}}_s=3.5$ are used.

For texture editing, an analogous classifier-free guidance formulation is employed:
\begin{equation*}
\tilde{\epsilon}_{\theta}^{\mathrm{T}}
= \epsilon^{\mathrm{T}}(\emptyset,\emptyset)
+ s^{\mathrm{T}}_{i}\bigl(\epsilon^{\mathrm{T}}(c_{i},\emptyset) - \epsilon^{\mathrm{T}}(\emptyset,\emptyset)\bigr)
+ s^{\mathrm{T}}_{mv}\bigl(\epsilon^{\mathrm{T}}(c_{i},c_{mv}) - \epsilon^{\mathrm{T}}(\emptyset,\emptyset)\bigr)
\end{equation*}

In this case, $s^{\mathrm{T}}_i$ denotes the edited image guidance scale and $s^{\mathrm{T}}_{mv}$ the multi-view guidance scale. At inference time, $s^{\mathrm{T}}_i=2.5$ and $s^{\mathrm{T}}_{mv}=3.5$ are used.

\section{Experiments}
\label{sec:experiments}
\begin{figure*}[t]
  \centering
  \setlength{\tabcolsep}{1pt} 
  \renewcommand{\arraystretch}{0.6}

  \begin{tabular}{@{}ccccccccc@{}}
    \toprule
    \multicolumn{2}{c}{\scriptsize \textbf{Source Textured Mesh}} &
    \multicolumn{1}{c}{\scriptsize \textbf{Edit Condition}} &
    \multicolumn{2}{c}{\scriptsize \textbf{Ours}} &
    \multicolumn{2}{c}{\scriptsize \textbf{3DEditFormer}} &
    \multicolumn{2}{c}{\scriptsize \textbf{EditP23}} \\
    
    \cmidrule(r){1-2} \cmidrule(lr){3-3} \cmidrule(lr){4-5} \cmidrule(lr){6-7} \cmidrule(l){8-9}

    \includegraphics[width=0.109\linewidth, trim={71.43 41.67 71.43 0}, clip]{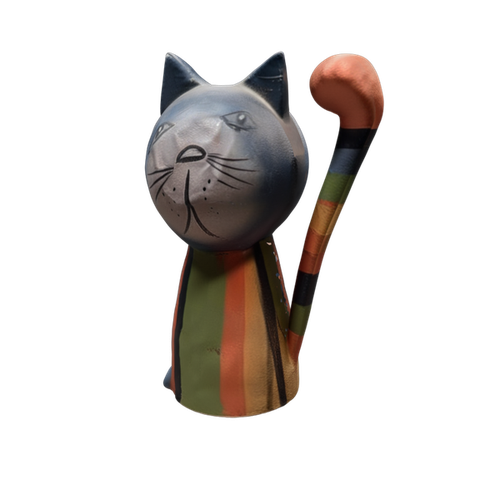} & \includegraphics[width=0.109\linewidth, trim={71.43 41.67 71.43 0}, clip]{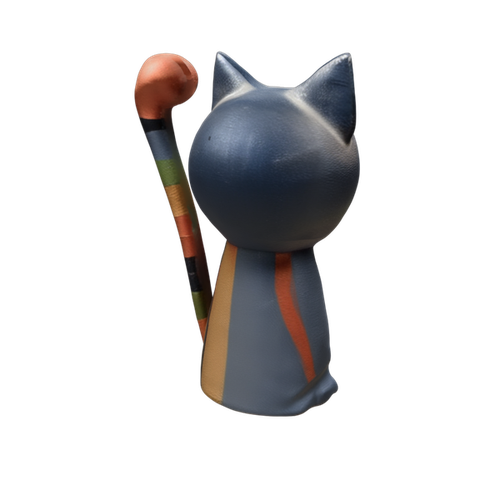} &
    \includegraphics[width=0.109\linewidth, trim={34.18 0 34.18 24.41}, clip]{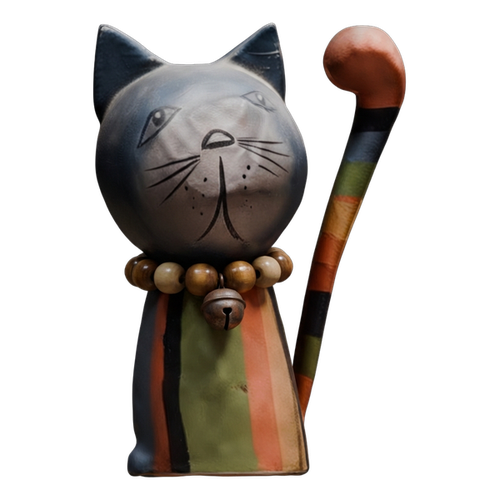} &
    \includegraphics[width=0.109\linewidth, trim={71.43 41.67 71.43 0}, clip]{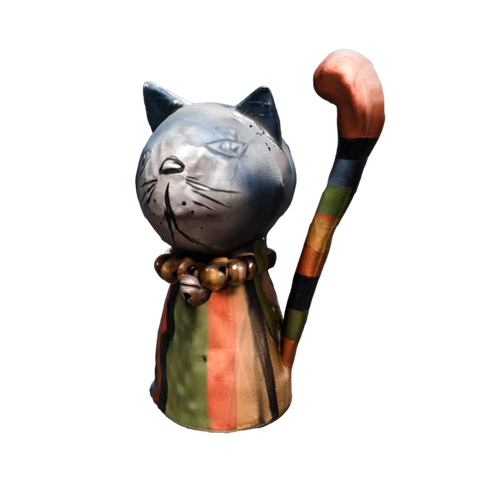} & \includegraphics[width=0.109\linewidth, trim={71.43 41.67 71.43 0}, clip]{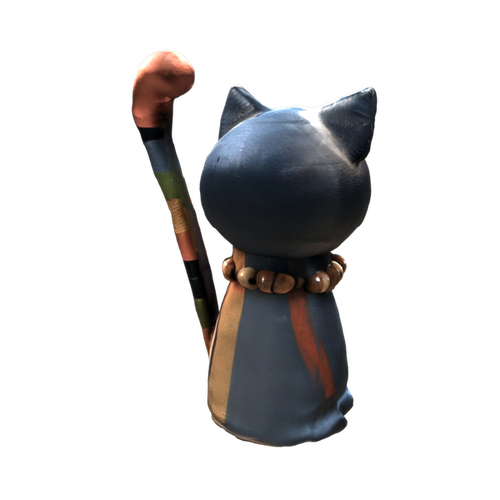} &
    \includegraphics[width=0.109\linewidth, trim={71.43 41.67 71.43 0}, clip]{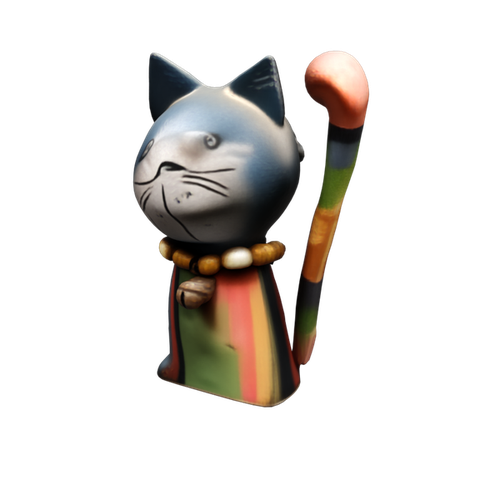} & \includegraphics[width=0.109\linewidth, trim={71.43 41.67 71.43 0}, clip]{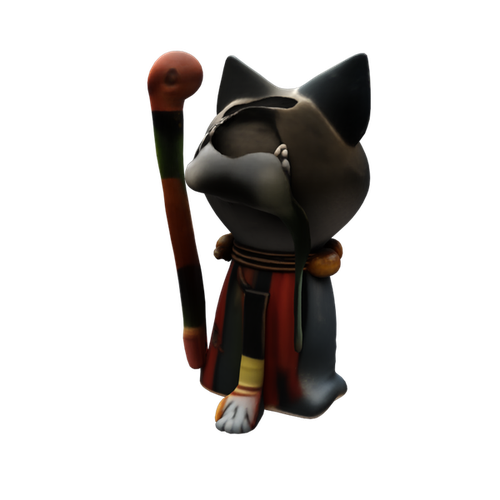} &
    \includegraphics[width=0.109\linewidth, trim={71.43 41.67 71.43 0}, clip]{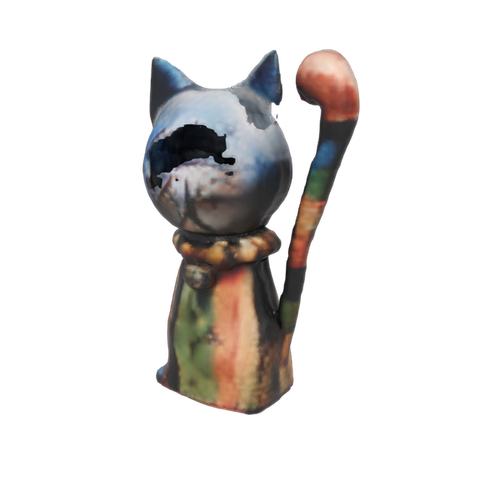} & \includegraphics[width=0.109\linewidth, trim={71.43 41.67 71.43 0}, clip]{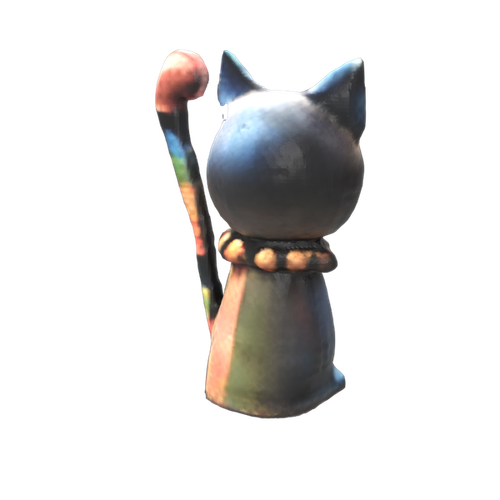} \\ [-12pt]
    
    \includegraphics[width=0.109\linewidth, trim={71.43 41.67 71.43 0}, clip]{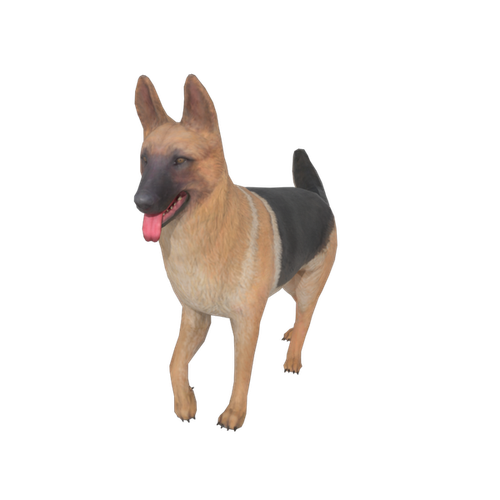} & \includegraphics[width=0.109\linewidth, trim={71.43 41.67 71.43 0}, clip]{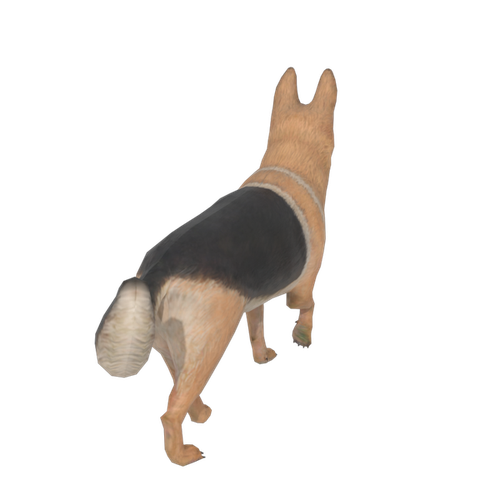} &
    \includegraphics[width=0.109\linewidth, trim={58.59 34.18 58.59 0}, clip]{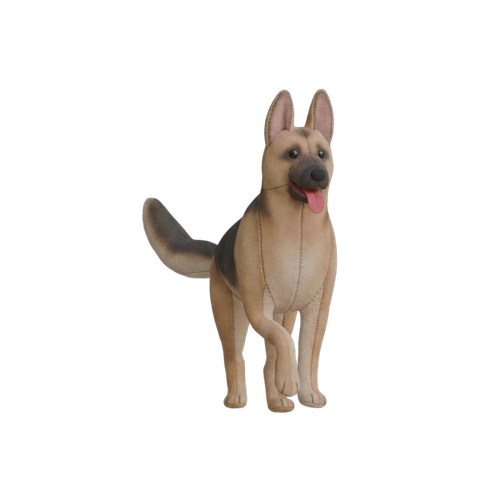} &
    \includegraphics[width=0.109\linewidth, trim={71.43 41.67 71.43 0}, clip]{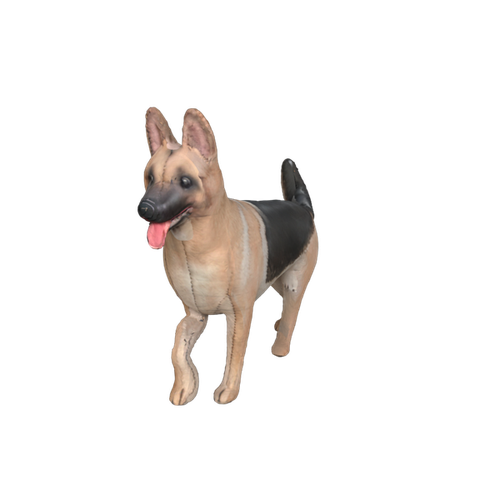} & \includegraphics[width=0.109\linewidth, trim={71.43 41.67 71.43 0}, clip]{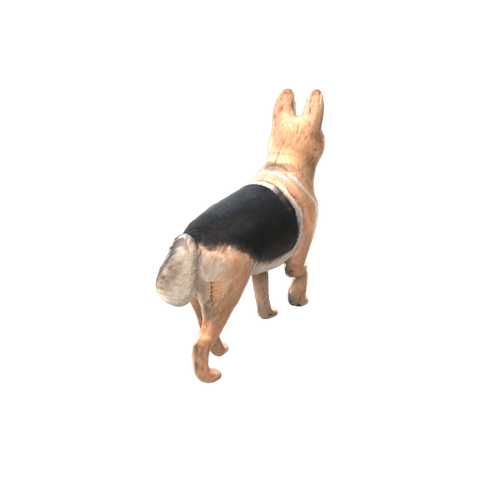} &
    \includegraphics[width=0.109\linewidth, trim={71.43 41.67 71.43 0}, clip]{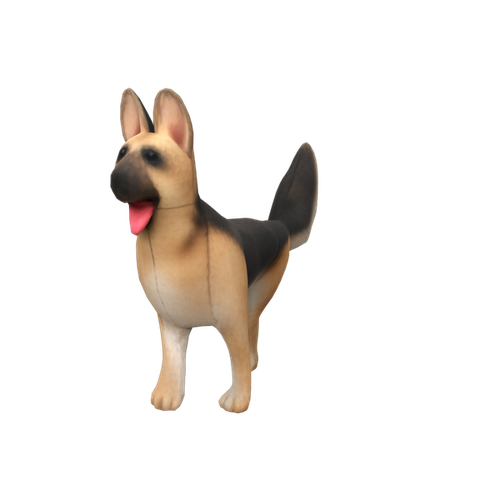} & \includegraphics[width=0.109\linewidth, trim={71.43 41.67 71.43 0}, clip]{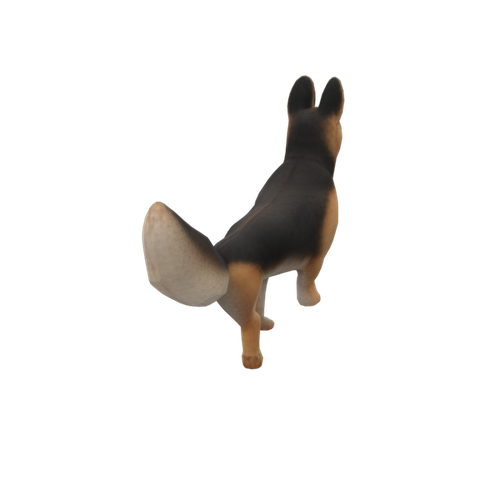} &
    \includegraphics[width=0.109\linewidth, trim={71.43 41.67 71.43 0}, clip]{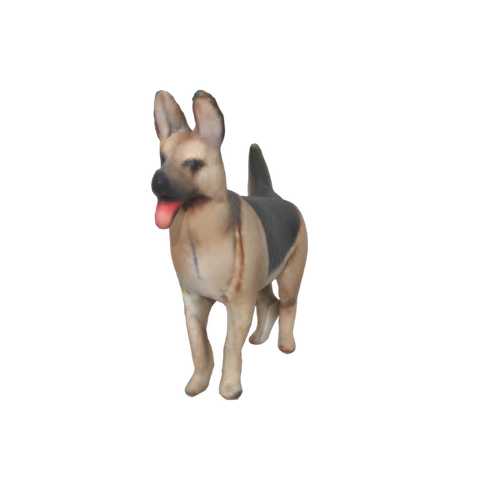} & \includegraphics[width=0.109\linewidth, trim={71.43 41.67 71.43 0}, clip]{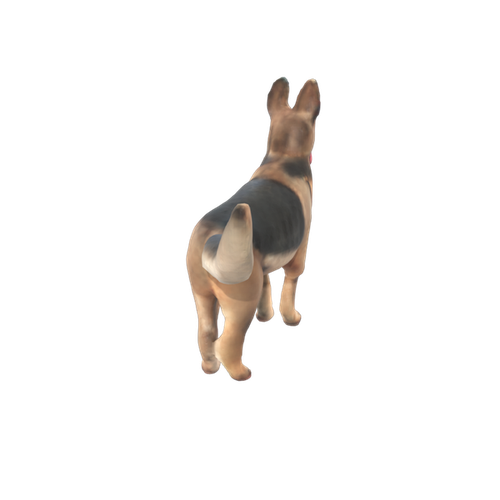} \\ [-9pt]

    \includegraphics[width=0.109\linewidth, trim={71.43 41.67 71.43 0}, clip]{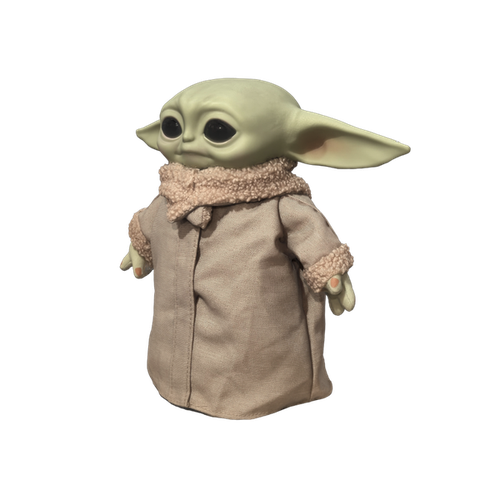} & \includegraphics[width=0.109\linewidth, trim={71.43 41.67 71.43 0}, clip]{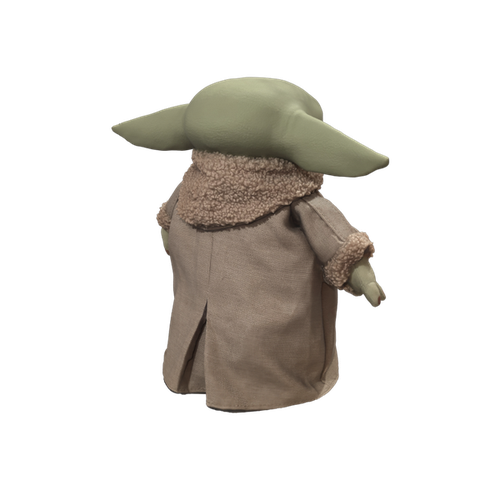} &
    \includegraphics[width=0.109\linewidth, trim={34.18 0 34.18 24.41}, clip]{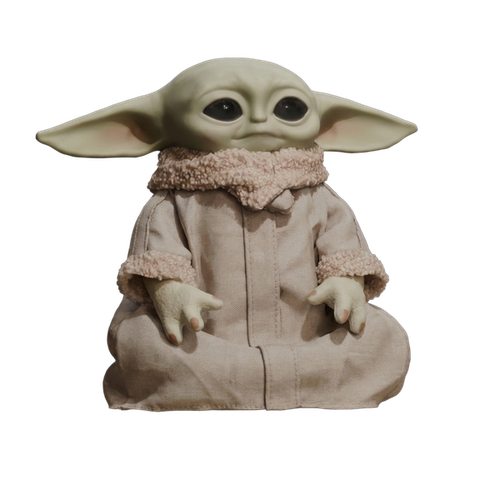} &
    \includegraphics[width=0.109\linewidth, trim={71.43 41.67 71.43 0}, clip]{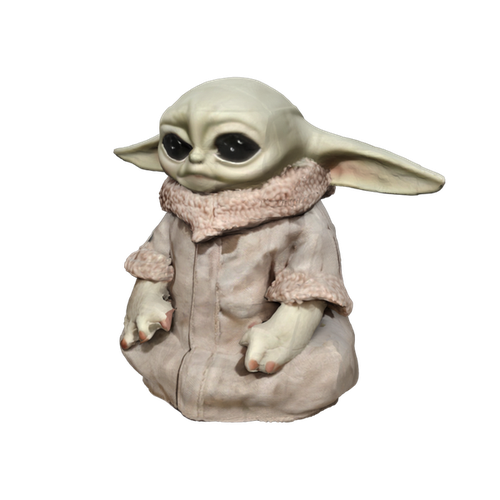} & \includegraphics[width=0.109\linewidth, trim={71.43 41.67 71.43 0}, clip]{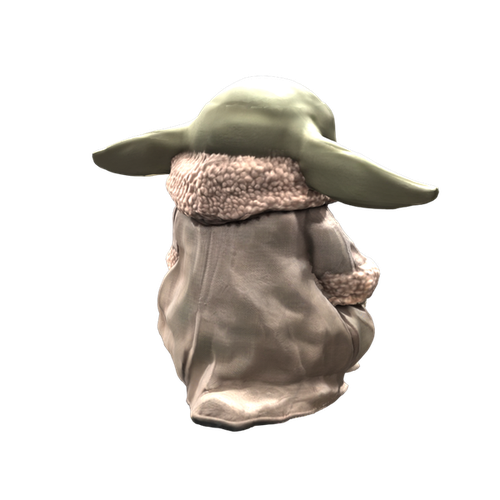} &
    \includegraphics[width=0.109\linewidth, trim={71.43 41.67 71.43 0}, clip]{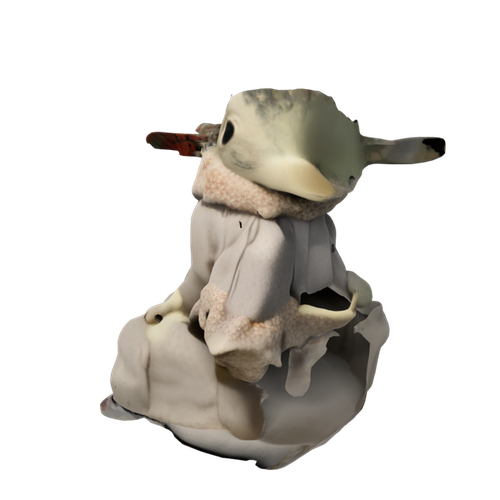} & \includegraphics[width=0.109\linewidth, trim={71.43 41.67 71.43 0}, clip]{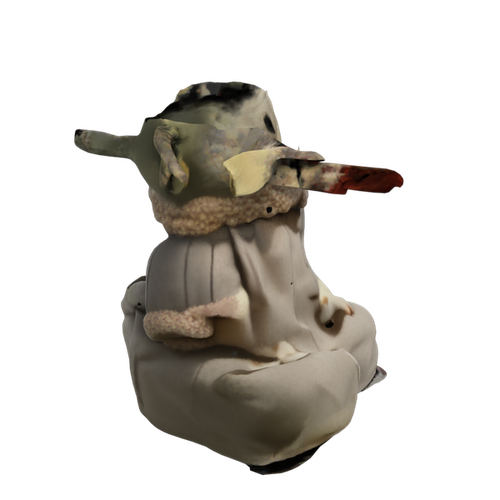} &
    \includegraphics[width=0.109\linewidth, trim={71.43 41.67 71.43 0}, clip]{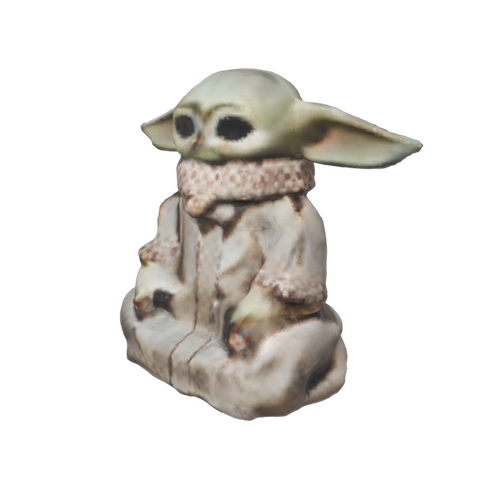} & \includegraphics[width=0.109\linewidth, trim={71.43 41.67 71.43 0}, clip]{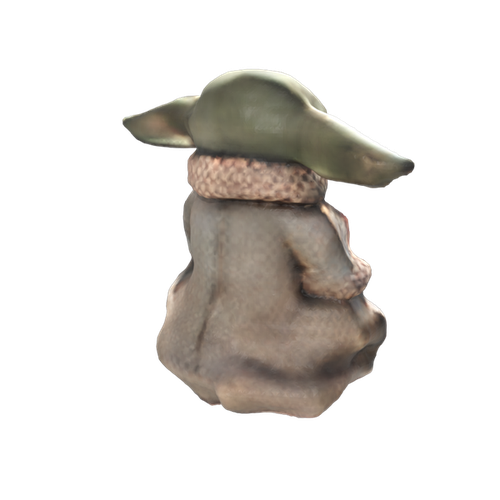} \\ [-10pt]

    \includegraphics[width=0.109\linewidth, trim={71.43 41.67 71.43 0}, clip]{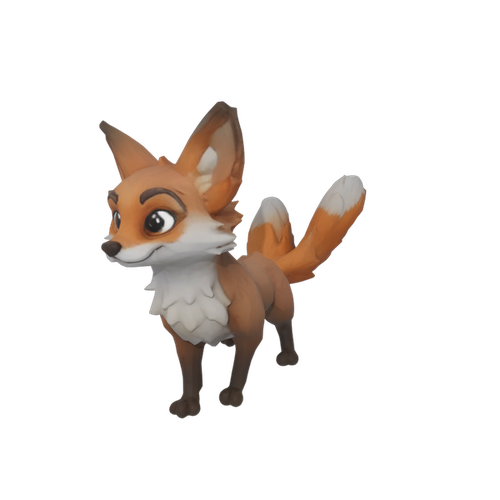} & \includegraphics[width=0.109\linewidth, trim={71.43 41.67 71.43 0}, clip]{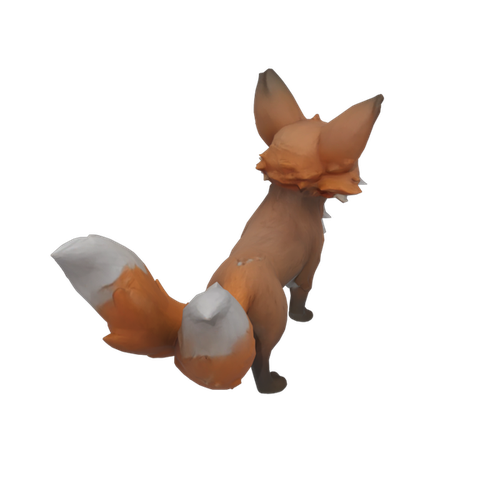} &
    \includegraphics[width=0.109\linewidth, trim={58.59 34.18 58.59 0}, clip]{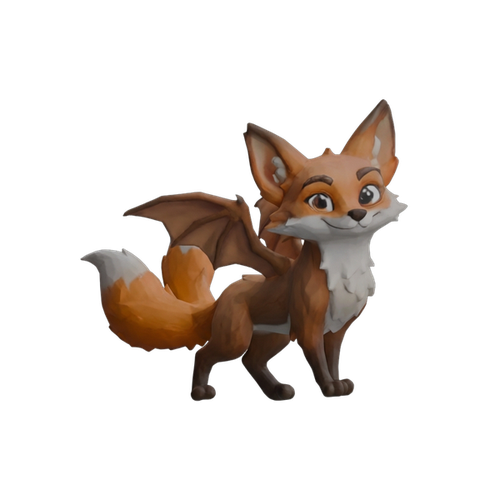} &
    \includegraphics[width=0.109\linewidth, trim={71.43 41.67 71.43 0}, clip]{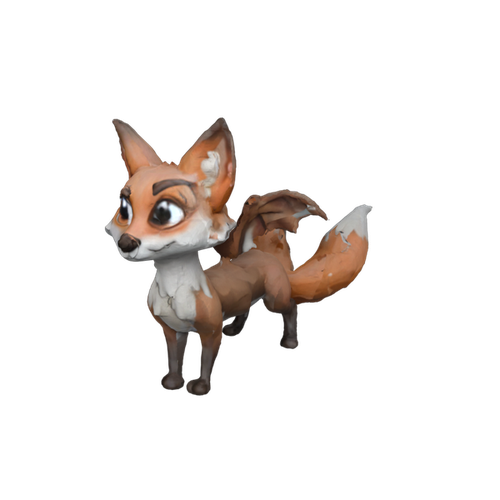} & \includegraphics[width=0.109\linewidth, trim={71.43 41.67 71.43 0}, clip]{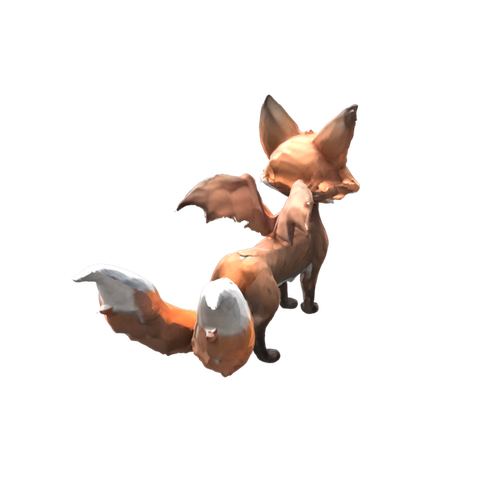} &
    \includegraphics[width=0.109\linewidth, trim={71.43 41.67 71.43 0}, clip]{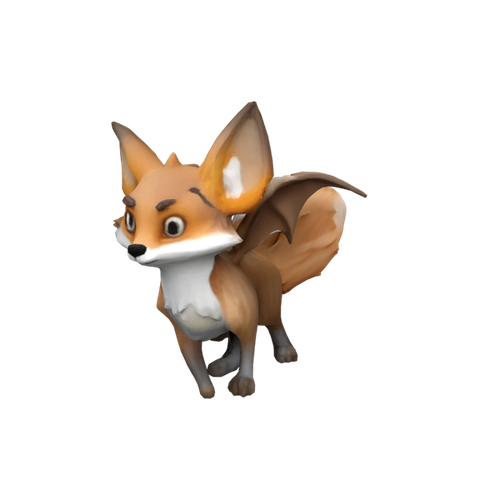} & \includegraphics[width=0.109\linewidth, trim={71.43 41.67 71.43 0}, clip]{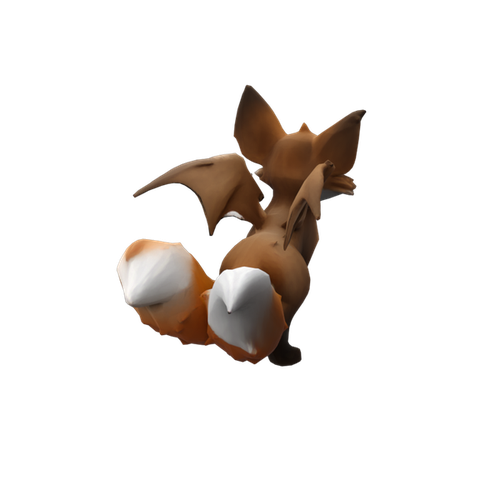} &
    \includegraphics[width=0.109\linewidth, trim={71.43 41.67 71.43 0}, clip]{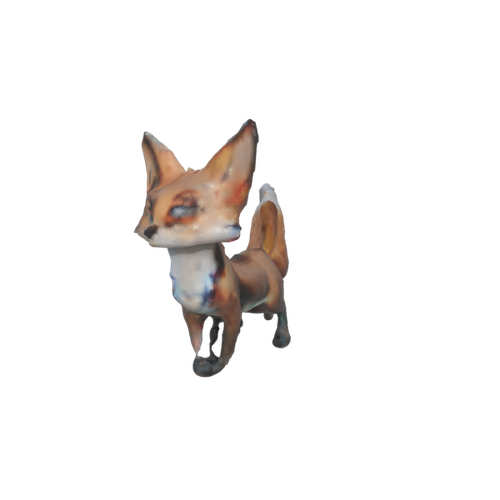} & \includegraphics[width=0.109\linewidth, trim={71.43 41.67 71.43 0}, clip]{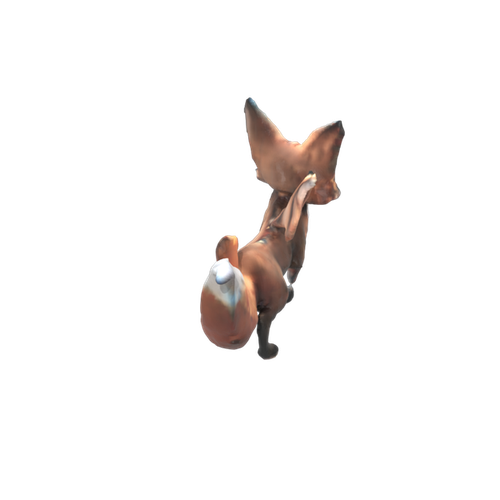} \\ [-3pt]

    \includegraphics[width=0.109\linewidth, trim={71.43 41.67 71.43 0}, clip]{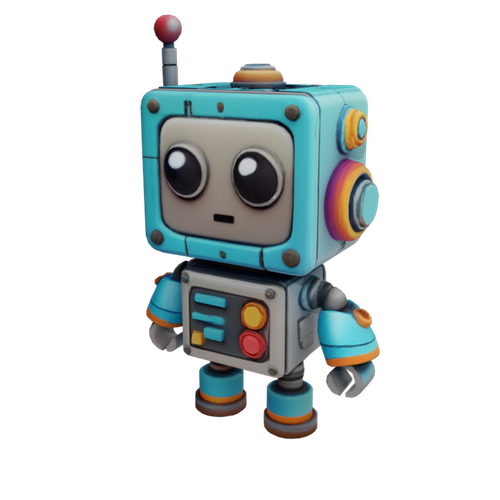} & \includegraphics[width=0.109\linewidth, trim={71.43 41.67 71.43 0}, clip]{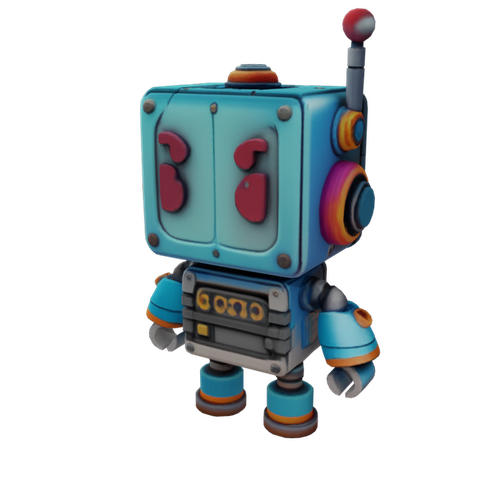} &
    \includegraphics[width=0.109\linewidth, trim={58.59 34.18 58.59 0}, clip]{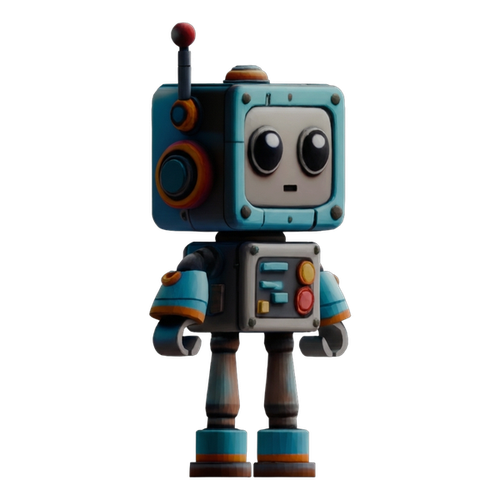} &
    \includegraphics[width=0.109\linewidth, trim={71.43 41.67 71.43 0}, clip]{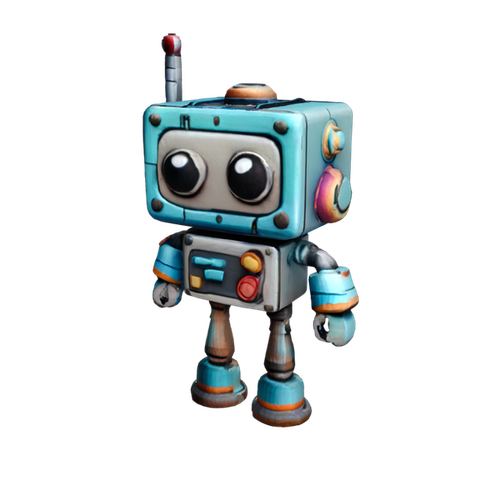} & \includegraphics[width=0.109\linewidth, trim={71.43 41.67 71.43 0}, clip]{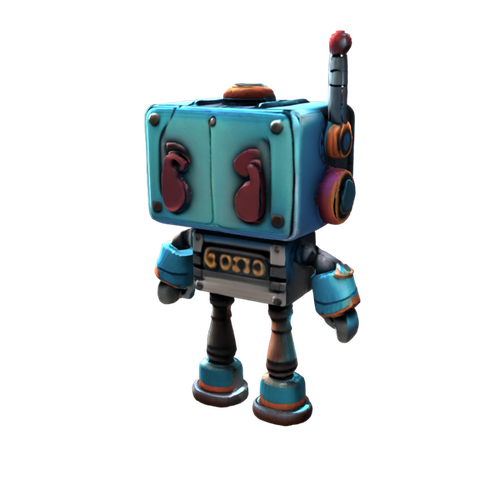} &
    \includegraphics[width=0.109\linewidth, trim={71.43 41.67 71.43 0}, clip]{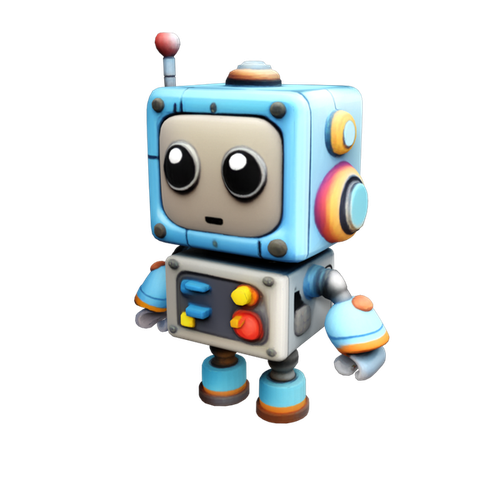} & \includegraphics[width=0.109\linewidth, trim={71.43 41.67 71.43 0}, clip]{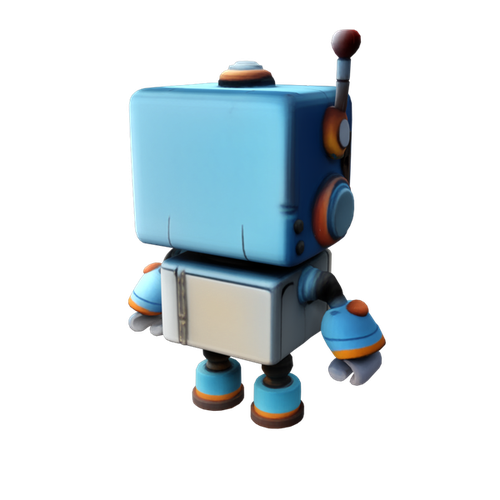} &
    \includegraphics[width=0.109\linewidth, trim={71.43 41.67 71.43 0}, clip]{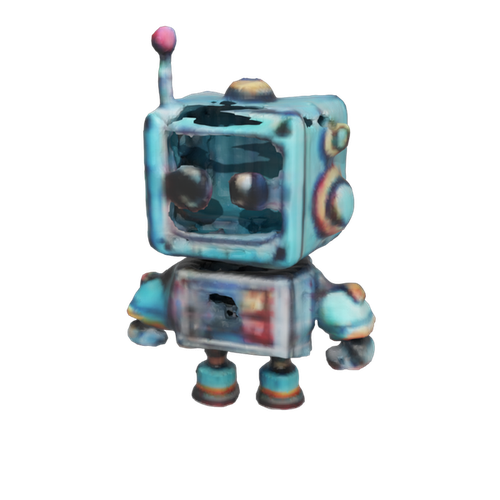} & \includegraphics[width=0.109\linewidth, trim={71.43 41.67 71.43 0}, clip]{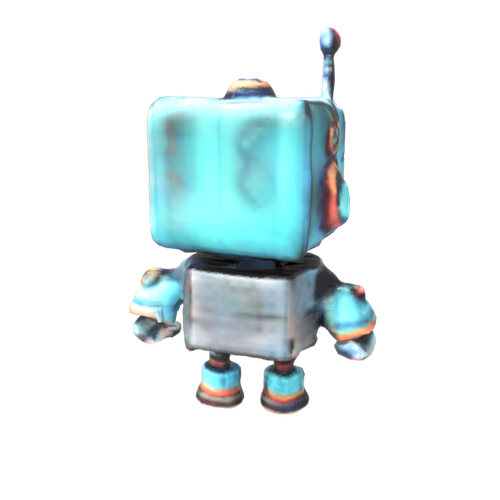} \\

    \bottomrule
  \end{tabular}

  \caption{\textbf{Qualitative Comparisons on BenchUp.} We compare our method against 3DEditFormer and EditP23. The first two columns show the source mesh (Front and Back views), followed by the editing condition. Our method (Cols 4--5) offers better condition alignment and preservation of the object's identity compared to the baselines.}
  \label{fig:qualitative_comparison}
\end{figure*}

\subsection{Evaluation Setup}
\subsubsection{Benchmark}

To evaluate our model compared to the state-of-the-art we first define a new benchmark designed to evaluate global 3D mesh editing, consisting of 24 diverse meshes and 100 edit conditions. The benchmark spans a wide range of editing operations, from localized geometric modifications to large-scale global transformations, including style deformations and pose changes. This benchmark addresses a gap in existing evaluation protocols, which primarily focus on localized edits~\cite{voxhammer2025} or exhibit limited mesh diversity~\cite{xia20253deditformer}, and are therefore insufficient for assessing global editing capabilities.

To construct the benchmark, we curate source geometries from Objaverse and augment them with meshes generated using TRELLIS 2.0 and Hunyuan 3d 2.0. We then generate diverse and semantically meaningful edit instructions using the multimodal capabilities of Gemini 3 Pro (Image Preview) \cite{Gemini3Pro2025}. For each source mesh, we render 20 views with azimuth uniformly sampled in the range $[-30^\circ, 30^\circ]$. These views are provided together with a meta-prompt specifying the edit categories: \textit{Parts}, \textit{Global-Deformation}, \textit{Global-Pose Change}, and \textit{Global-Texture/Material}. For each view, a target category is sampled and the model is instructed to generate a corresponding edit while preserving all other aspects of the image, including lighting, viewpoint, and the object’s overall identity. The exact meta-prompt we used in shared in \supp{appendix:eval}. Finally, we manually filter the generated examples to remove negligible edits or failures to preserve object identity.

\subsubsection{Metrics}

We evaluate performance along two complementary axes: \emph{Condition Alignment}, which measures how well the edited mesh reflects the target edit specified by the condition image, and \emph{Occluded Region Fidelity}, which assesses whether regions that are not visible, or only partially visible, in the condition image are properly preserved.
For both axes, semantic correspondence is measured using feature similarity computed with CLIP-I \cite{hessel2022clipscorereferencefreeevaluationmetric} and DINO-I (using a DINOv2 backbone \cite{oquab2024dinov2learningrobustvisual}). For \emph{Condition Alignment}, we additionally report SSIM to evaluate pixel-level fidelity, LPIPS \cite{zhang2018perceptual} to capture perceptual similarity, and CLIP-Dir, which measures alignment between the source-to-prediction and source-to-condition in CLIP space.

\subsubsection{Baselines}

We compare against two recent mask-free 3D editing methods representing distinct paradigms. \textbf{EditP23} propagates a single edited view to a consistent multi-view grid using
edit-aware denoising during inference. It computes a differential edit direction via delta velocity prediction to isolate the edit signal.
The resulting multi-view grid is then reconstructed into a 3D mesh using Instant Mesh \cite{xu2024instantmeshefficient3dmesh}. \textbf{3DEditFormer} finetunes TRELLIS Dual-Attention layers and injects
multi-stage source-shape features extrarcted during inference in a frozen TRELLIS instance, combining fine-grained structural features from late diffusion timesteps and
semantic transition features from early timesteps.

\subsection{Quantitative Results}
Quantitative results are reported in \cref{tab:comparison_filtered_single}. Our method outperforms previous methods across all metrics. Notably, the results highlight that \algoname achieves superior condition alignment without sacrificing preservation of unedited regions, highlighting its ability to resolve the typical trade-off between edit fidelity and source-shape consistency.
Additional evaluations of our method’s reconstruction fidelity compared to the baselines are provided in \supp{appendix:eval}.
\begin{table}[t]
\small
\centering
\caption{\textbf{Quantitative comparison on BenchUp.} We evaluate image-guided 3D shape editing by measuring \textbf{Condition Alignment} and \textbf{Occluded Region Fidelity}. Bold indicates best performance.}
\label{tab:comparison_filtered_single}
\setlength{\tabcolsep}{1.2mm} %
\resizebox{\linewidth}{!}{
\begin{tabular}{lccccccc}
\toprule
\multirow{2}{*}{Method} & \multicolumn{5}{c}{Condition Alignment} & \multicolumn{2}{c}{Occluded Region Fid.} \\
\cmidrule(lr){2-6} \cmidrule(lr){7-8}
 & SSIM$\uparrow$ & LPIPS$\downarrow$ & CLIP-I$\uparrow$ & DINO-I$\uparrow$ & C-Dir$\uparrow$ & CLIP-I$\uparrow$ & DINO-I$\uparrow$ \\
\midrule
3DEditFormer & 0.733 & 0.270 & 0.908 & 0.849 & 0.441 & 0.877 & 0.736 \\
EditP23 & 0.759 & 0.254 & 0.917 & 0.851 & 0.455 & 0.880 & 0.748 \\
Ours & \textbf{0.763} & \textbf{0.198} & \textbf{0.943} & \textbf{0.915} & \textbf{0.520} & \textbf{0.928} & \textbf{0.878} \\
\bottomrule
\end{tabular}
}
\end{table}

\subsection{Qualitative Results}
We present visual comparisons in \cref{fig:qualitative_comparison}.
As shown, our method demonstrates superior \emph{Condition Alignment} across the different edit types defined in our benchmark.
Notably, in geometric modifications our approach successfully propagates the structural changes commanded by the target image into the 3D mesh,
whereas baseline methods often struggle to integrate the new topology with the source geometry. 
Crucially, we observe a significant improvement in \emph{Occluded Region Fidelity}.
While the other methods often hallucinate inconsistent textures or degrade geometry in regions not visible in the conditioning image, our approach preserves the source object’s semantic identity in these occluded areas.
\subsection{User Study}
\begin{figure}
  \centering
  \includegraphics[width=1\linewidth]{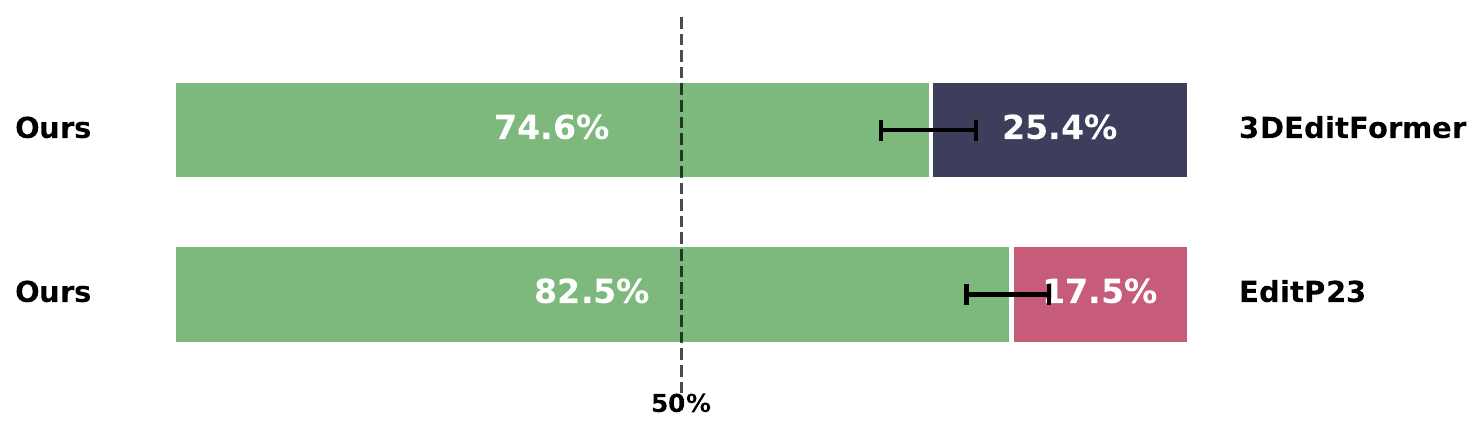}
  \caption{\textbf{User study results.} We show the \% of participants who preferred our method when compared to each baseline. Error bars are the 95\% confidence interval.}
  \label{fig:user_study}
\end{figure}

To further validate our results, we conduct a user preference study comparing our method against our baselines. We use a two alternative forced choice setup where participants view the original object, target edit, and two results (ours vs. one baseline), then select which better achieves the edit while preserving original details. We collected 664 comparisons from 34 participants. Results in \cref{fig:user_study} show \algoname   was strongly preferred by participants.

\subsection{Ablation}

\begin{table}[t]
\small
\centering
\caption{\textbf{Ablation Study Results.} Comparison of variants. \textbf{Bold} denotes best, \underline{underline} second best.}
\label{tab:ablation_comparison_single}
\setlength{\tabcolsep}{1.0mm} %
\resizebox{\linewidth}{!}{
\begin{tabular}{lccccccc}
\toprule
\multirow{2}{*}{Method} & \multicolumn{5}{c}{Condition Alignment} & \multicolumn{2}{c}{Occluded Region Fid.} \\
\cmidrule(lr){2-6} \cmidrule(lr){7-8}
 & SSIM$\uparrow$ & LPIPS$\downarrow$ & CLIP-I$\uparrow$ & DINO-I$\uparrow$ & C-Dir$\uparrow$ & CLIP-I$\uparrow$ & DINO-I$\uparrow$ \\
\midrule
w/o Motion & \underline{0.769} & \underline{0.196} & \textbf{0.943} & 0.909 & 0.505 & \textbf{0.932} & \textbf{0.884} \\
256 latents & 0.766 & 0.206 & \underline{0.941} & 0.909 & \underline{0.525} & 0.917 & 0.869 \\
512 latents & 0.768 & 0.220 & 0.918 & 0.868 & 0.506 & 0.905 & 0.837 \\
Concat MV & \textbf{0.779} & \textbf{0.187} & \textbf{0.943} & \underline{0.912} & \textbf{0.555} & 0.897 & 0.832 \\
\midrule
Ours & 0.763 & 0.198 & \textbf{0.943} & \textbf{0.915} & 0.520 & \underline{0.928} & \underline{0.878} \\
\bottomrule
\end{tabular}
}
\end{table}

\subsubsection{Geometry Editing Pipeline}
We conduct two ablation studies on the geometry editing pipeline. First, we vary the number of latent vectors sampled from the source shape’s latent representation, evaluating three settings: 256, 512, and 1024 latents, where 1024 is the configuration used in all other experiments. While earlier analyses indicate that reconstructing shapes from 256 or 512 randomly sampled latents yields results comparable to using the full latent set, this ablation demonstrates that training with 1024 latents leads to substantially improved performance.
Quantitative results for this study are reported in \cref{tab:ablation_comparison_single}.
\begin{figure}[t] 
  \centering 
  \setlength{\tabcolsep}{0pt} 
  \renewcommand{\arraystretch}{0.85}

  \begin{tabular*}{\linewidth}{@{\extracolsep{\fill}} cccc @{}} 
    \toprule 
    \scriptsize \textbf{Input Image} & 
    \scriptsize \textbf{Source Mesh} & 
    \scriptsize \textbf{Ours (w/ DFM)} & 
    \scriptsize \textbf{Baseline (w/o DFM)} \\ 
    \midrule 
    \\[-4ex] %
    
    \includegraphics[width=0.2\linewidth, trim={48.83 34.18 48.83 34.18}, clip]{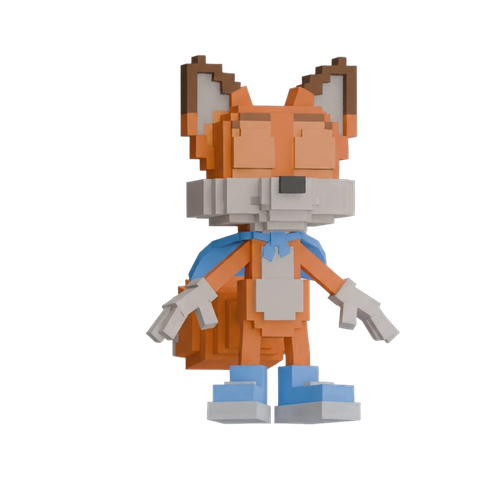} & 
    \includegraphics[width=0.2\linewidth, trim={83.33 41.67 83.33 41.67}, clip]{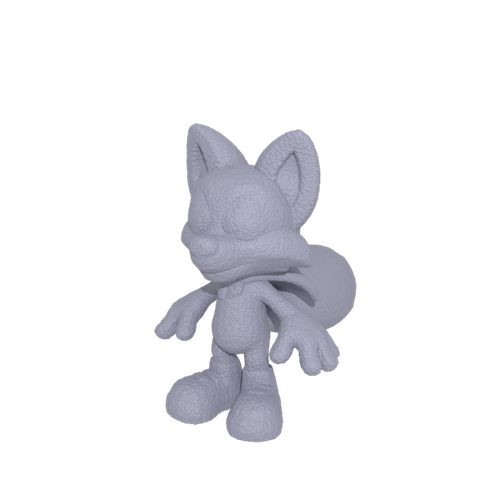} & 
    \includegraphics[width=0.2\linewidth, trim={59.52 41.67 59.52 41.67}, clip]{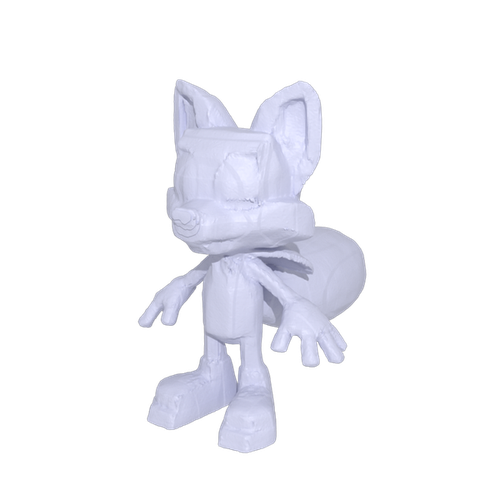} & 
    \includegraphics[width=0.2\linewidth, trim={83.33 41.67 83.33 41.67}, clip]{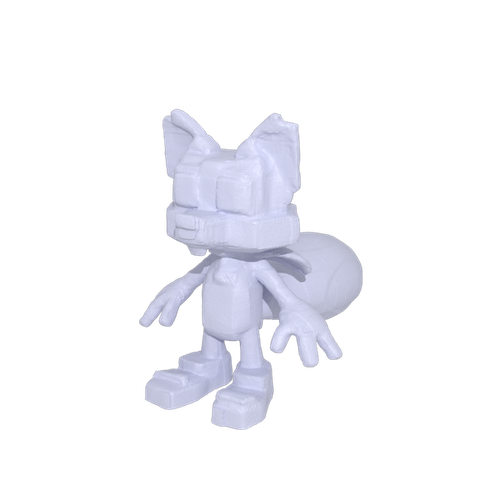} \\ [-8pt]
    
    \includegraphics[width=0.2\linewidth, trim={48.83 14.65 48.83 14.65}, clip]{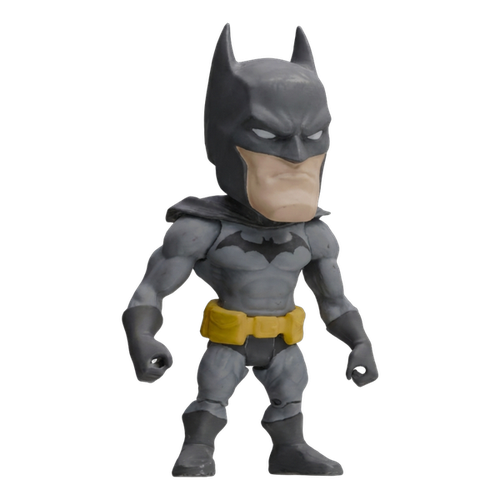} & 
    \includegraphics[width=0.2\linewidth, trim={83.33 41.67 83.33 41.67}, clip]{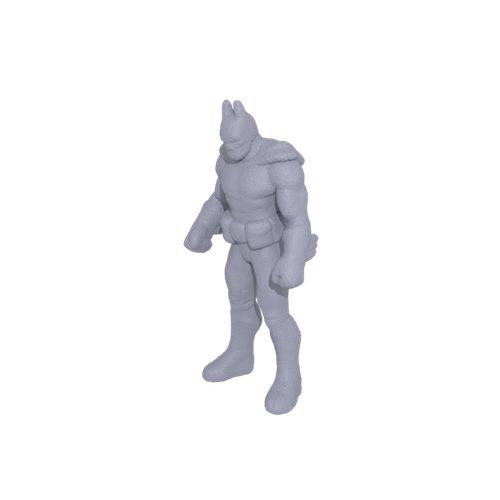} & 
    \includegraphics[width=0.2\linewidth, trim={83.33 41.67 83.33 41.67}, clip]{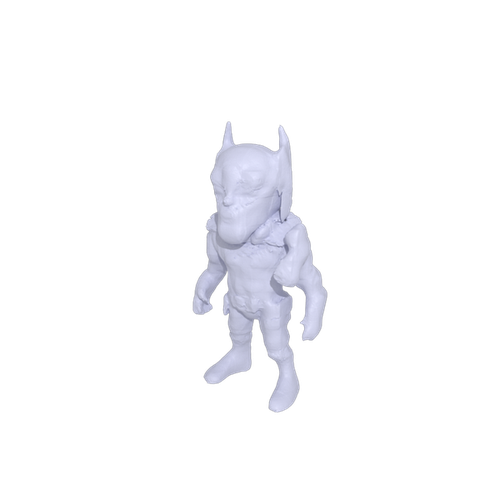} & 
    \includegraphics[width=0.2\linewidth, trim={83.33 41.67 83.33 41.67}, clip]{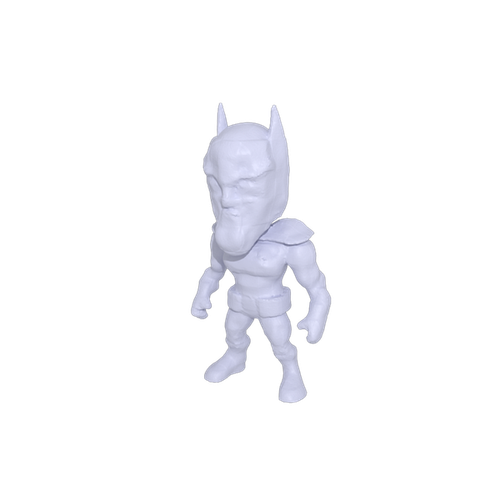} \\ [-8pt]
    
    \includegraphics[width=0.2\linewidth, trim={48.83 34.18 48.83 34.18}, clip]{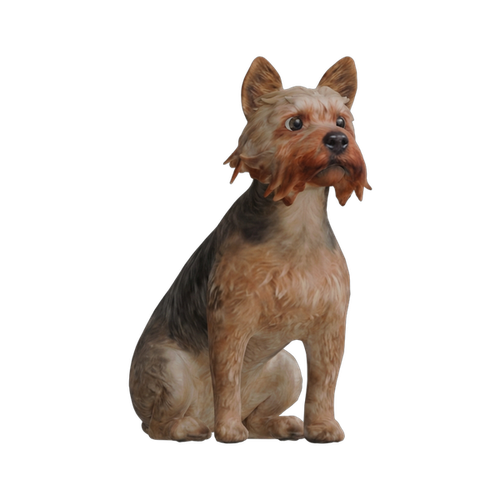} & 
    \includegraphics[width=0.2\linewidth, trim={83.33 41.67 83.33 41.67}, clip]{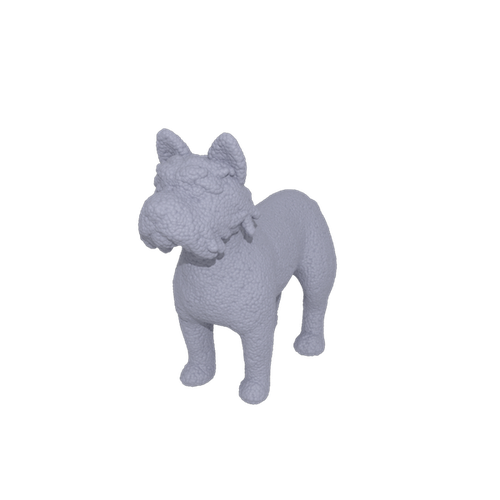} & 
    \includegraphics[width=0.2\linewidth, trim={95.24 53.57 95.24 53.57}, clip]{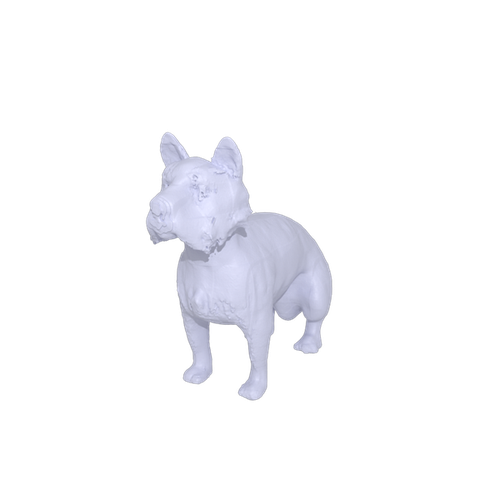} & 
    \includegraphics[width=0.2\linewidth, trim={95.24 53.57 95.24 53.57}, clip]{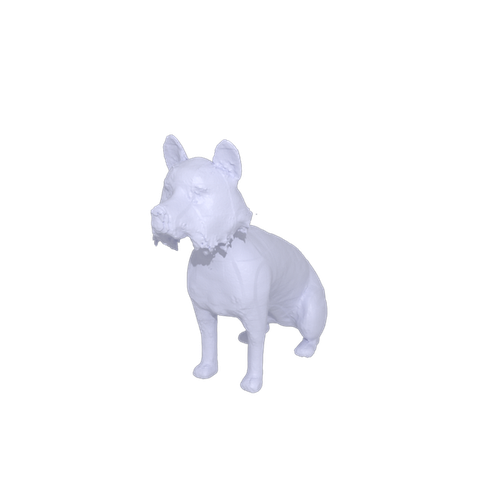} \\

    \bottomrule 
  \end{tabular*} 

  \caption{\textbf{Geometry Editing Ablation.} Qualitative comparison showing that DFM samples (third column) improve pose and global edits compared to the baseline (fourth column).} 
  \label{fig:motion_table} 
\end{figure}

Next, we evaluate the impact of including DFM training samples. Qualitatively, DFM samples not only enable pose changes but also improve the model’s ability to perform global edits and to deviate, within reasonable limits, from the source mesh at a global scale. As shown in \cref{fig:motion_table}, a model trained under the same setting but without DFM samples exhibits noticeably degraded performance compared to the full model.
Quantitatively (\cref{tab:ablation_comparison_single}), the variant trained without DFM achieves the highest scores on occluded-region fidelity metrics. This can be attributed to parts-based supervision, which introduces changes primarily in localized regions and encourages the model to adhere closely to the original shape layout. However, this comes at the cost of reduced flexibility and a weaker ability to impose coherent global structural changes.

\subsubsection{Texture Editing Pipeline}
\begin{figure}[t]
\centering
\setlength{\tabcolsep}{2pt}
\renewcommand{\arraystretch}{1.05}

\begin{tabular*}{\columnwidth}{@{\extracolsep{\fill}}
>{\centering\arraybackslash}m{0.16\columnwidth}
>{\centering\arraybackslash}m{0.12\columnwidth}
>{\centering\arraybackslash}m{0.12\columnwidth}
>{\centering\arraybackslash}m{0.12\columnwidth}
>{\centering\arraybackslash}m{0.12\columnwidth}
>{\centering\arraybackslash}m{0.12\columnwidth}
>{\centering\arraybackslash}m{0.12\columnwidth}
@{}}
\toprule
\parbox[c]{0.16\columnwidth}{\centering\textbf{Edited}\\\textbf{Image}}
& \multicolumn{2}{c}{\raisebox{-0.5\height}{\includegraphics[width=0.20\columnwidth]{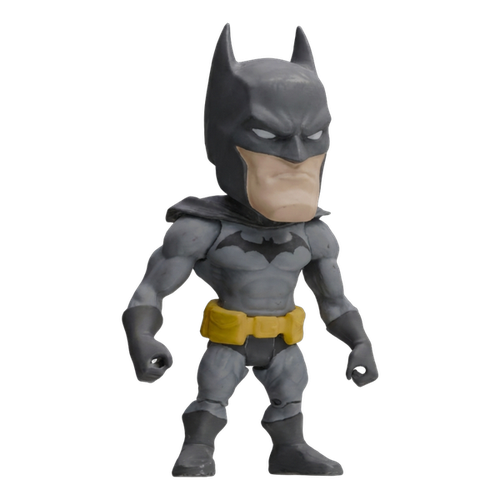}}}
& \multicolumn{2}{c}{\raisebox{-0.5\height}{\includegraphics[width=0.20\columnwidth]{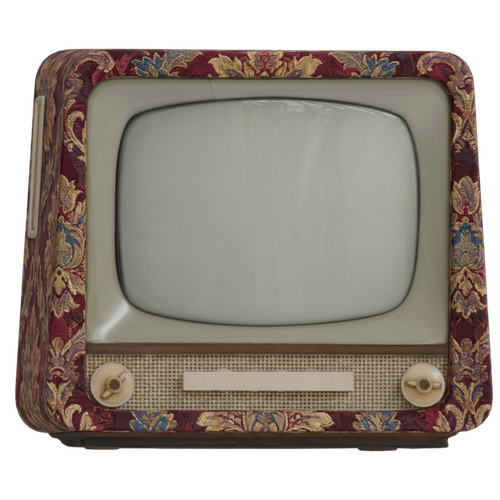}}}
& \multicolumn{2}{c}{\raisebox{-0.5\height}{\includegraphics[width=0.20\columnwidth]{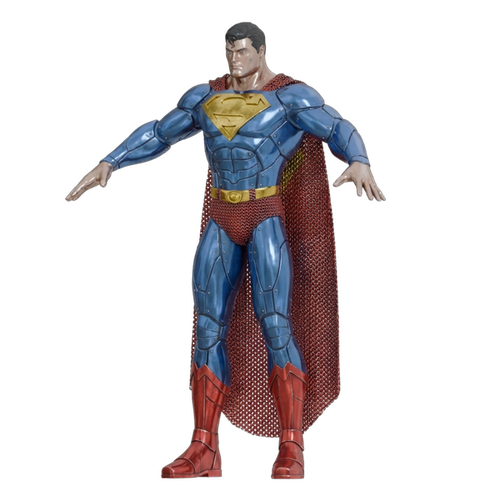}}}
\\
\midrule
& Front & Back & Front & Back & Front & Back \\
\midrule
\parbox[c]{0.16\columnwidth}{\centering\textbf{Source}\\\textbf{Texture}}
& \includegraphics[trim=59.52pt 59.52pt 59.52pt 59.52pt, clip, width=0.12\columnwidth]{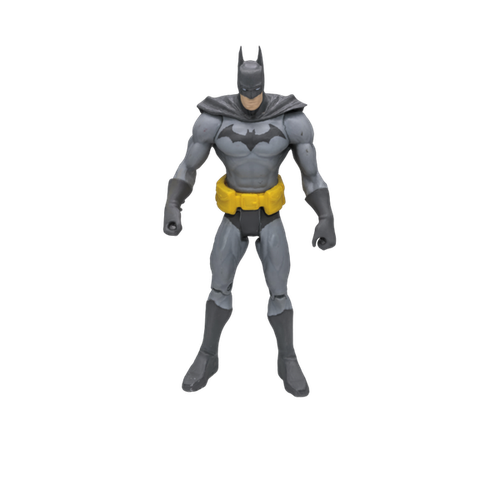}
& \includegraphics[trim=59.52pt 59.52pt 59.52pt 59.52pt, clip, width=0.12\columnwidth]{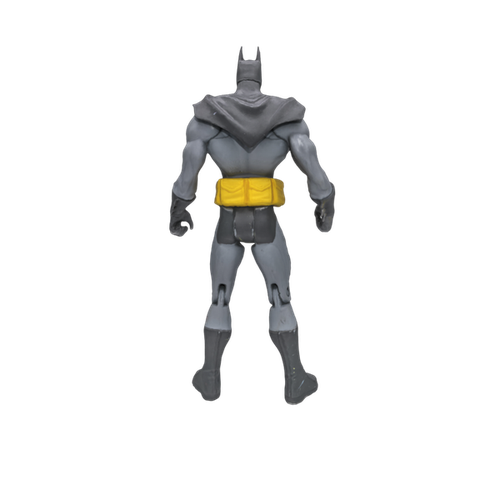}
& \includegraphics[trim=59.52pt 59.52pt 59.52pt 59.52pt, clip, width=0.12\columnwidth]{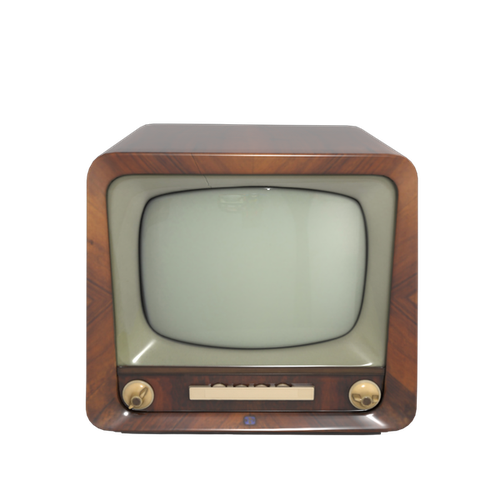}
& \includegraphics[trim=59.52pt 59.52pt 59.52pt 59.52pt, clip, width=0.12\columnwidth]{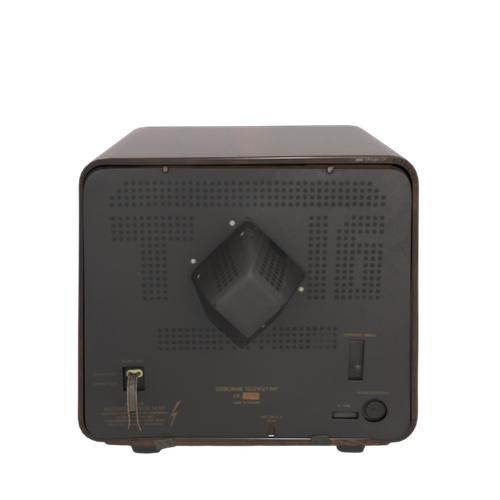}
& \includegraphics[trim=59.52pt 59.52pt 59.52pt 59.52pt, clip, width=0.12\columnwidth]{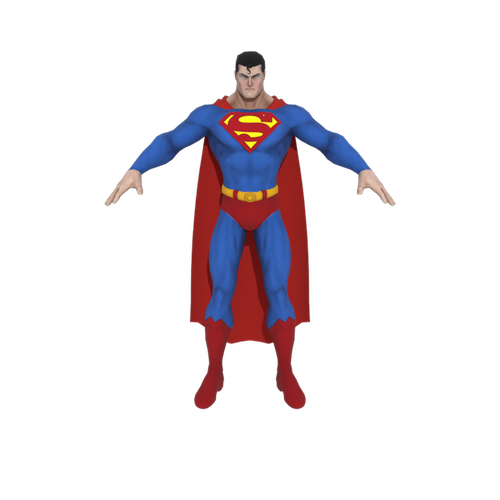}
& \includegraphics[trim=59.52pt 59.52pt 59.52pt 59.52pt, clip, width=0.12\columnwidth]{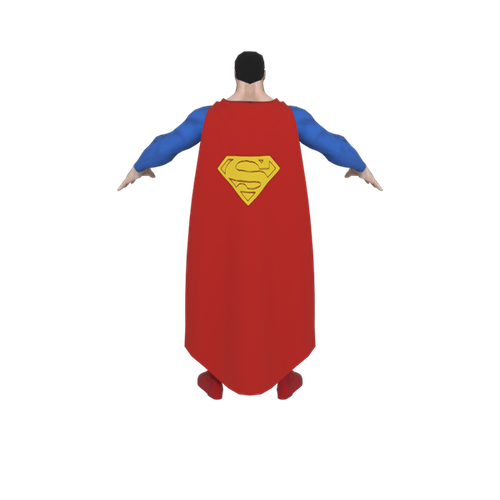}
\\
\midrule
\parbox[c]{0.16\columnwidth}{\centering $s_{i}{=}6.5$\\$s_{mv}{=}2.5$}
& \includegraphics[trim=59.52pt 59.52pt 59.52pt 59.52pt, clip, width=0.12\columnwidth]{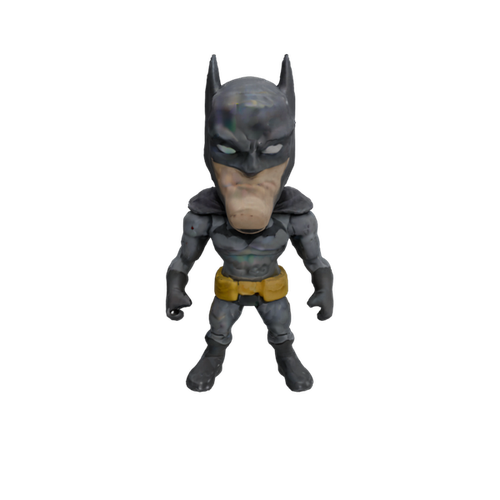}
& \includegraphics[trim=59.52pt 59.52pt 59.52pt 59.52pt, clip, width=0.12\columnwidth]{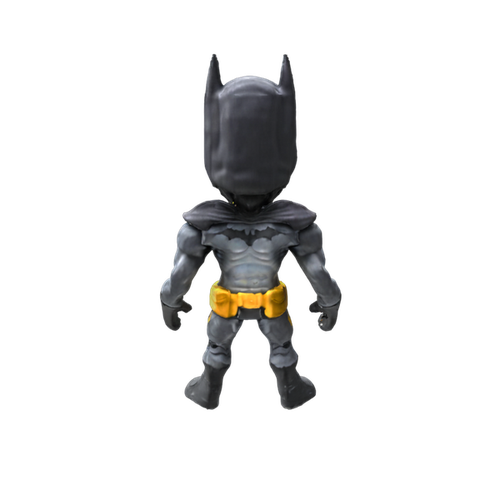}
& \includegraphics[trim=59.52pt 59.52pt 59.52pt 59.52pt, clip, width=0.12\columnwidth]{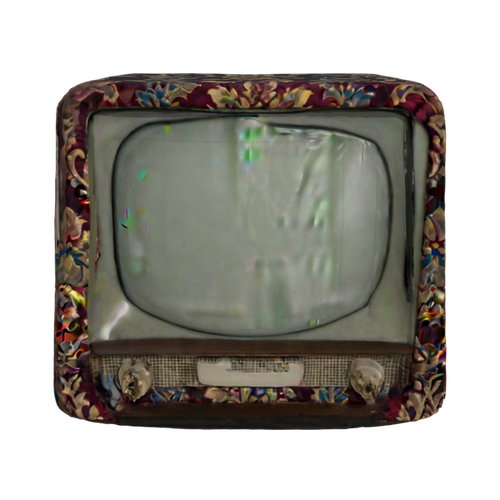}
& \includegraphics[trim=59.52pt 59.52pt 59.52pt 59.52pt, clip, width=0.12\columnwidth]{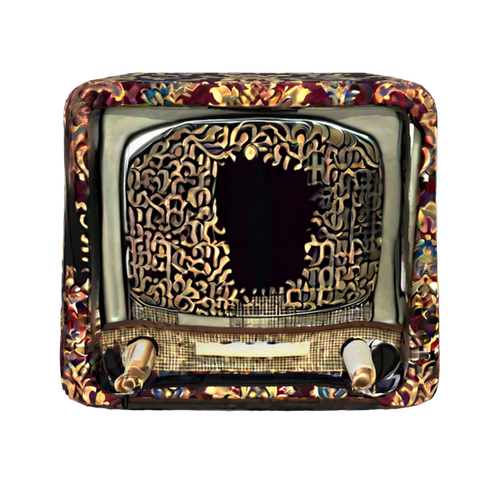}
& \includegraphics[trim=59.52pt 59.52pt 59.52pt 59.52pt, clip, width=0.12\columnwidth]{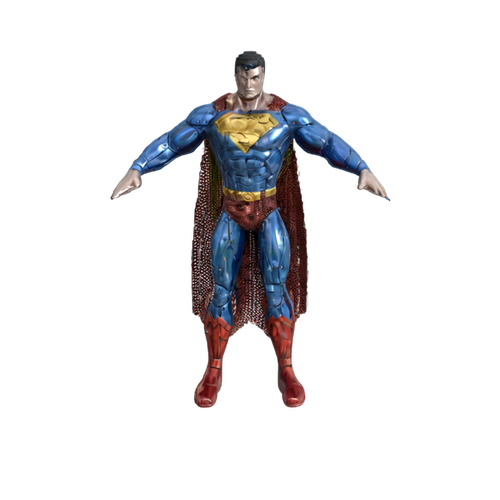}
& \includegraphics[trim=59.52pt 59.52pt 59.52pt 59.52pt, clip, width=0.12\columnwidth]{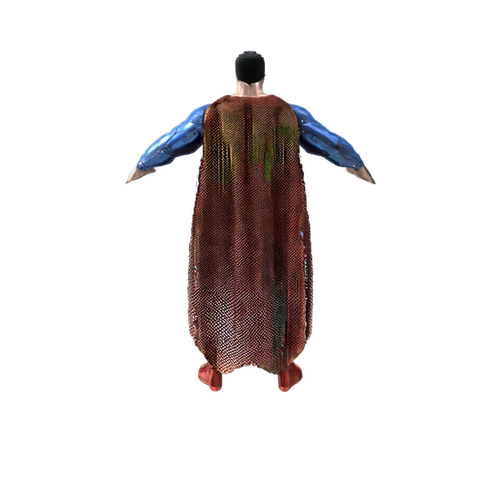}
\\
\parbox[c]{0.16\columnwidth}{\centering $s_{i}{=}5.5$\\$s_{mv}{=}3.5$}
& \includegraphics[trim=59.52pt 59.52pt 59.52pt 59.52pt, clip, width=0.12\columnwidth]{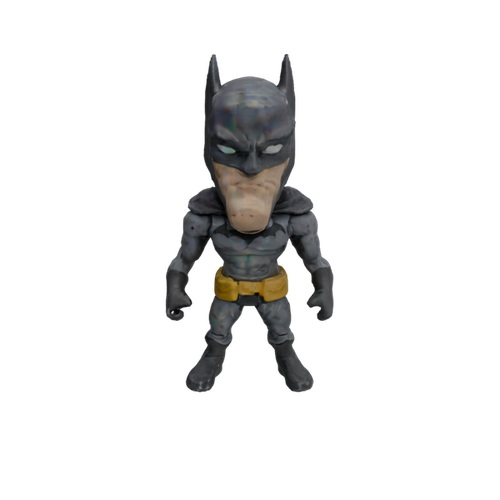}
& \includegraphics[trim=59.52pt 59.52pt 59.52pt 59.52pt, clip, width=0.12\columnwidth]{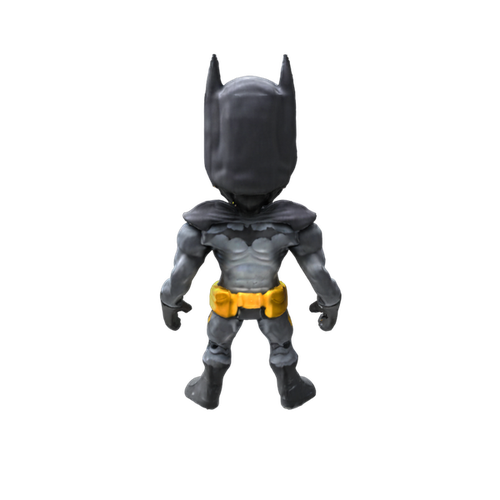}
& \includegraphics[trim=59.52pt 59.52pt 59.52pt 59.52pt, clip, width=0.12\columnwidth]{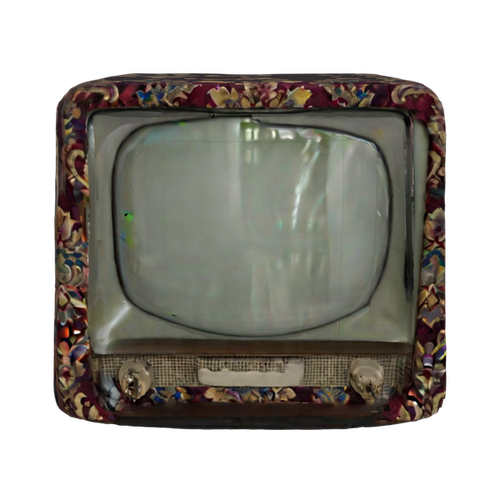}
& \includegraphics[trim=59.52pt 59.52pt 59.52pt 59.52pt, clip, width=0.12\columnwidth]{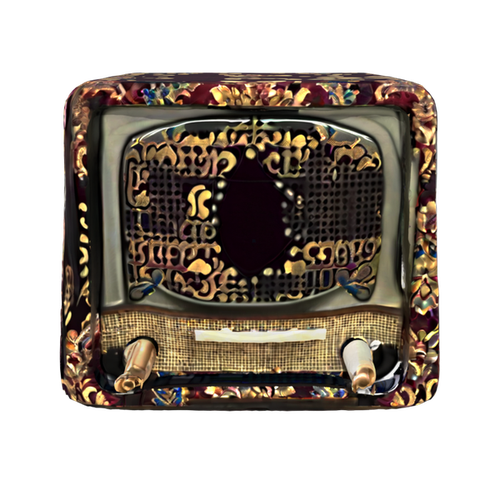}
& \includegraphics[trim=59.52pt 59.52pt 59.52pt 59.52pt, clip, width=0.12\columnwidth]{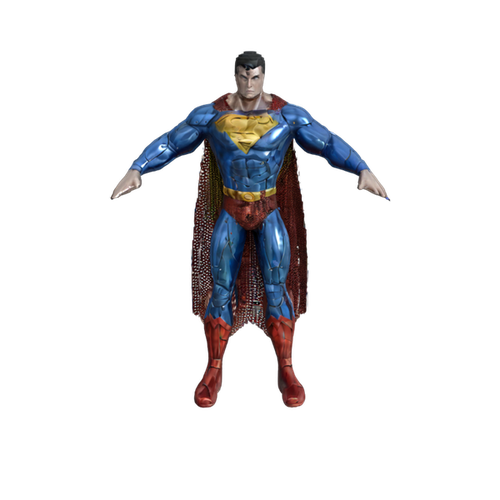}
& \includegraphics[trim=59.52pt 59.52pt 59.52pt 59.52pt, clip, width=0.12\columnwidth]{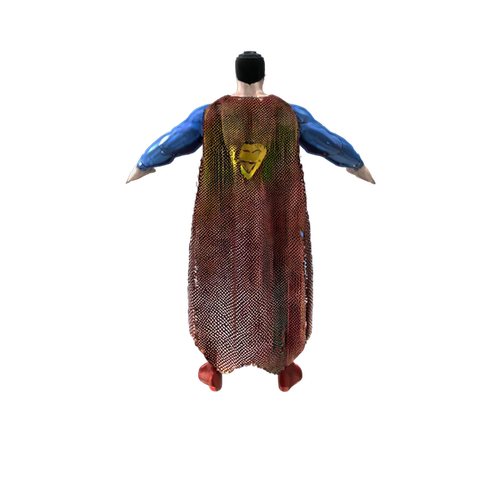}
\\
\parbox[c]{0.16\columnwidth}{\centering $s_{i}{=}3.5$\\$s_{mv}{=}5.5$}
& \includegraphics[trim=59.52pt 59.52pt 59.52pt 59.52pt, clip, width=0.12\columnwidth]{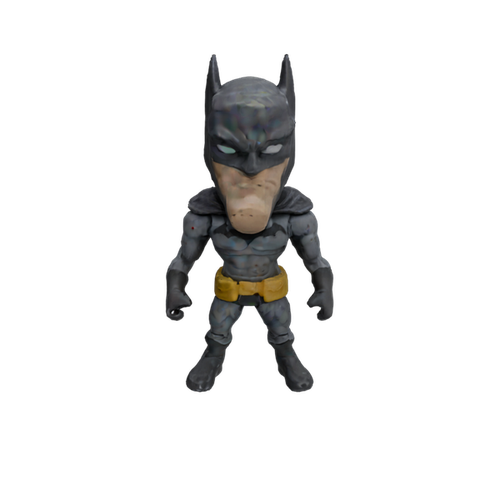}
& \includegraphics[trim=59.52pt 59.52pt 59.52pt 59.52pt, clip, width=0.12\columnwidth]{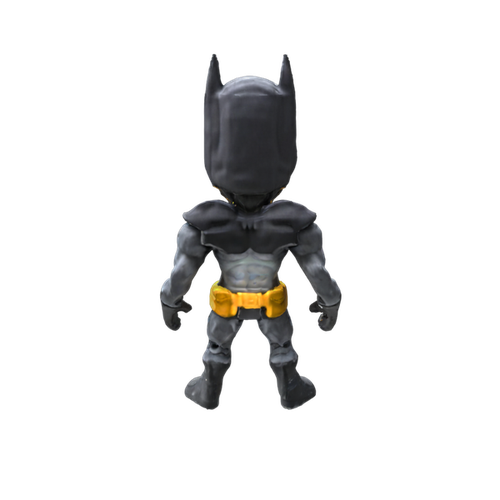}
& \includegraphics[trim=59.52pt 59.52pt 59.52pt 59.52pt, clip, width=0.12\columnwidth]{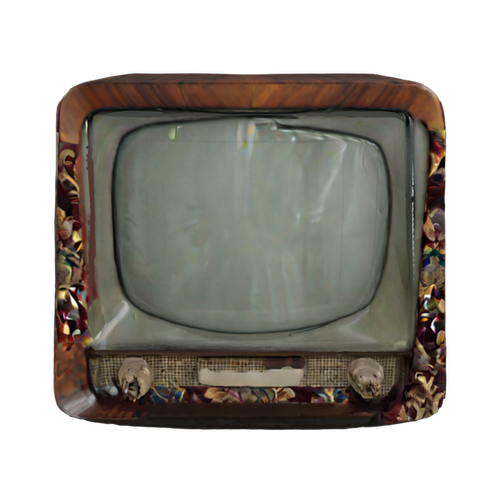}
& \includegraphics[trim=59.52pt 59.52pt 59.52pt 59.52pt, clip, width=0.12\columnwidth]{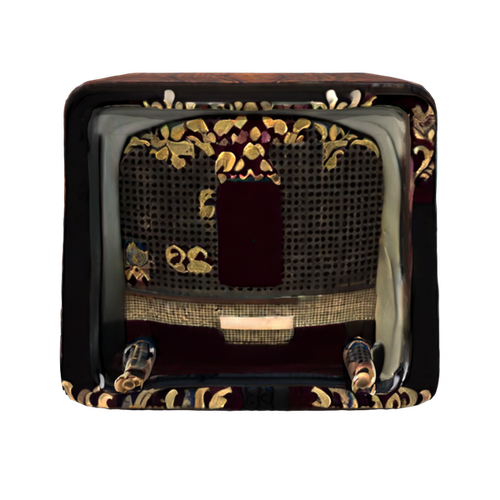}
& \includegraphics[trim=59.52pt 59.52pt 59.52pt 59.52pt, clip, width=0.12\columnwidth]{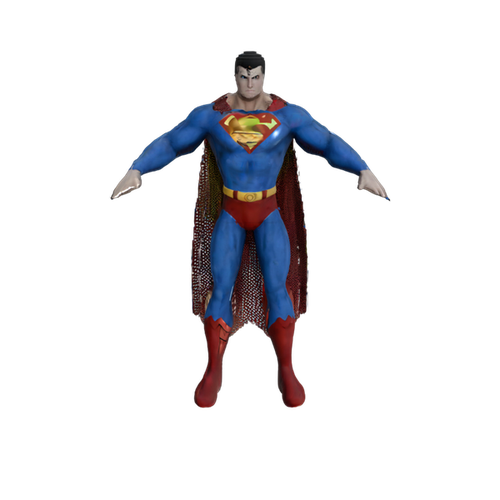}
& \includegraphics[trim=59.52pt 59.52pt 59.52pt 59.52pt, clip, width=0.12\columnwidth]{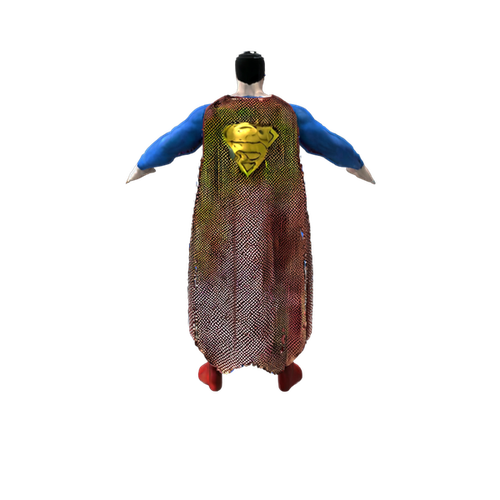}
\\
\parbox[c]{0.16\columnwidth}{\centering $s_{i}{=}2.5$\\$s_{mv}{=}6.5$}
& \includegraphics[trim=59.52pt 59.52pt 59.52pt 59.52pt, clip, width=0.12\columnwidth]{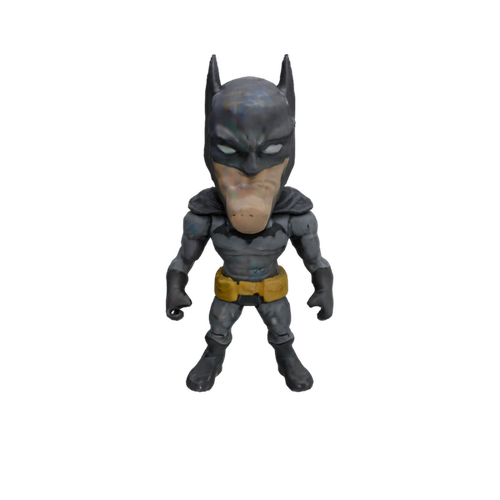}
& \includegraphics[trim=59.52pt 59.52pt 59.52pt 59.52pt, clip, width=0.12\columnwidth]{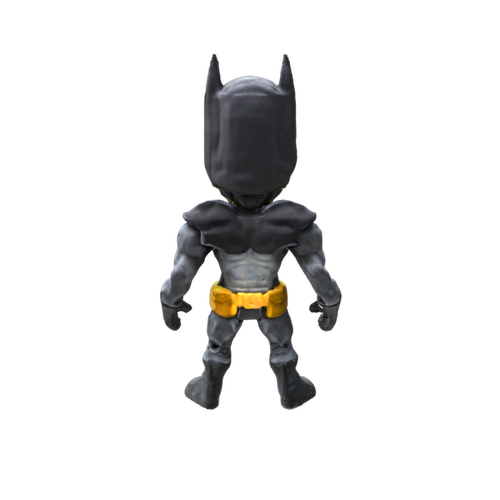}
& \includegraphics[trim=59.52pt 59.52pt 59.52pt 59.52pt, clip, width=0.12\columnwidth]{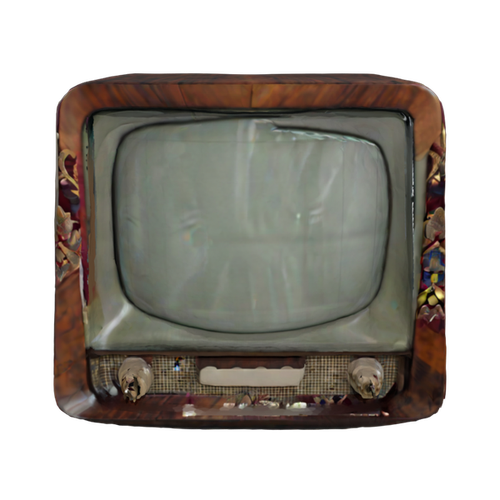}
& \includegraphics[trim=59.52pt 59.52pt 59.52pt 59.52pt, clip, width=0.12\columnwidth]{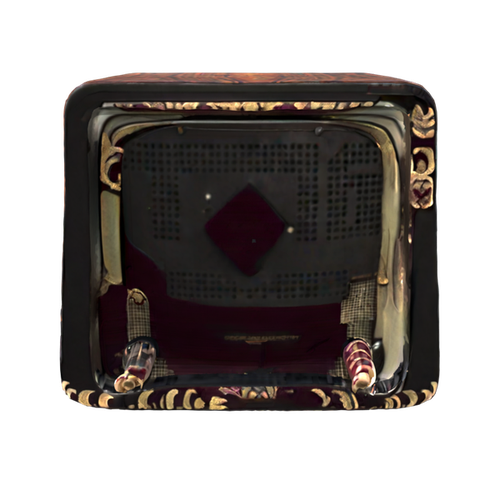}
& \includegraphics[trim=59.52pt 59.52pt 59.52pt 59.52pt, clip, width=0.12\columnwidth]{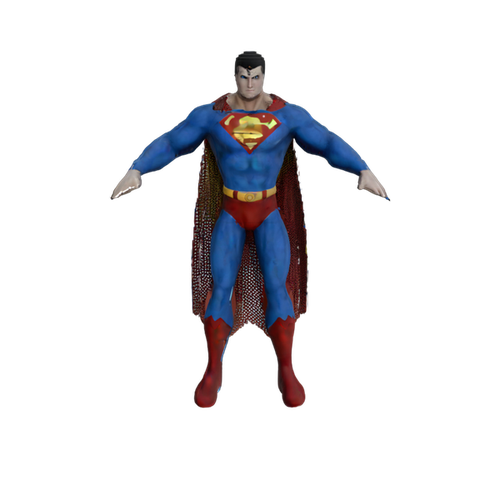}
& \includegraphics[trim=59.52pt 59.52pt 59.52pt 59.52pt, clip, width=0.12\columnwidth]{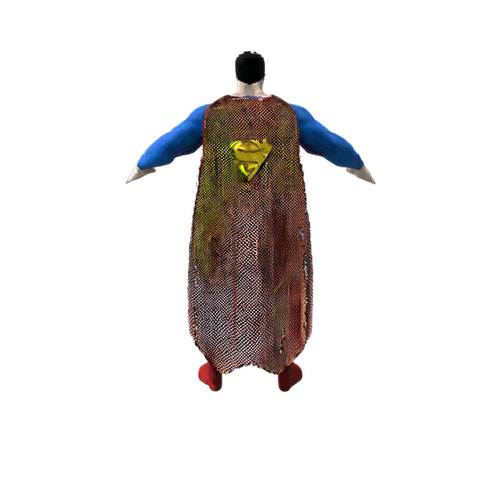}
\\
\bottomrule
\end{tabular*}

\caption{\textbf{Texture CFG tuning}. Editing results of \algoname Texture Editing Pipeline, under varying classifier-free guidance (CFG) scales for the edited reference image ($s_i$) and the source multi-view ($s_{mv}$) inputs. For clarity, we omit the superscript $T$. The results illustrate a tunable trade-off between edit-condition alignment and preservation of fine-grained details from the original texture, controlled by the CFG scales.}
\label{fig:cfg_ablation}
\end{figure}

We evaluated alternative approach for injecting the source multi-view condition. In this variant, the source multi-view images are concatenated with the UNet input noise along the channel dimension. To support this, the first convolution layer of the UNet is adapted to accept eight input channels instead of the original four. Quantitative results for this ablation are reported in \cref{tab:ablation_comparison_single}. While ConcatMV achieves the highest condition-alignment scores, it performs poorly in occluded-region fidelity. Our final approach provides the best trade-off between these two aspects.
Finally, we explore different combinations of classifier-free guidance (CFG) scales for the edited image and the source multi-view condition. Qualitative results in \cref{fig:cfg_ablation} demonstrate that these scales provide a controllable trade-off between adherence to the edit condition and preservation of the original texture.

\ifarxiv
\enlargethispage{60pt}
\subsection{Failure Cases Analysis}
\algoname exhibits degraded performance on assets with fine-grained geometric details or non-object-centric compositions.
Since all shapes are decoded through the pretrained Step1X-3D VAE, the output quality is bounded by its representational capacity; for assets with thin structures or intricate detail, the VAE encoder--decoder itself introduces smoothing and detail loss.
In the original image-to-3D setting the backbone can partially compensate by leveraging the input image as a strong appearance prior, re-synthesizing details that the latent code fails to preserve.
Our editing formulation, however, is conditioned on the source shape latent and trained to preserve the input geometry except where the edit condition specifies a change, making it more susceptible to representational drift in the latent space.
\cref{fig:failure_Cases} illustrates this with several representative examples: comparing each source shape with its VAE-only reconstruction confirms that much of the observed degradation is attributable to the limited expressiveness of the shape latent space rather than to the editing process itself.
\ifarxiv
\begin{figure*}[!htbp]
\else
\begin{figure*}[t]
\fi
  \centering
  \setlength{\tabcolsep}{1pt}
  \renewcommand{\arraystretch}{0.6}

  \newlength{\fcRowHeight}
  \setlength{\fcRowHeight}{0.115\linewidth}

  \setlength{\arrayrulewidth}{0.6pt}%
  \ifarxiv
    \def\fcBoxWidth{0.92\linewidth}%
  \else
    \def\fcBoxWidth{\linewidth}%
  \fi
  \resizebox{\fcBoxWidth}{!}{%
  \begin{tabular}[t]{@{}cc@{}}
    \hline
    \multicolumn{2}{c}{\scriptsize \textbf{Input Shape}} \\
    \hline
    \includegraphics[height=\fcRowHeight, trim={17.86 17.86 17.86 17.86}, clip]{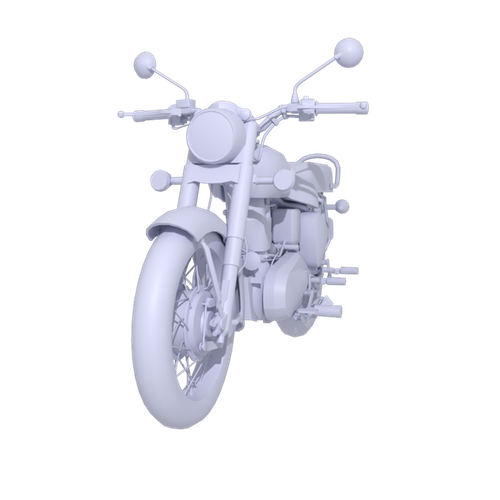} & \includegraphics[height=\fcRowHeight, trim={17.86 17.86 17.86 17.86}, clip]{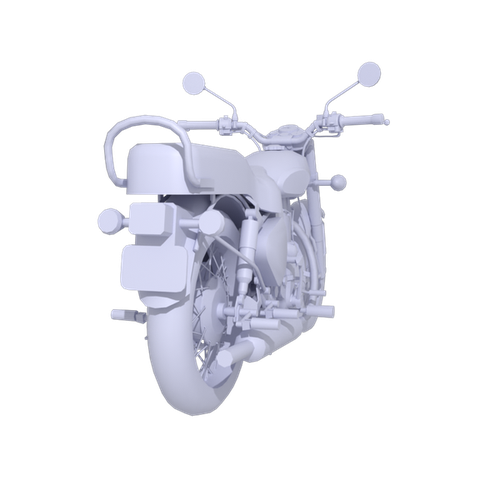} \\
    \includegraphics[height=\fcRowHeight, trim={17.86 17.86 17.86 17.86}, clip]{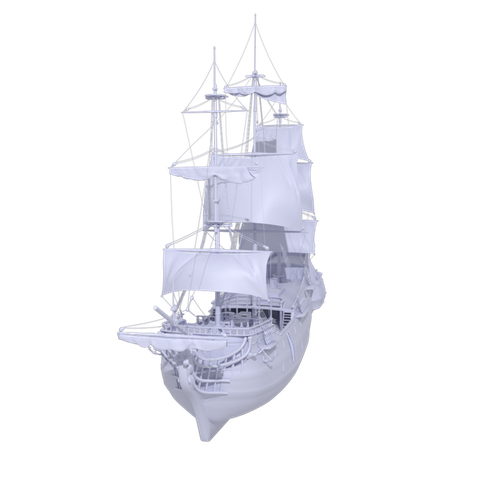} & \includegraphics[height=\fcRowHeight, trim={17.86 17.86 17.86 17.86}, clip]{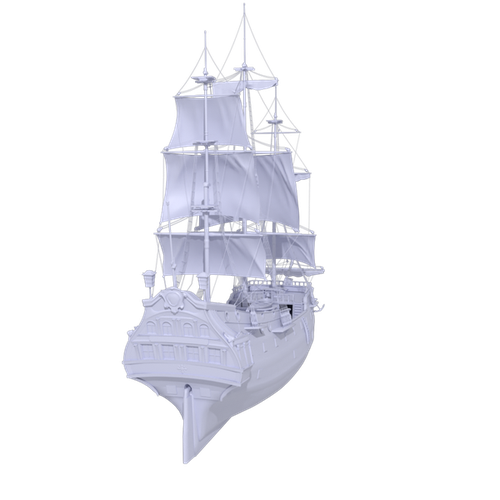} \\
    \includegraphics[height=\fcRowHeight, trim={17.86 17.86 17.86 17.86}, clip]{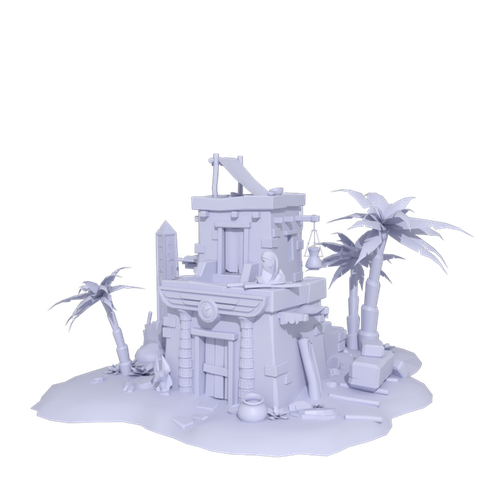} & \includegraphics[height=\fcRowHeight, trim={17.86 17.86 17.86 17.86}, clip]{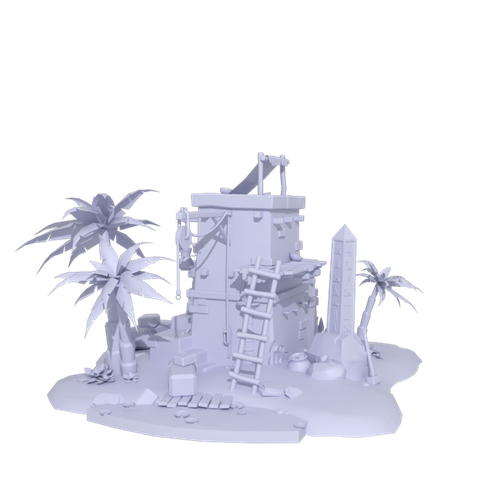} \\
    \hline
  \end{tabular}%
  \vrule width 0.6pt%
  \begin{tabular}[t]{@{}cc@{}}
    \hline
    \multicolumn{2}{c}{\scriptsize \textbf{VAE Output}} \\
    \hline
    \includegraphics[height=\fcRowHeight, trim={17.86 17.86 17.86 17.86}, clip]{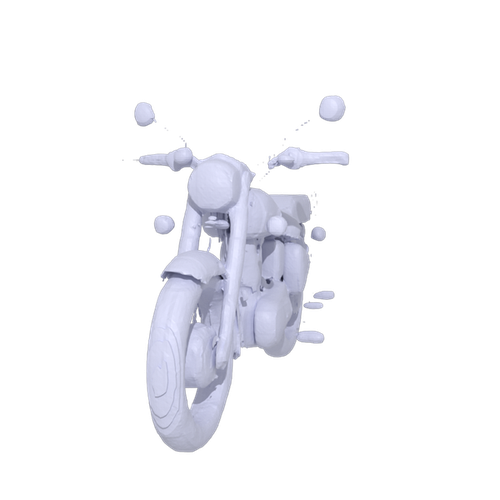} & \includegraphics[height=\fcRowHeight, trim={17.86 17.86 17.86 17.86}, clip]{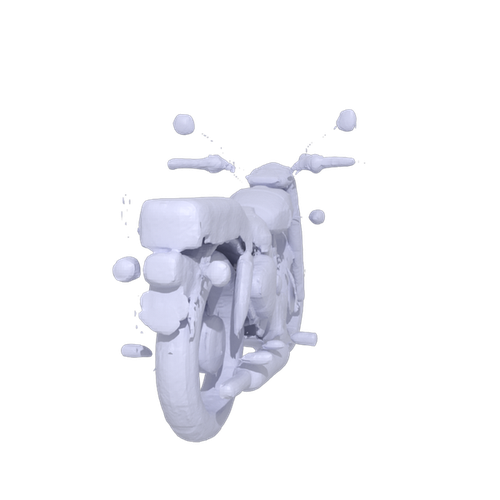} \\
    \includegraphics[height=\fcRowHeight, trim={17.86 17.86 17.86 17.86}, clip]{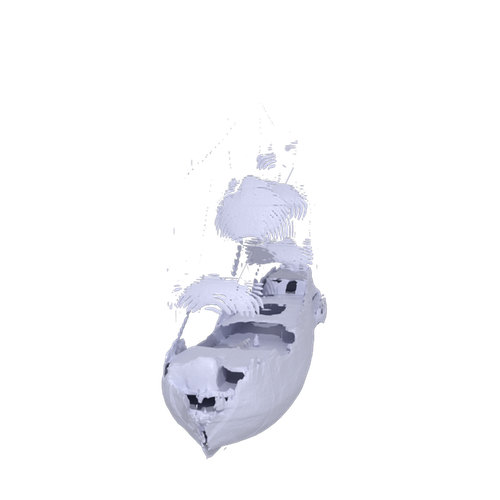} & \includegraphics[height=\fcRowHeight, trim={17.86 17.86 17.86 17.86}, clip]{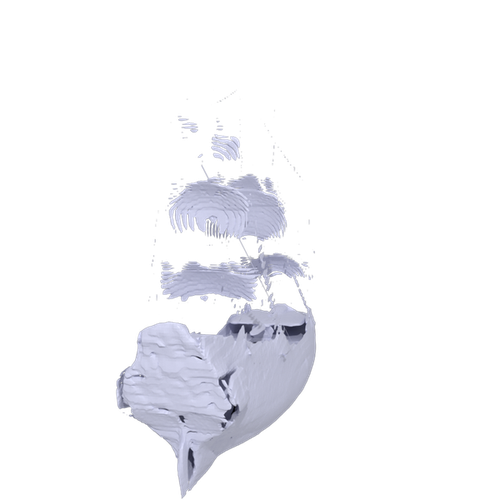} \\
    \includegraphics[height=\fcRowHeight, trim={17.86 17.86 17.86 17.86}, clip]{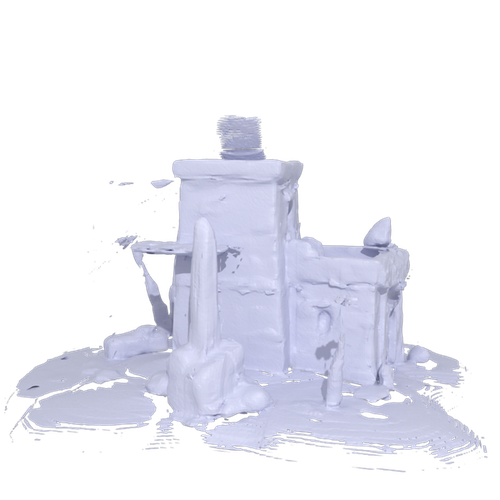} & \includegraphics[height=\fcRowHeight, trim={17.86 17.86 17.86 17.86}, clip]{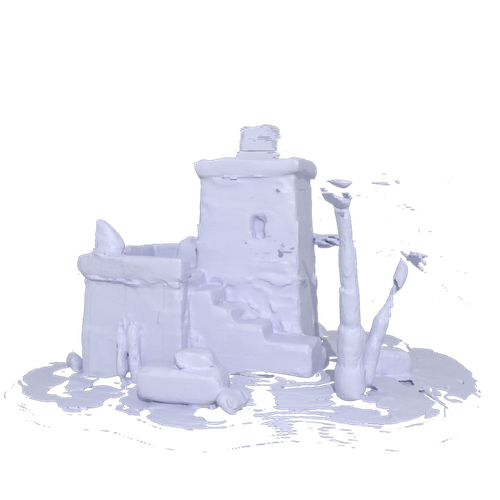} \\
    \hline
  \end{tabular}%
  \vrule width 0.6pt%
  \begin{tabular}[t]{@{}c@{}}
    \hline
    \scriptsize \textbf{Edit Condition} \\
    \hline
    \includegraphics[height=\fcRowHeight, trim={34.18 34.18 34.18 34.18}, clip]{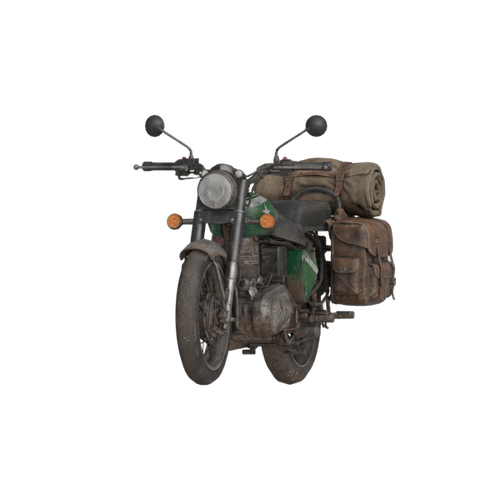} \\
    \includegraphics[height=\fcRowHeight, trim={34.18 34.18 34.18 34.18}, clip]{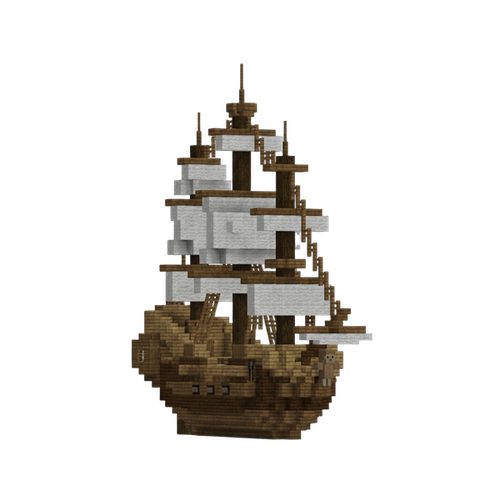} \\
    \includegraphics[height=\fcRowHeight]{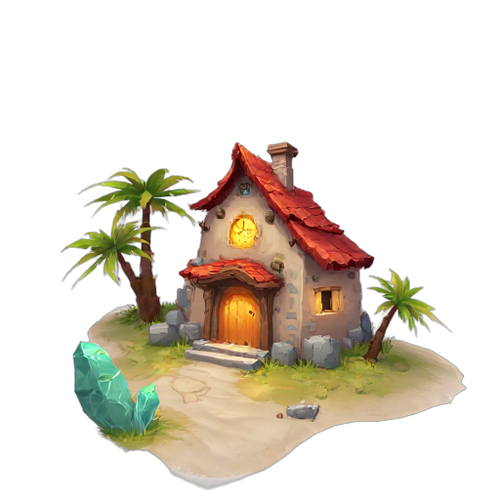} \\
    \hline
  \end{tabular}%
  \vrule width 0.6pt%
  \begin{tabular}[t]{@{}cc@{}}
    \hline
    \multicolumn{2}{c}{\scriptsize \textbf{Output Shape}} \\
    \hline
    \includegraphics[height=\fcRowHeight, trim={17.86 17.86 17.86 17.86}, clip]{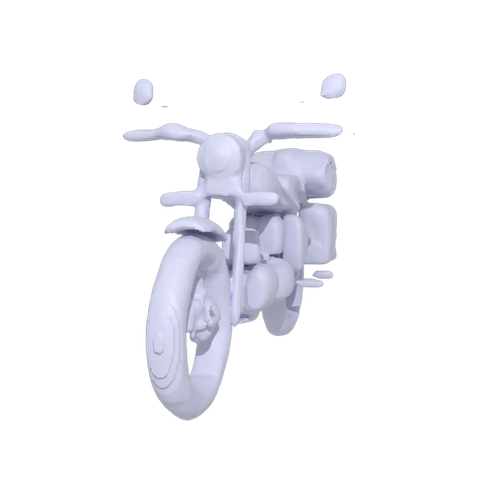} & \includegraphics[height=\fcRowHeight, trim={17.86 17.86 17.86 17.86}, clip]{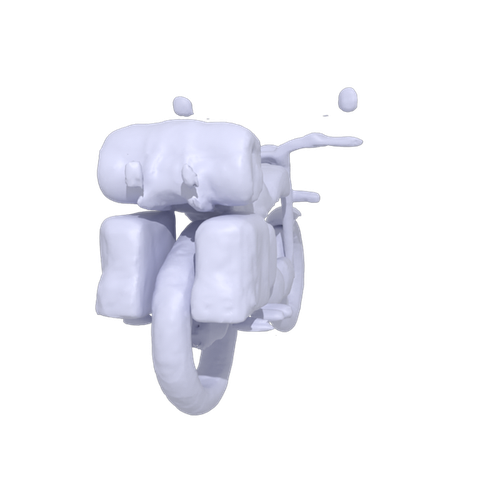} \\
    \includegraphics[height=\fcRowHeight, trim={17.86 17.86 17.86 17.86}, clip]{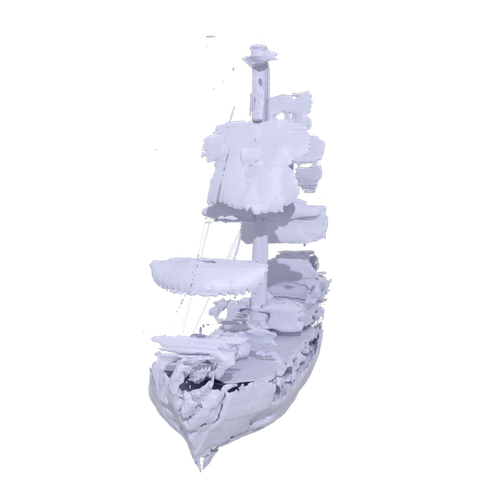} & \includegraphics[height=\fcRowHeight, trim={17.86 17.86 17.86 17.86}, clip]{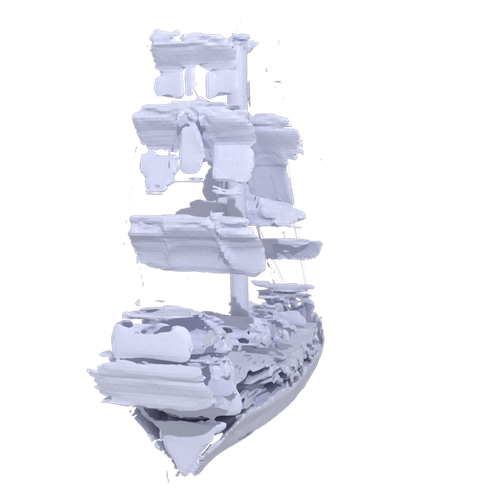} \\
    \includegraphics[height=\fcRowHeight, trim={17.86 17.86 17.86 17.86}, clip]{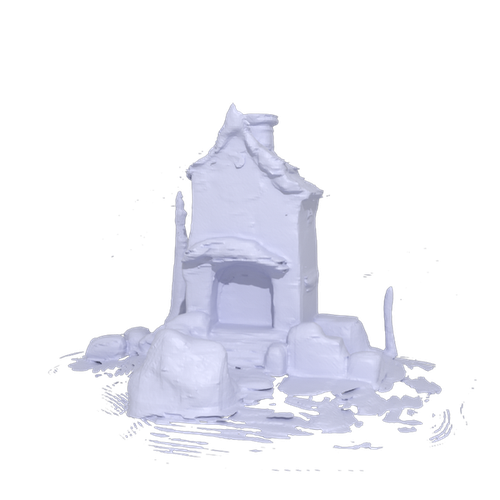} & \includegraphics[height=\fcRowHeight, trim={17.86 17.86 17.86 17.86}, clip]{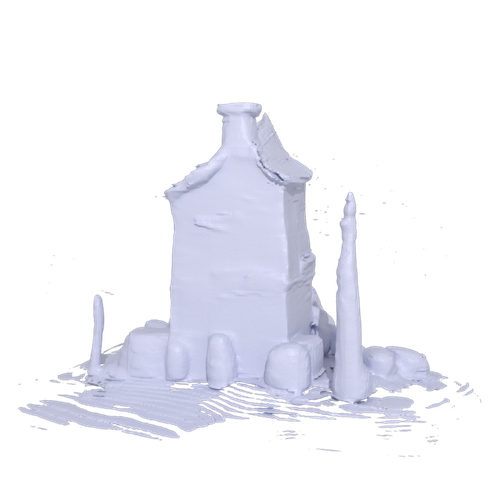} \\
    \hline
  \end{tabular}%
  }%

  \captionof{figure}{\textbf{Failure Cases Analysis.}Columns 1--2 show the \textbf{Input Shape} (front/back), 3--4 show the \textbf{VAE Output} (front/back). Column 5 is the \textbf{Edit Condition} and 6--7 present \textbf{Output Shape} (front/back).}
  \label{fig:failure_Cases}
\end{figure*}

\FloatBarrier
\fi

\section{Conclusion, Limitations and Future Work}
\label{sec:conclusions}
We have presented ShapeUP, a feed-forward approach to 3D editing built on a simple but deliberate design choice. Rather than introducing specialized editing machinery, we formulate 3D editing as a supervised latent-to-latent translation problem within a native 3D foundation model, jointly conditioned on the source shape and an edited image. This perspective anchors generation to the identity of the original asset while learning how visual edits in 2D translate into coherent changes in 3D, enabling both local and global edits without optimization, explicit masks, or multi-view reconstruction.

Our formulation naturally leads to a mask-free, image-conditioned framework for 3D editing, but it also shifts much of the burden toward the construction of suitable supervision data. In our work, we introduced a novel use of 3D motion data, leveraging temporally distant frames to generate supervision that goes beyond local part edits and meaningfully expands the scope of editable transformations, particularly for global pose and deformation changes. At the same time, our current training set remains relatively small and is biased toward mostly closed, object-centric assets. Addressing these cases through larger and more diverse training data
\ifarxiv
\else
, as well as a deeper study of guidance trade-offs between edit fidelity and identity preservation, 
\fi represents a promising direction for future work. We hope \algoname encourages further exploration of learnable, native-3D editing frameworks built on foundation models as a practical path toward controllable 3D content creation.

\ifanonymous\else
\begin{acks}
  We thank Roi Bar-On for his early feedback and helpful suggestions. We also thank the anonymous reviewers for their valuable feedback.
  This research was supported in part by the Israel Science Foundation (grants no. 2492/20 and 1473/24), Len Blavatnik and the Blavatnik family foundation. 
\end{acks}
\fi

\ifarxiv
    \begin{figure*}[t]
  \centering
  \setlength{\tabcolsep}{1pt} 
  \renewcommand{\arraystretch}{0.6}

\resizebox{0.99\linewidth}{!}{
\begin{tabular}{@{}cc!{\vrule width 0.6pt}ccc!{\vrule width 0.6pt}ccc@{}}
    \toprule
    \multicolumn{2}{c}{\scriptsize \textbf{Source Textured Mesh}} &
    \multicolumn{1}{c}{\scriptsize \textbf{Edit Condition}} &
    \multicolumn{2}{c}{\scriptsize \textbf{Edited}} &
    \multicolumn{1}{c}{\scriptsize \textbf{Edit Condition}} &
    \multicolumn{2}{c}{\scriptsize \textbf{Edited}} \\
    
    \cmidrule(r){1-2} \cmidrule(lr){3-3} \cmidrule(lr){4-5} \cmidrule(lr){6-6} \cmidrule(l){7-8}

    \\[-14pt]
    \includegraphics[width=0.12\linewidth, trim={17.86 17.86 17.86 17.86}, clip]{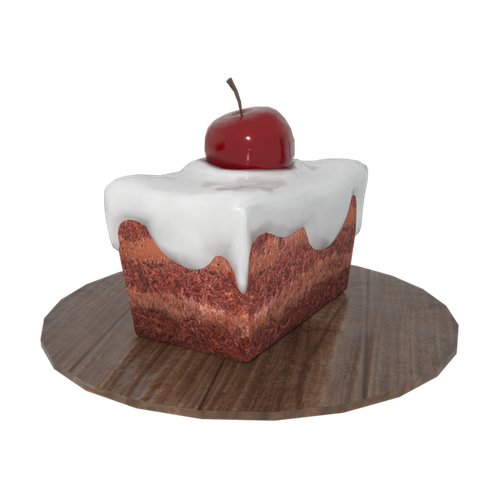} & \includegraphics[width=0.12\linewidth, trim={17.86 17.86 17.86 17.86}, clip]{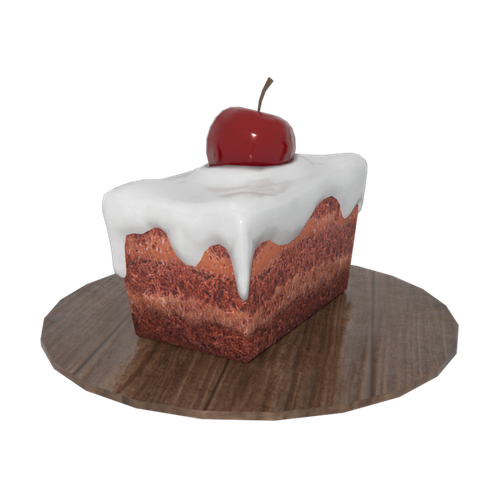} &
    \includegraphics[width=0.13\linewidth, trim={34.18 34.18 34.18 34.18}, clip]{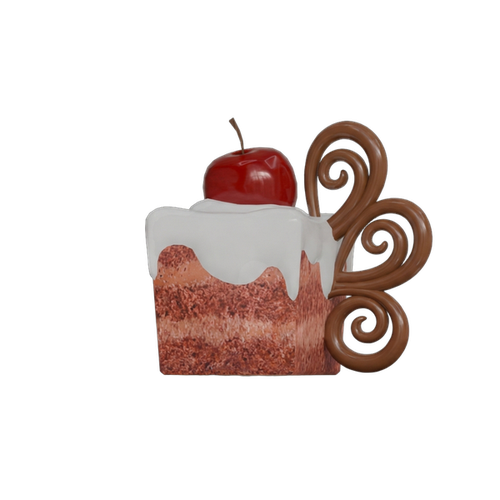} &
    \includegraphics[width=0.12\linewidth, trim={17.86 17.86 17.86 17.86}, clip]{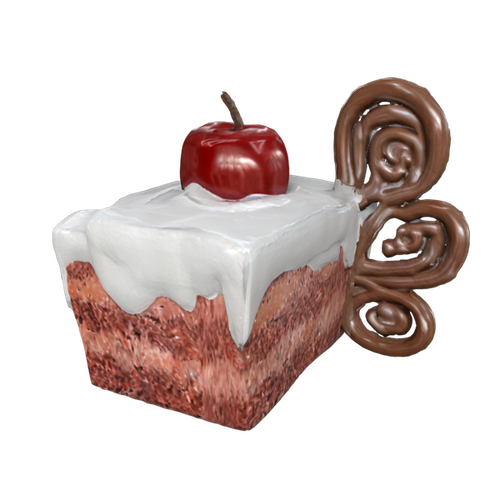} & \includegraphics[width=0.12\linewidth, trim={17.86 17.86 17.86 17.86}, clip]{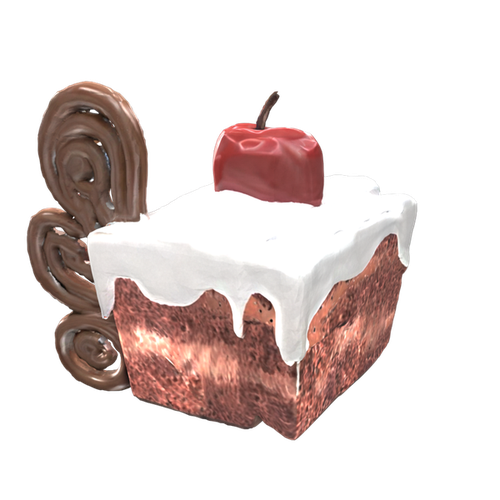} &
    \includegraphics[width=0.13\linewidth, trim={34.18 34.18 34.18 34.18}, clip]{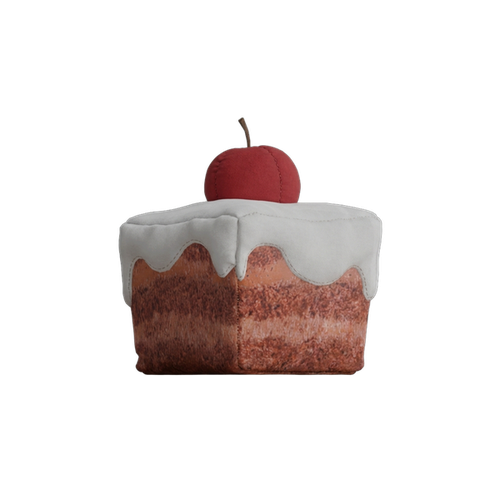} &
    \raisebox{3pt}{\includegraphics[width=0.115\linewidth, trim={0 0 0 0}, clip]{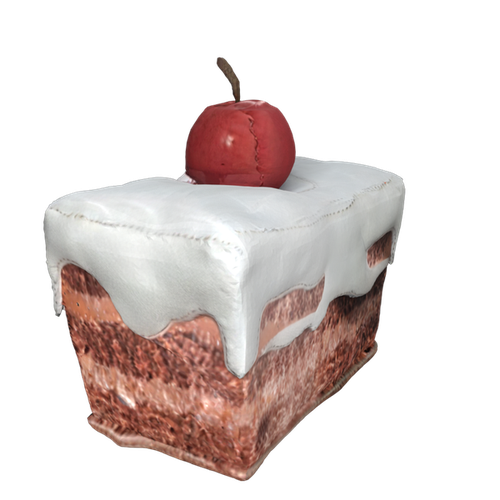}} & \raisebox{3pt}{\includegraphics[width=0.115\linewidth, trim={0 0 0 0}, clip]{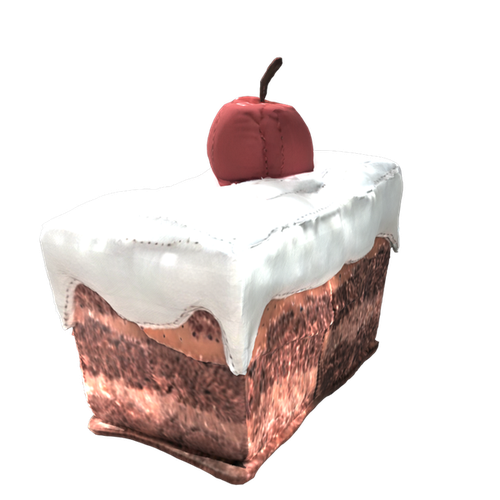}} \\ [-7pt]

    \includegraphics[width=0.12\linewidth, trim={17.86 17.86 17.86 17.86}, clip]{images/qualitative_results/Source_meshes/german_shep_60deg.png} & \includegraphics[width=0.12\linewidth, trim={17.86 17.86 17.86 17.86}, clip]{images/qualitative_results/Source_meshes/german_shep_240deg.png} &
    \includegraphics[width=0.12\linewidth, trim={39.06 39.06 39.06 39.06}, clip]{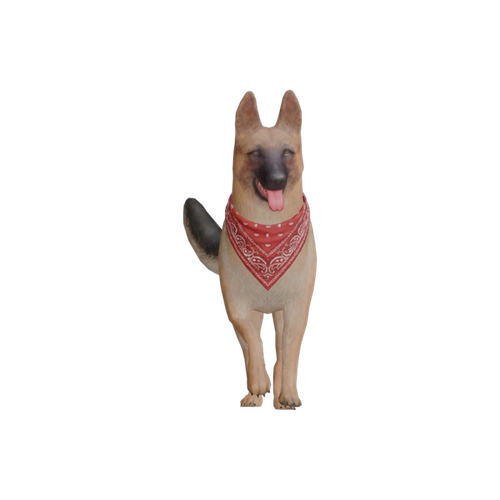} &
    \includegraphics[width=0.12\linewidth, trim={47.62 47.62 47.62 47.62}, clip]{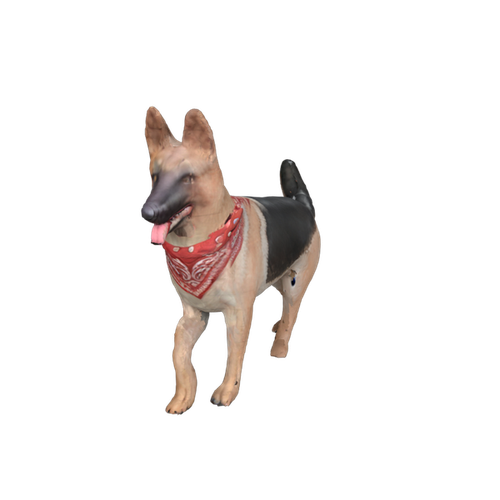} & \includegraphics[width=0.12\linewidth, trim={59.52 59.52 59.52 59.52}, clip]{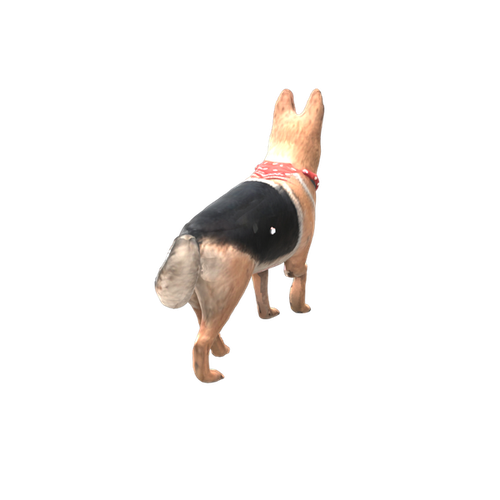} &
    \includegraphics[width=0.12\linewidth, trim={39.06 39.06 39.06 39.06}, clip]{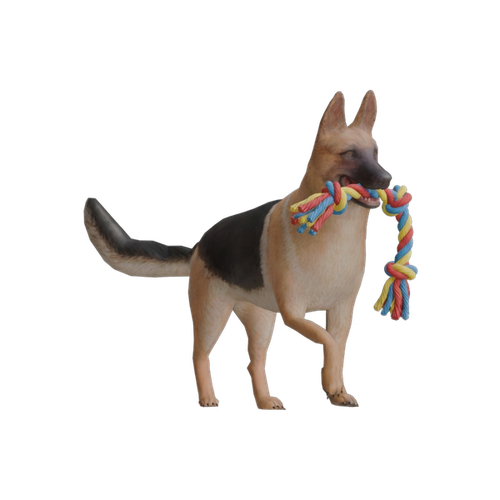} &
    \includegraphics[width=0.12\linewidth, trim={47.62 47.62 47.62 47.62}, clip]{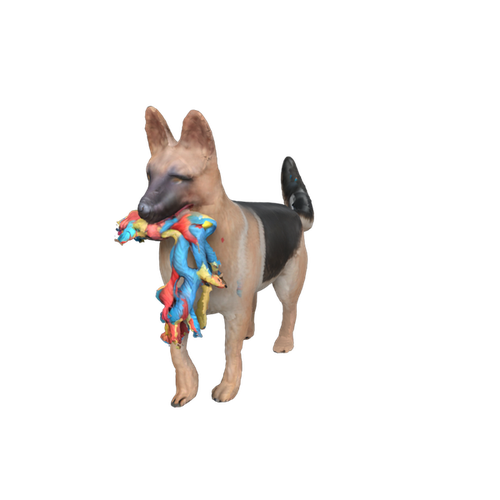} & \includegraphics[width=0.12\linewidth, trim={59.52 59.52 59.52 59.52}, clip]{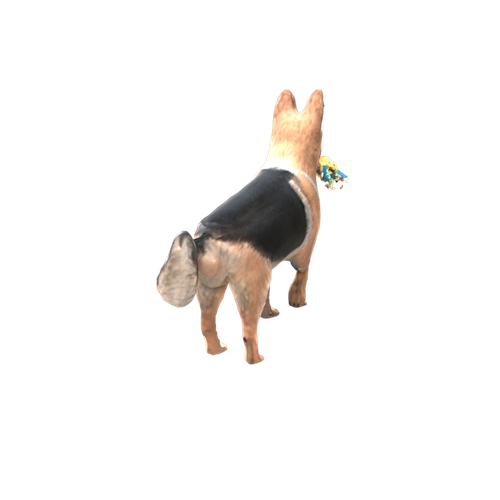} \\ [-8pt]

    \includegraphics[width=0.12\linewidth, trim={17.86 17.86 17.86 17.86}, clip]{images/qualitative_results/Source_meshes/grogu_60deg.png} & \includegraphics[width=0.12\linewidth, trim={17.86 17.86 17.86 17.86}, clip]{images/qualitative_results/Source_meshes/grogu_240deg.png} &
    \raisebox{3pt}{\includegraphics[width=0.115\linewidth, trim={0 0 0 0}, clip]{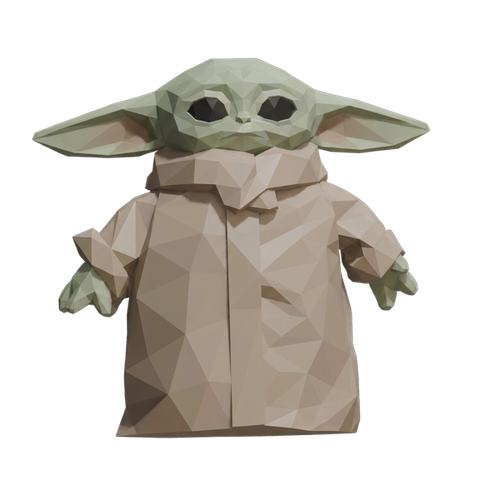}} &
    \includegraphics[width=0.12\linewidth, trim={17.86 17.86 17.86 17.86}, clip]{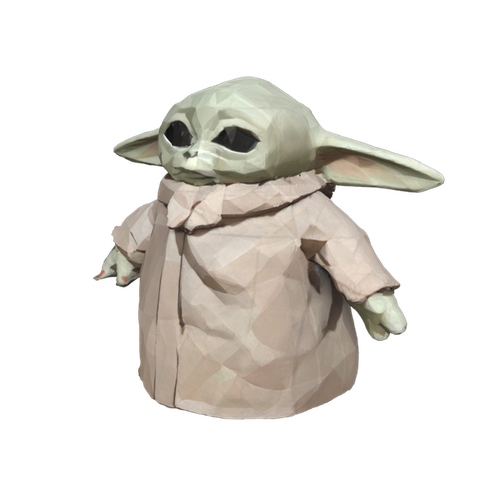} & \includegraphics[width=0.12\linewidth, trim={17.86 17.86 17.86 17.86}, clip]{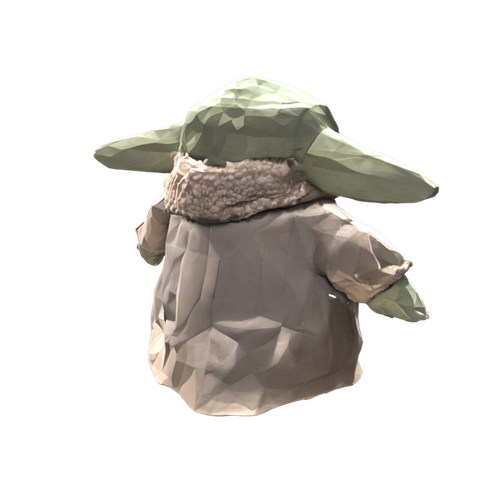} &
    \raisebox{3pt}{\includegraphics[width=0.115\linewidth, trim={0 0 0 0}, clip]{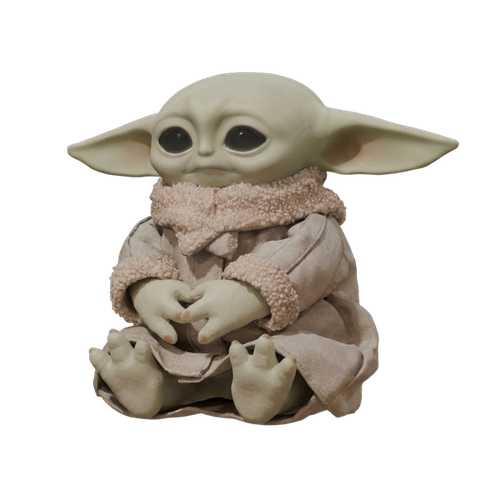}} &
    \includegraphics[width=0.12\linewidth, trim={17.86 17.86 17.86 17.86}, clip]{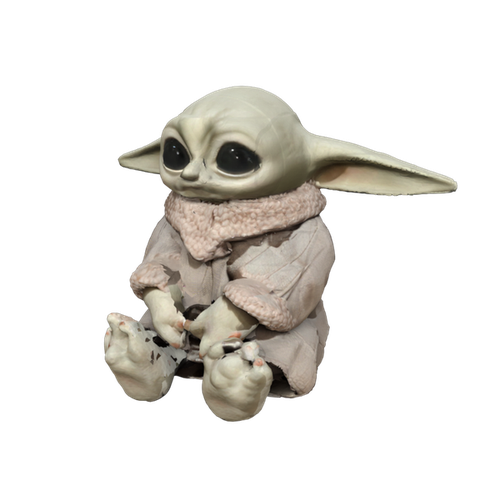} & \includegraphics[width=0.12\linewidth, trim={17.86 17.86 17.86 17.86}, clip]{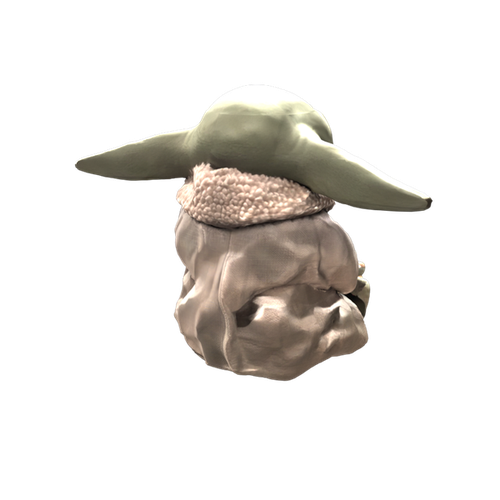} \\ [-5pt]

    \includegraphics[width=0.12\linewidth, trim={17.86 17.86 17.86 17.86}, clip]{images/qualitative_results/Source_meshes/robot_60deg.png} & \includegraphics[width=0.12\linewidth, trim={17.86 17.86 17.86 17.86}, clip]{images/qualitative_results/Source_meshes/robot_240deg.png} &
    \includegraphics[width=0.12\linewidth, trim={14.65 14.65 14.65 14.65}, clip]{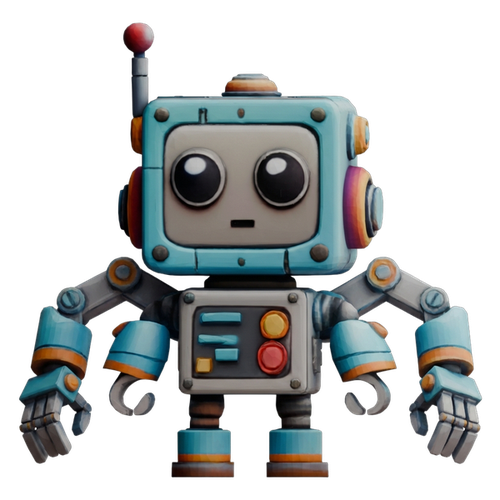} &
    \includegraphics[width=0.12\linewidth, trim={17.86 17.86 17.86 17.86}, clip]{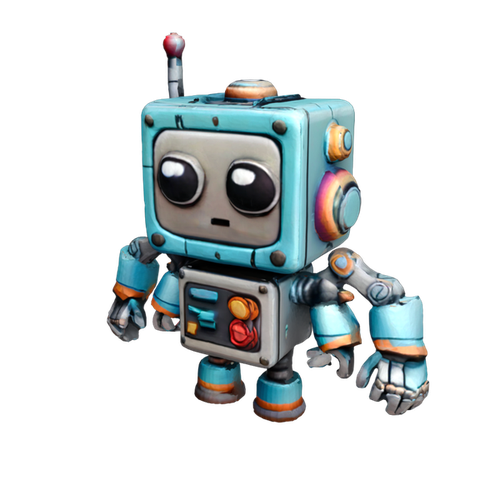} & \includegraphics[width=0.12\linewidth, trim={17.86 17.86 17.86 17.86}, clip]{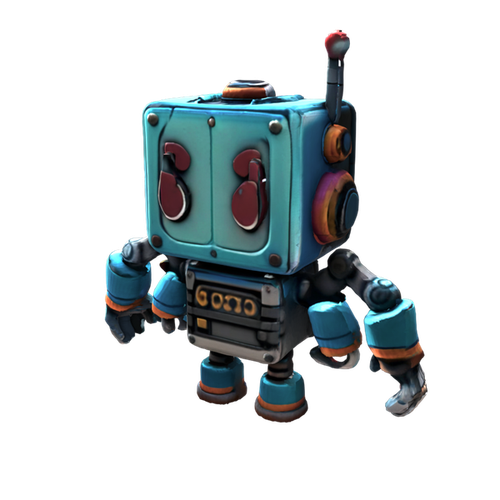} &
    \includegraphics[width=0.12\linewidth, trim={14.65 14.65 14.65 14.65}, clip]{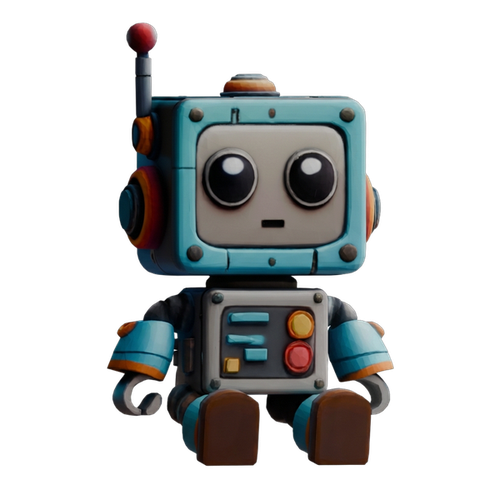} &
    \includegraphics[width=0.12\linewidth, trim={17.86 17.86 17.86 17.86}, clip]{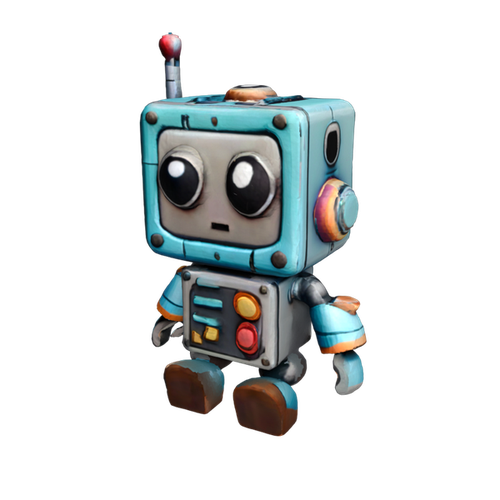} & \includegraphics[width=0.12\linewidth, trim={17.86 17.86 17.86 17.86}, clip]{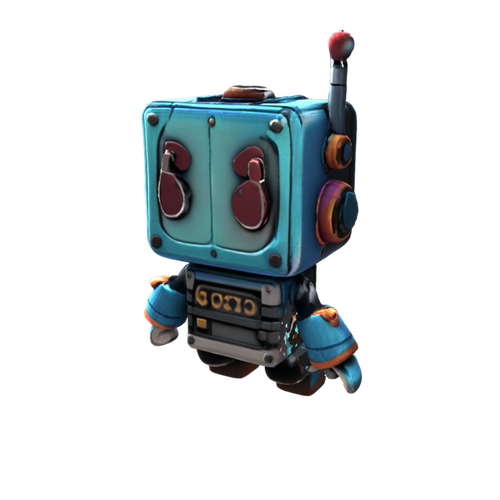}  \\

    \includegraphics[width=0.12\linewidth, trim={17.86 17.86 17.86 17.86}, clip]{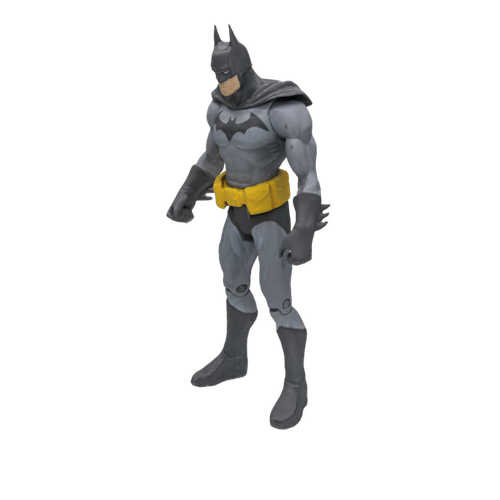} & \includegraphics[width=0.12\linewidth, trim={17.86 17.86 17.86 17.86}, clip]{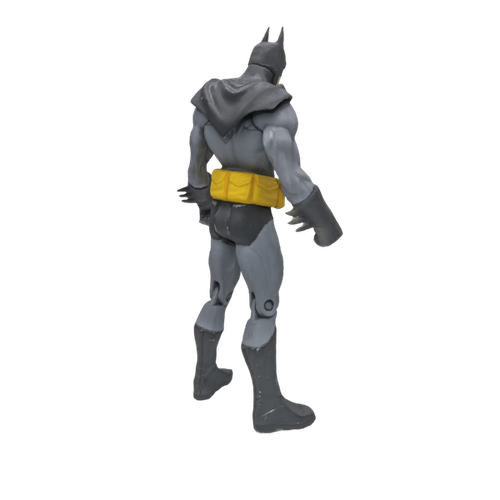} &
    \raisebox{3pt}{\includegraphics[width=0.115\linewidth, trim={0 0 0 0}, clip]{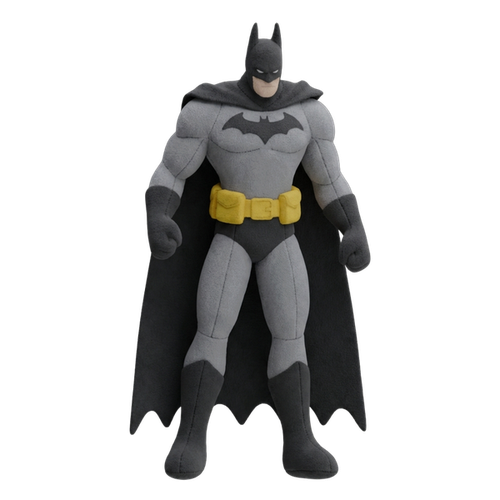}} &
    \includegraphics[width=0.12\linewidth, trim={17.86 17.86 17.86 17.86}, clip]{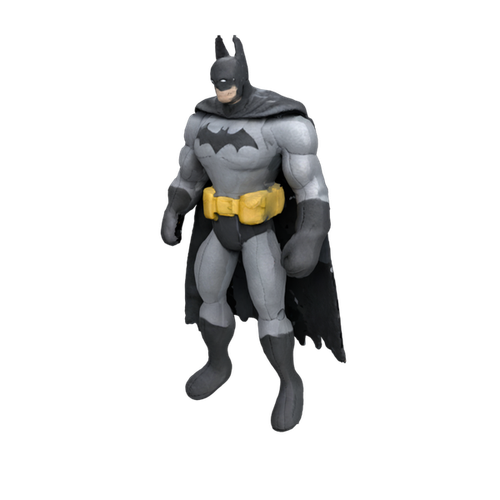} & \includegraphics[width=0.12\linewidth, trim={17.86 17.86 17.86 17.86}, clip]{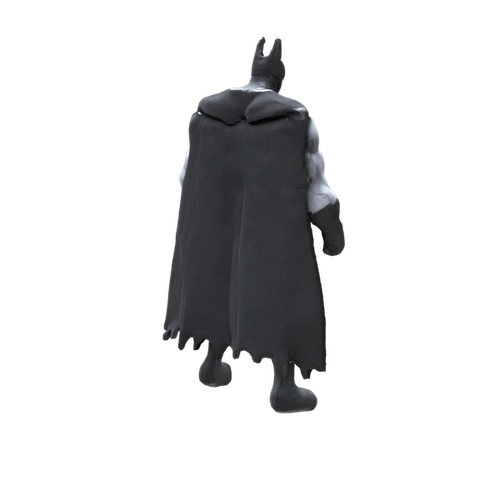} &
    \raisebox{3pt}{\includegraphics[width=0.115\linewidth, trim={0 0 0 0}, clip]{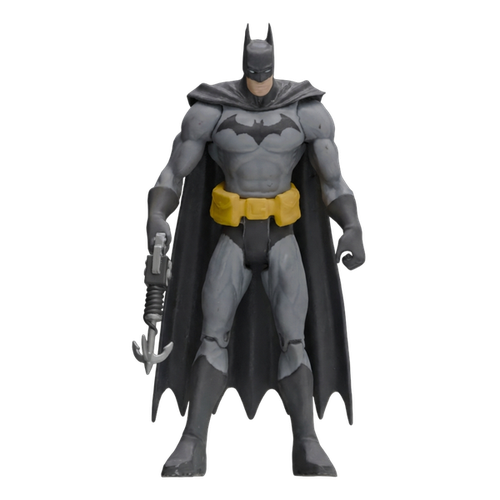}} &
    \includegraphics[width=0.12\linewidth, trim={17.86 17.86 17.86 17.86}, clip]{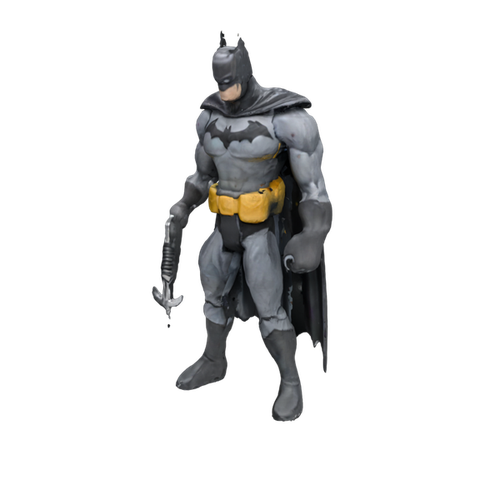} & \includegraphics[width=0.12\linewidth, trim={17.86 17.86 17.86 17.86}, clip]{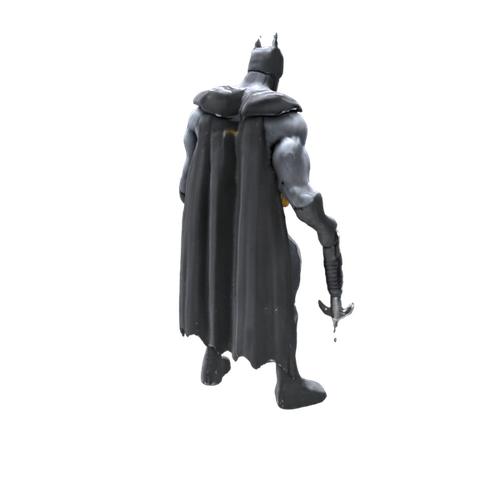} \\ [-3pt]

    \includegraphics[width=0.12\linewidth, trim={17.86 17.86 17.86 17.86}, clip]{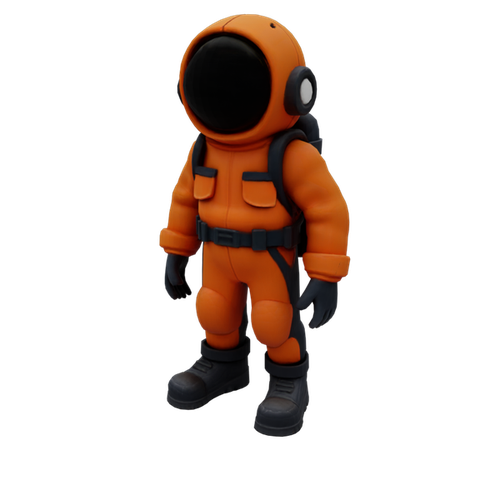} & \includegraphics[width=0.12\linewidth, trim={17.86 17.86 17.86 17.86}, clip]{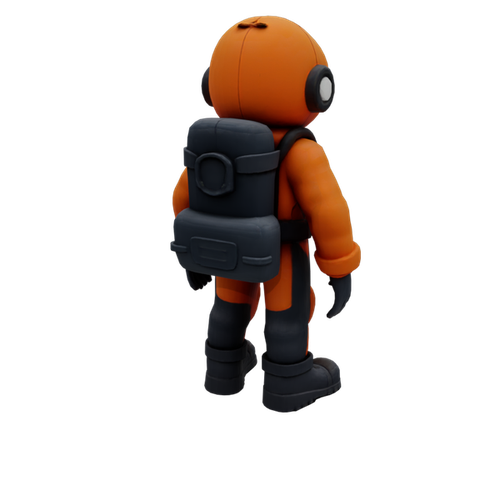} &
    \includegraphics[width=0.12\linewidth, trim={14.65 14.65 14.65 14.65}, clip]{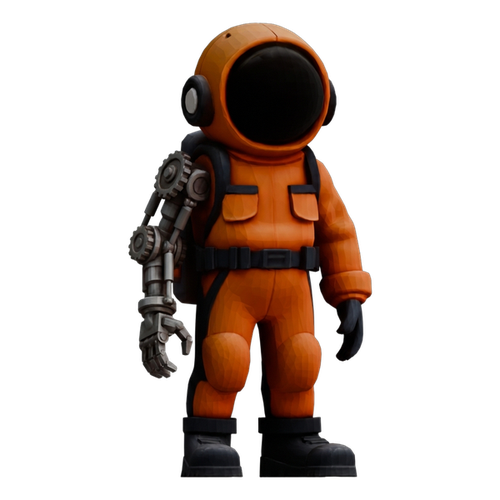} &
    \includegraphics[width=0.12\linewidth, trim={17.86 17.86 17.86 17.86}, clip]{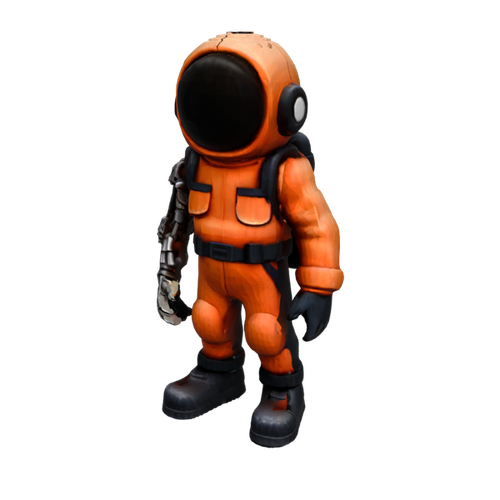} & \includegraphics[width=0.12\linewidth, trim={17.86 17.86 17.86 17.86}, clip]{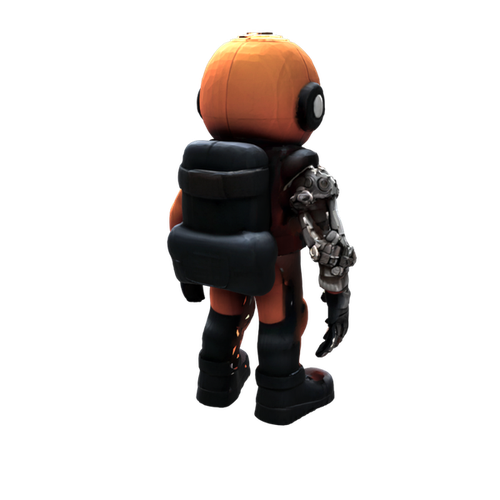} &
    \includegraphics[width=0.12\linewidth, trim={14.65 14.65 14.65 14.65}, clip]{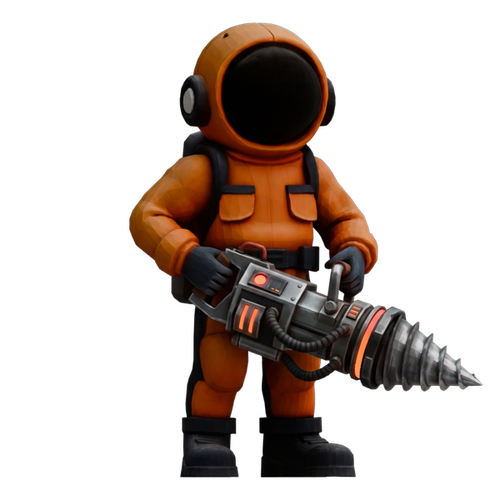} &
    \includegraphics[width=0.12\linewidth, trim={17.86 17.86 17.86 17.86}, clip]{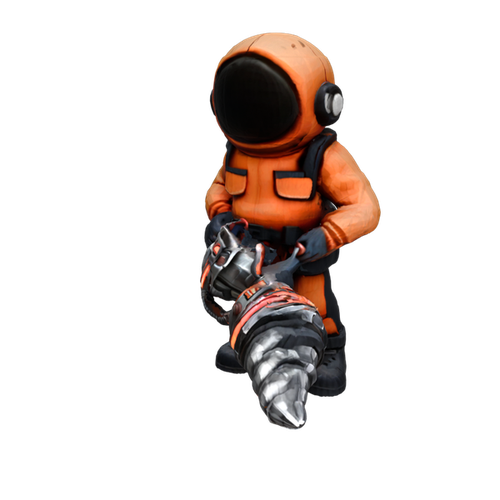} & \includegraphics[width=0.12\linewidth, trim={17.86 17.86 17.86 17.86}, clip]{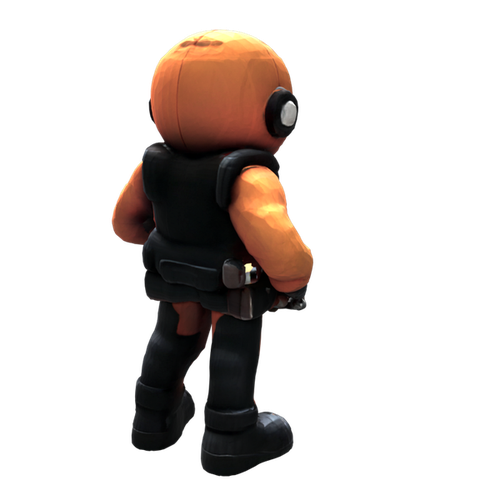} \\ [-3pt]

    \includegraphics[width=0.12\linewidth, trim={47.62 47.62 47.62 47.62}, clip]{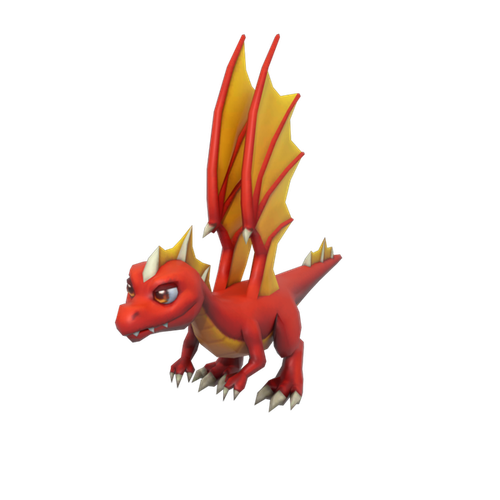} & \includegraphics[width=0.12\linewidth, trim={47.62 47.62 47.62 47.62}, clip]{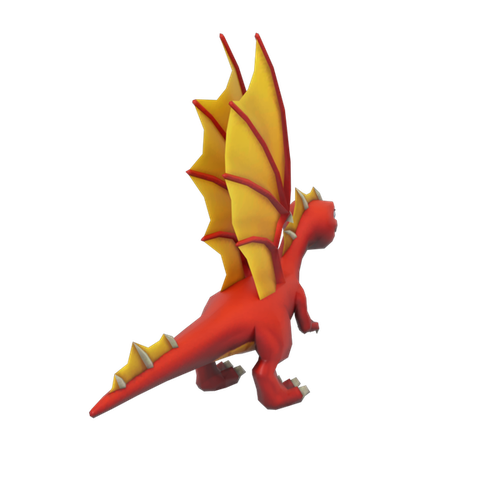} &
    \includegraphics[width=0.12\linewidth, trim={39.06 39.06 39.06 39.06}, clip]{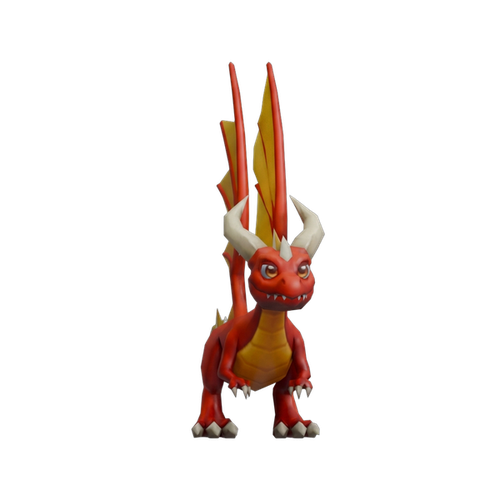} &
    \includegraphics[width=0.12\linewidth, trim={47.62 47.62 47.62 47.62}, clip]{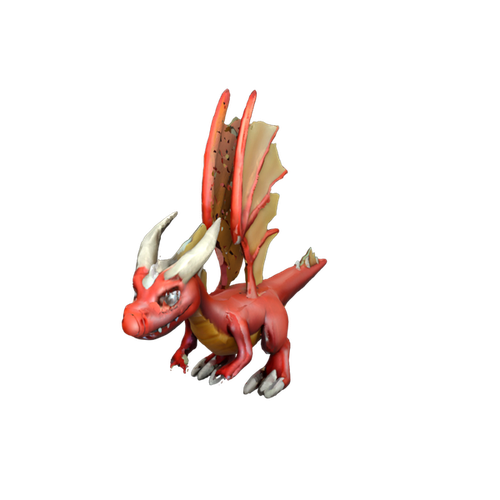} & \includegraphics[width=0.12\linewidth, trim={47.62 47.62 47.62 47.62}, clip]{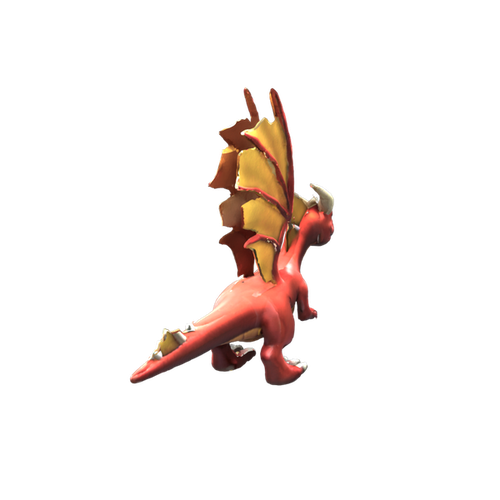} &
    \includegraphics[width=0.12\linewidth, trim={39.06 39.06 39.06 39.06}, clip]{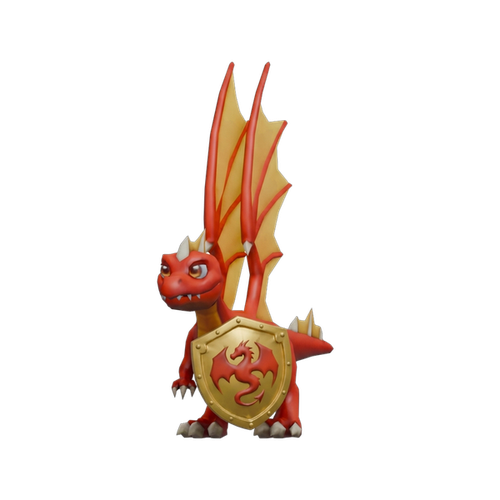} &
    \includegraphics[width=0.12\linewidth, trim={47.62 47.62 47.62 47.62}, clip]{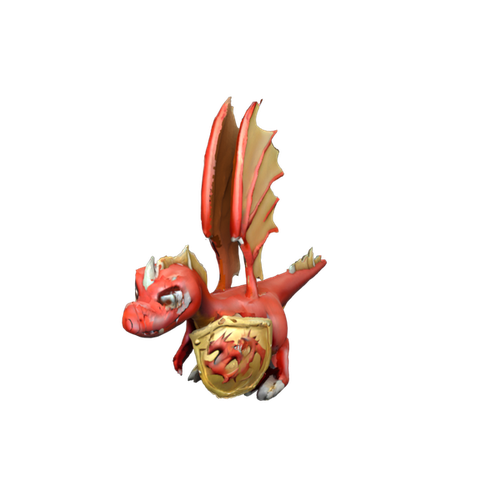} & \includegraphics[width=0.12\linewidth, trim={47.62 47.62 47.62 47.62}, clip]{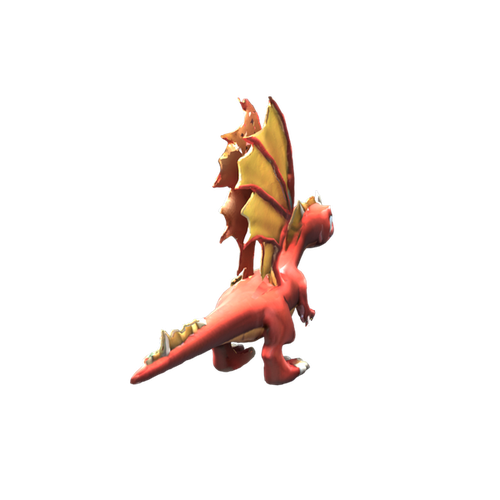} \\ [-3pt]

    \includegraphics[width=0.12\linewidth, trim={17.86 17.86 17.86 17.86}, clip]{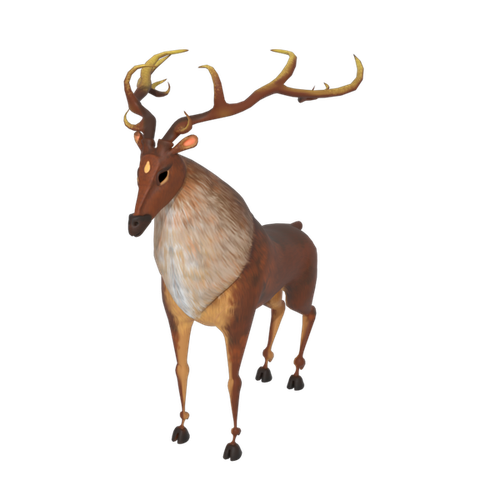} & \includegraphics[width=0.12\linewidth, trim={17.86 17.86 17.86 17.86}, clip]{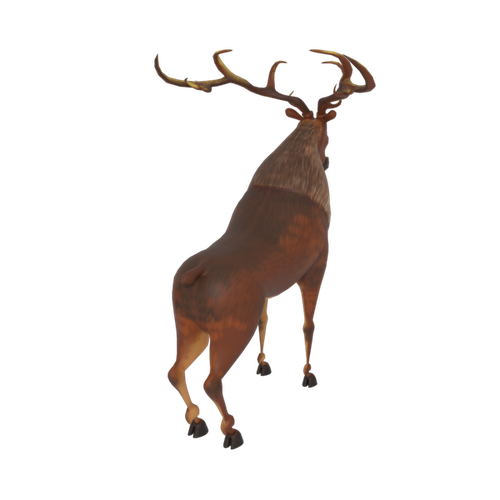} &
    \includegraphics[width=0.12\linewidth, trim={14.65 14.65 14.65 14.65}, clip]{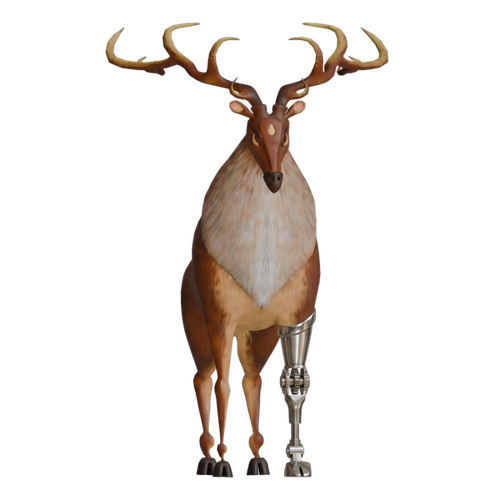} &
    \includegraphics[width=0.12\linewidth, trim={17.86 17.86 17.86 17.86}, clip]{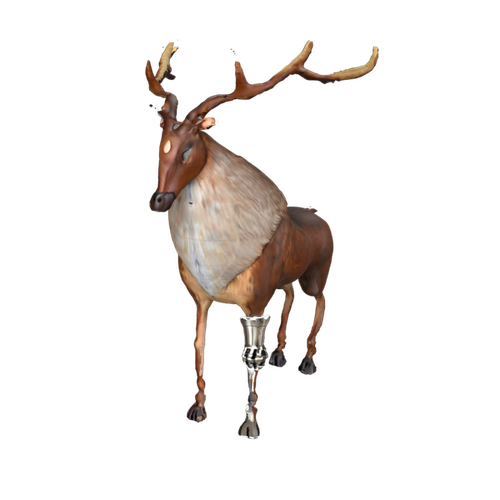} & \includegraphics[width=0.12\linewidth, trim={17.86 17.86 17.86 17.86}, clip]{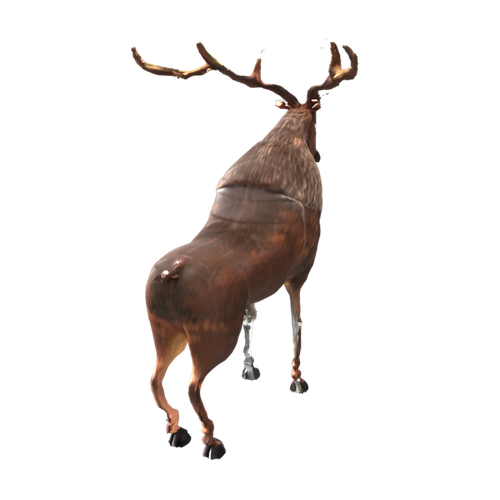} &
    \includegraphics[width=0.12\linewidth, trim={14.65 14.65 14.65 14.65}, clip]{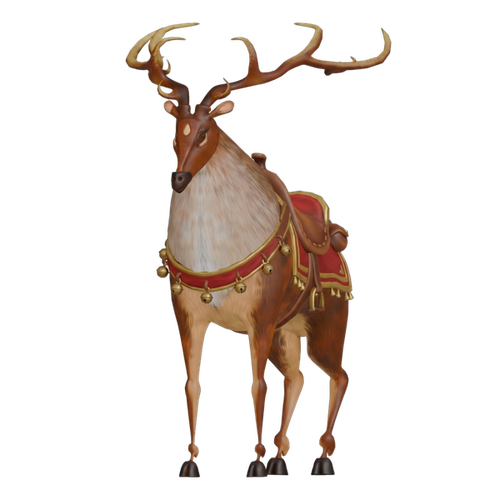} &
    \includegraphics[width=0.12\linewidth, trim={17.86 17.86 17.86 17.86}, clip]{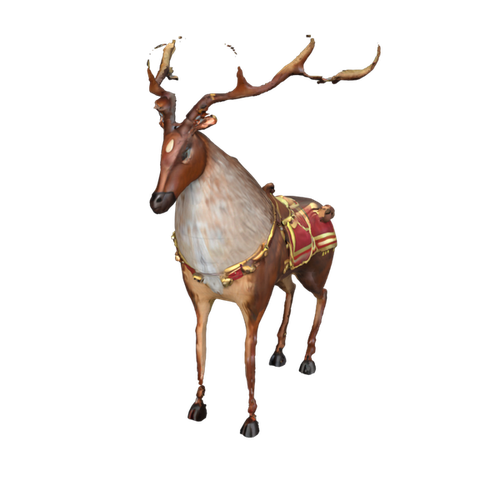} & \includegraphics[width=0.12\linewidth, trim={17.86 17.86 17.86 17.86}, clip]{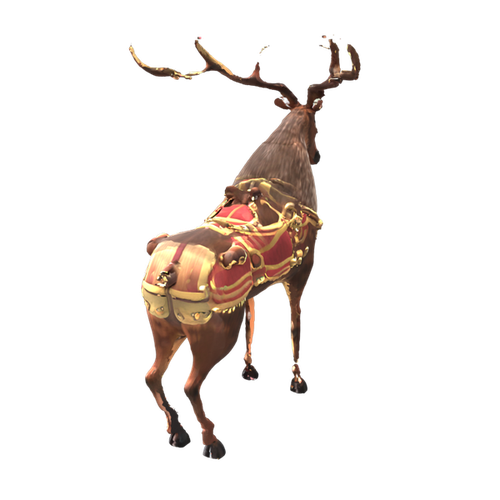} \\ [-1pt]

    \includegraphics[width=0.12\linewidth, trim={17.86 17.86 17.86 17.86}, clip]{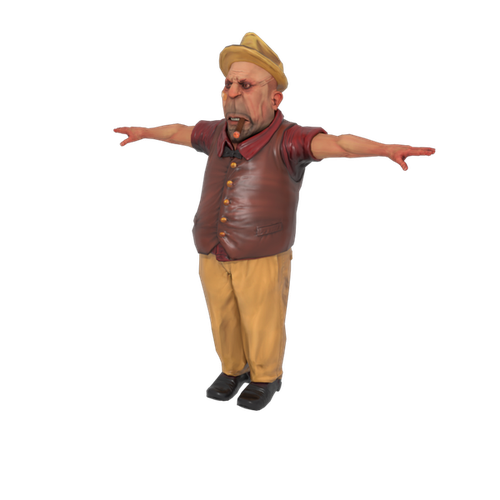} & \includegraphics[width=0.12\linewidth, trim={17.86 17.86 17.86 17.86}, clip]{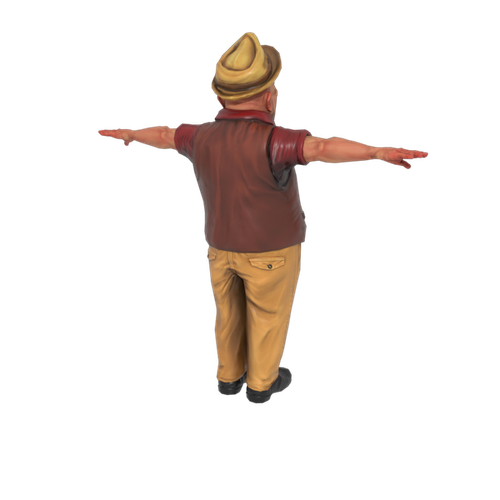} &
    \raisebox{3pt}{\includegraphics[width=0.115\linewidth, trim={0 0 0 0}, clip]{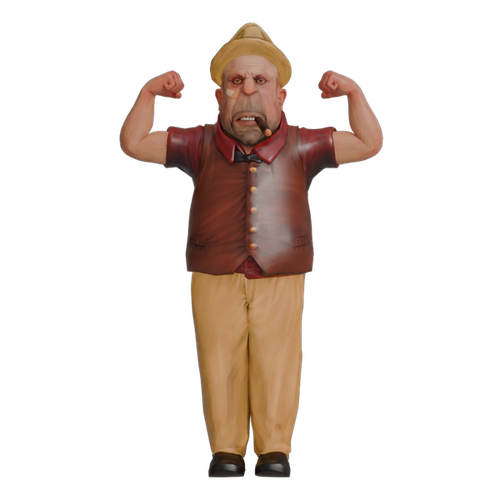}} &
    \includegraphics[width=0.12\linewidth, trim={17.86 17.86 17.86 17.86}, clip]{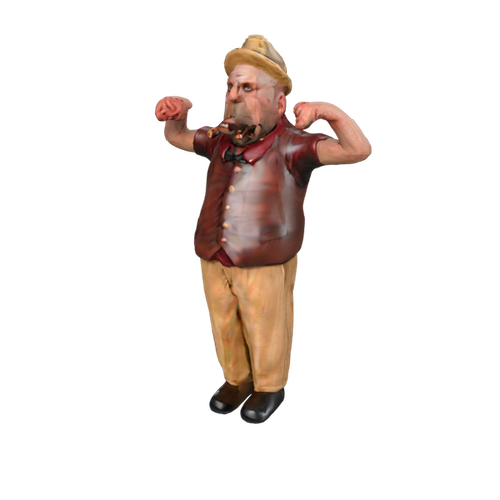} & \includegraphics[width=0.12\linewidth, trim={17.86 17.86 17.86 17.86}, clip]{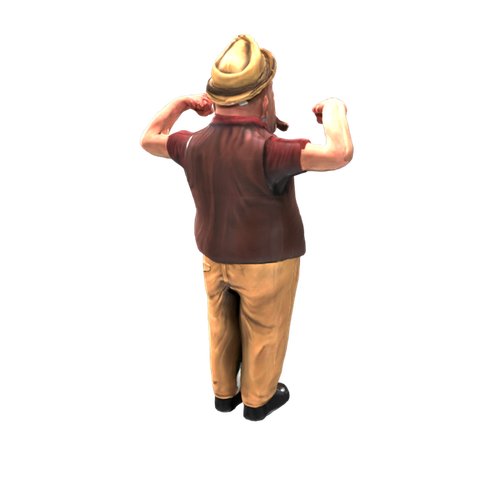} &
    \raisebox{3pt}{\includegraphics[width=0.115\linewidth, trim={0 0 0 0}, clip]{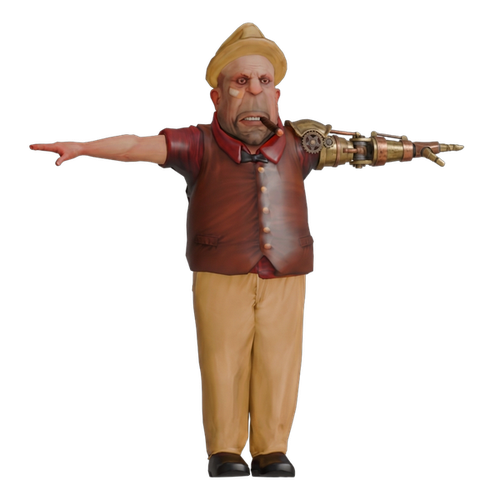}} &
    \includegraphics[width=0.12\linewidth, trim={17.86 17.86 17.86 17.86}, clip]{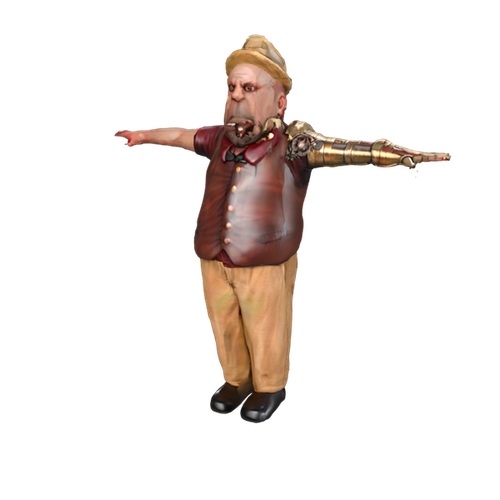} & \includegraphics[width=0.12\linewidth, trim={17.86 17.86 17.86 17.86}, clip]{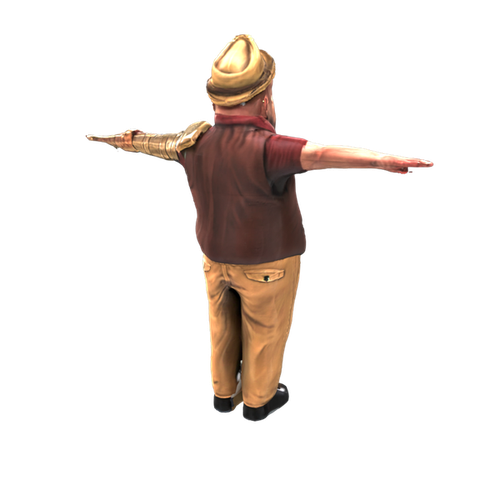} \\ [-4pt]

    \includegraphics[width=0.12\linewidth, trim={17.86 17.86 17.86 17.86}, clip]{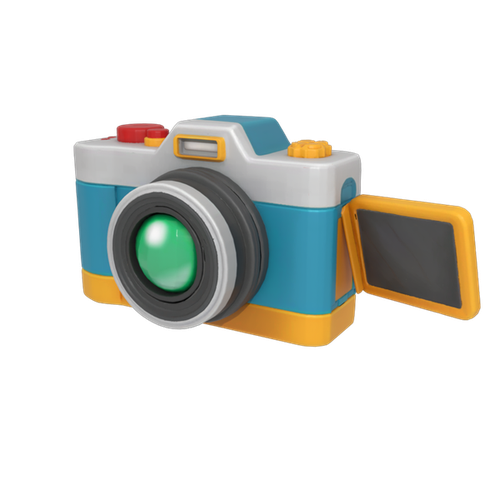} & \includegraphics[width=0.12\linewidth, trim={17.86 17.86 17.86 17.86}, clip]{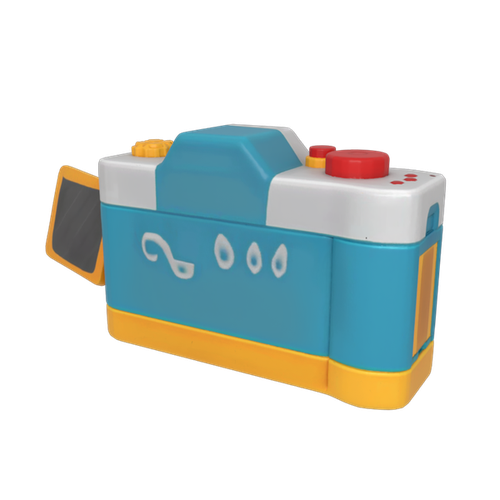} &
    \includegraphics[width=0.12\linewidth, trim={14.65 14.65 14.65 14.65}, clip]{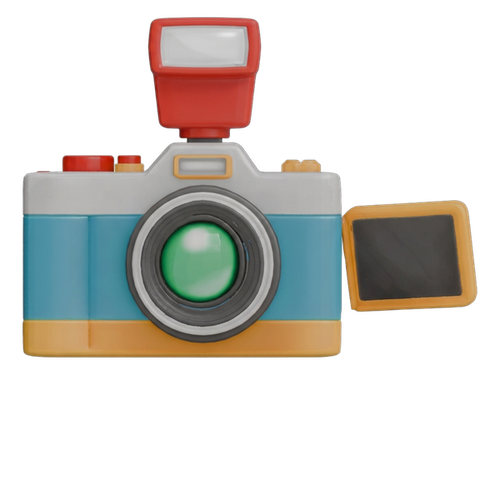} &
    \includegraphics[width=0.12\linewidth, trim={17.86 17.86 17.86 17.86}, clip]{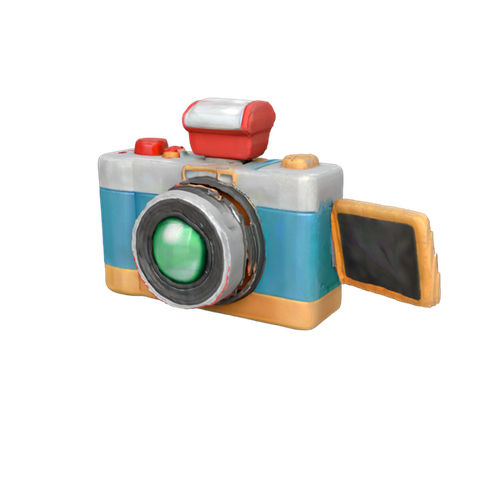} & \includegraphics[width=0.12\linewidth, trim={17.86 17.86 17.86 17.86}, clip]{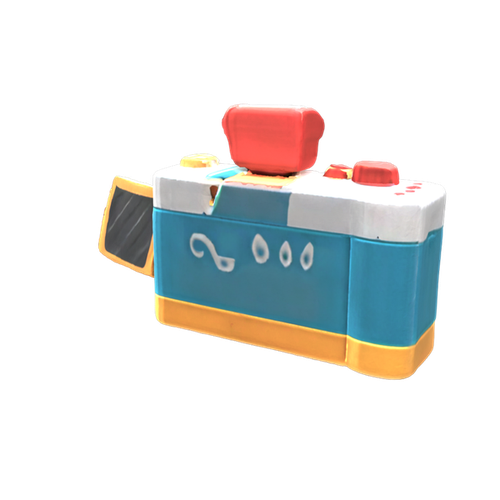} &
    \includegraphics[width=0.12\linewidth, trim={14.65 14.65 14.65 4.88}, clip]{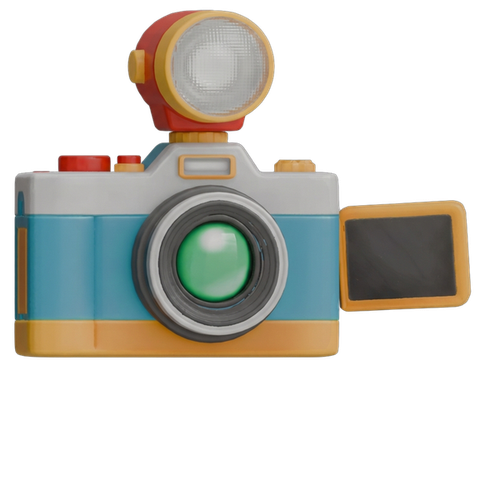} &
    \includegraphics[width=0.12\linewidth, trim={17.86 17.86 17.86 17.86}, clip]{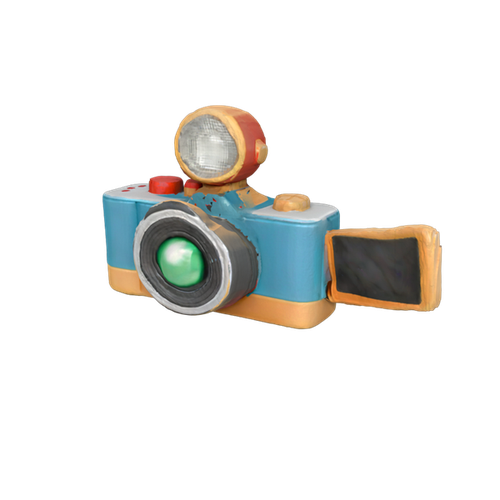} & \includegraphics[width=0.12\linewidth, trim={17.86 17.86 17.86 17.86}, clip]{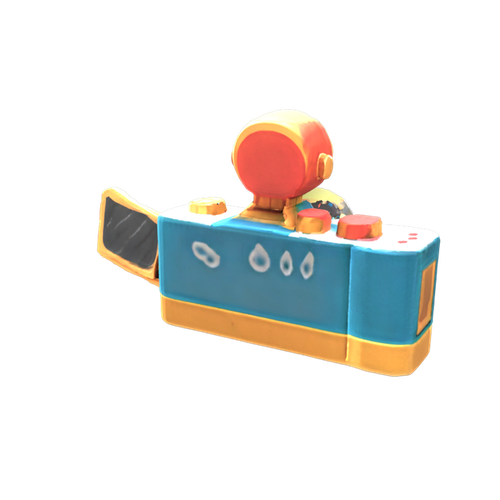} \\ [-6pt]

    \bottomrule
  \end{tabular}
}

  \caption{\textbf{Qualitative Results.} Columns 1--2 show the \textbf{Source Textured Mesh} (front/back). Columns 3--5 and 6--8 present \textbf{Edit A} and \textbf{Edit B}: each includes the \textbf{Edit Condition} and the edited mesh (front/back), demonstrating diverse semantic changes on one object.}
  \label{fig:more_results}
\end{figure*}

    \begin{figure*}[t]
  \centering
  \setlength{\tabcolsep}{1pt} 
  \renewcommand{\arraystretch}{0.6}

\resizebox{0.98\linewidth}{!}{%
\begin{tabular}{@{}cc!{\vrule width 0.6pt}ccc!{\vrule width 0.6pt}ccc@{}}
    \toprule
    \multicolumn{2}{c}{\scriptsize \textbf{Source Textured Mesh}} &
    \multicolumn{1}{c}{\scriptsize \textbf{Edit Condition}} &
    \multicolumn{2}{c}{\scriptsize \textbf{Edited}} &
    \multicolumn{1}{c}{\scriptsize \textbf{Edit Condition}} &
    \multicolumn{2}{c}{\scriptsize \textbf{Edited}} \\
    
    \cmidrule(r){1-2} \cmidrule(lr){3-3} \cmidrule(lr){4-5} \cmidrule(lr){6-6} \cmidrule(l){7-8}

    \includegraphics[width=0.12\linewidth, trim={17.86 17.86 17.86 17.86}, clip]{images/qualitative_results/Source_meshes/dannys_cat_60deg.png} & \includegraphics[width=0.12\linewidth, trim={17.86 17.86 17.86 17.86}, clip]{images/qualitative_results/Source_meshes/dannys_cat_240deg.png} &
    \includegraphics[width=0.115\linewidth, trim={0 0 0 0}, clip]{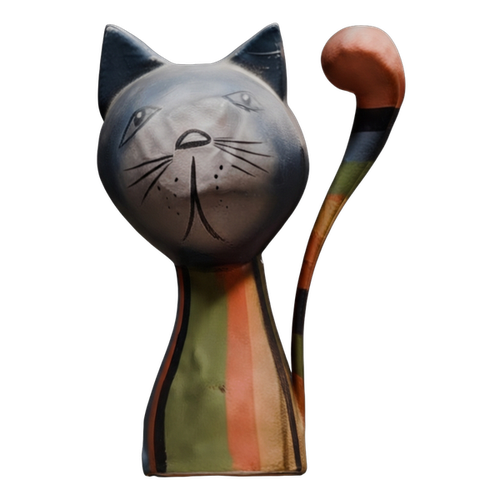} &
    \includegraphics[width=0.12\linewidth, trim={17.86 17.86 17.86 17.86}, clip]{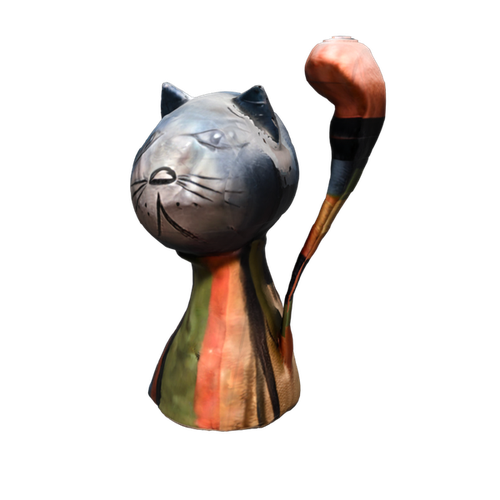} & \includegraphics[width=0.12\linewidth, trim={17.86 17.86 17.86 17.86}, clip]{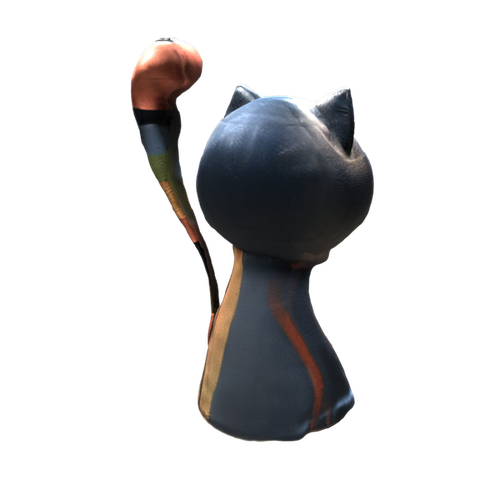} &
    \includegraphics[width=0.115\linewidth, trim={0 0 0 0}, clip]{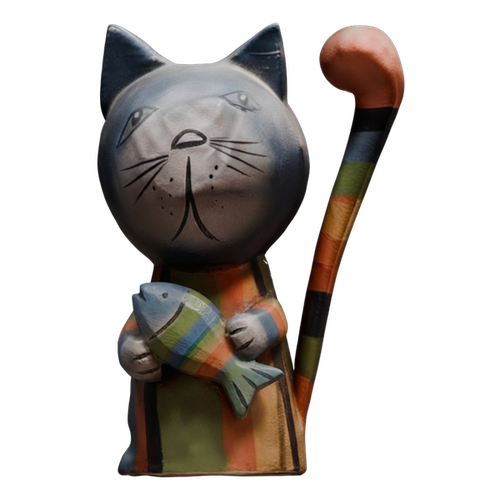} &
    \includegraphics[width=0.12\linewidth, trim={17.86 17.86 17.86 17.86}, clip]{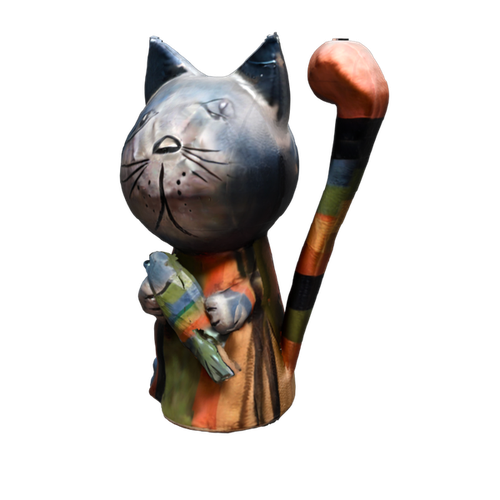} & \includegraphics[width=0.12\linewidth, trim={17.86 17.86 17.86 17.86}, clip]{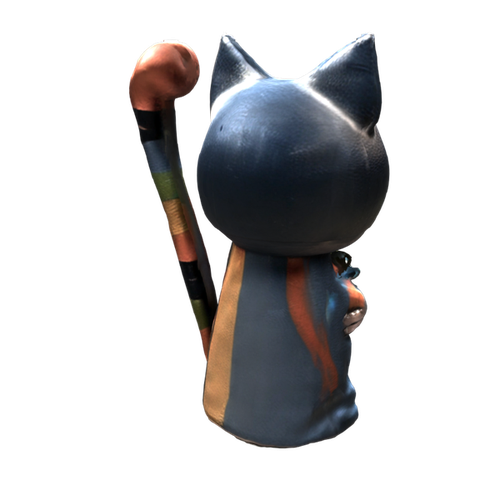} \\ [-3pt]
    
    \includegraphics[width=0.12\linewidth, trim={17.86 17.86 17.86 17.86}, clip]{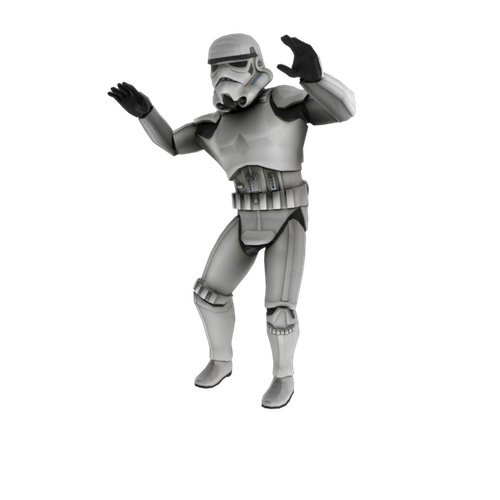} & \includegraphics[width=0.12\linewidth, trim={17.86 17.86 17.86 17.86}, clip]{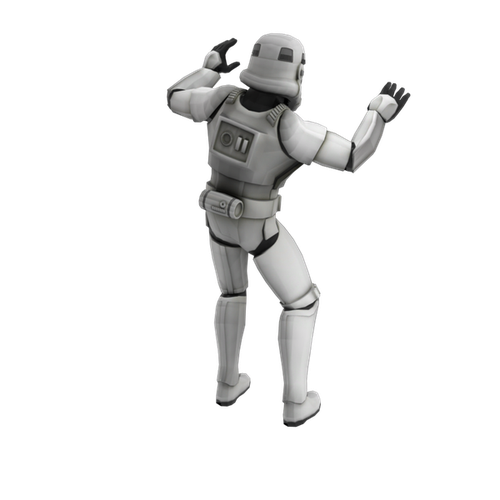} &
    \raisebox{3pt}{\includegraphics[width=0.115\linewidth, trim={0 0 0 0}, clip]{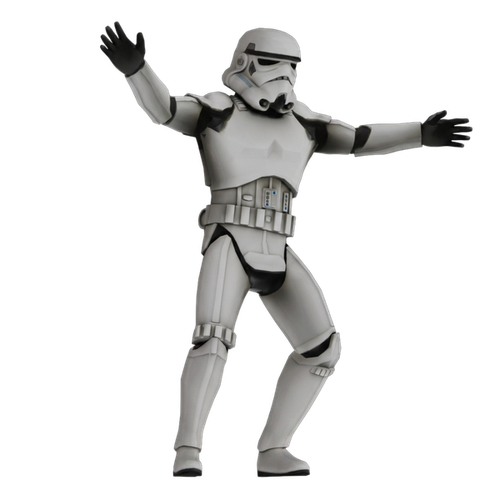}} &
    \includegraphics[width=0.12\linewidth, trim={47.62 47.62 47.62 47.62}, clip]{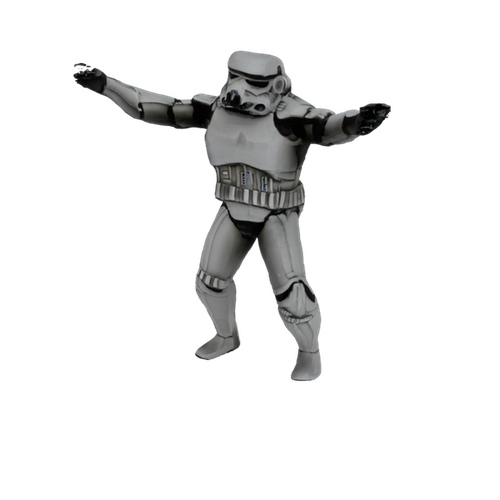} & \includegraphics[width=0.12\linewidth, trim={59.52 59.52 59.52 41.67}, clip]{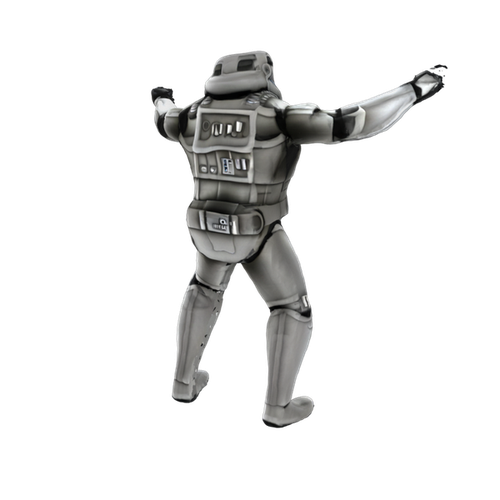} &
    \raisebox{3pt}{\includegraphics[width=0.115\linewidth, trim={0 0 0 0}, clip]{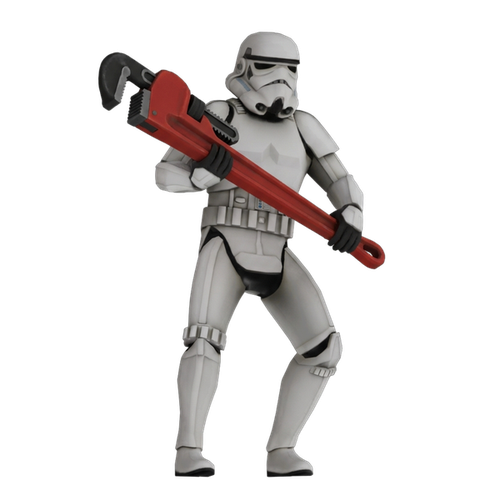}} &
    \includegraphics[width=0.12\linewidth, trim={47.62 47.62 47.62 23.81}, clip]{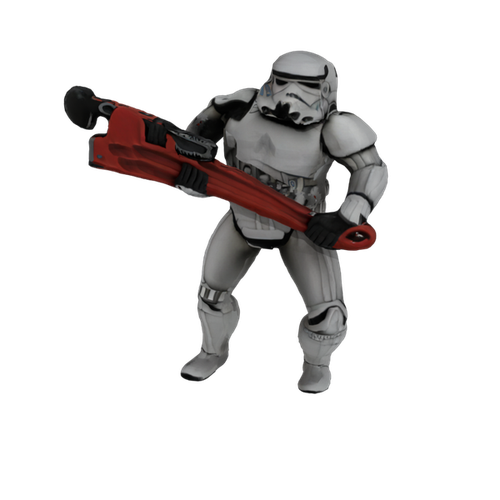} & \includegraphics[width=0.12\linewidth, trim={47.62 47.62 47.62 23.81}, clip]{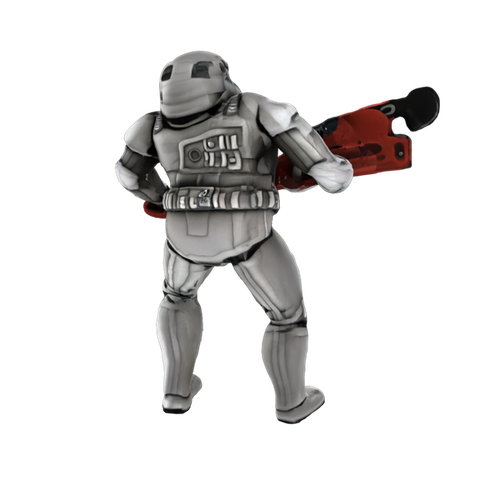} \\ [-5pt]

    \includegraphics[width=0.12\linewidth, trim={17.86 17.86 17.86 17.86}, clip]{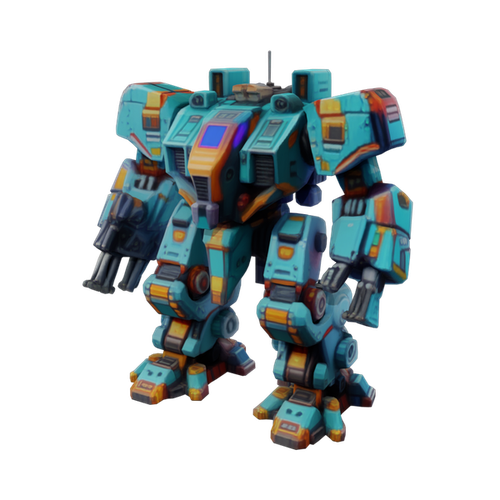} & \includegraphics[width=0.12\linewidth, trim={17.86 17.86 17.86 17.86}, clip]{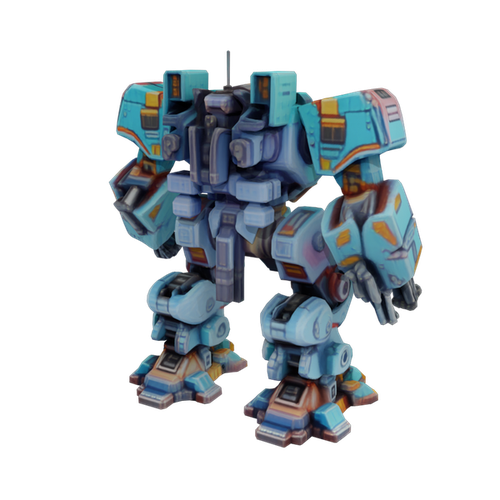} &
    \raisebox{3pt}{\includegraphics[width=0.115\linewidth, trim={0 0 0 0}, clip]{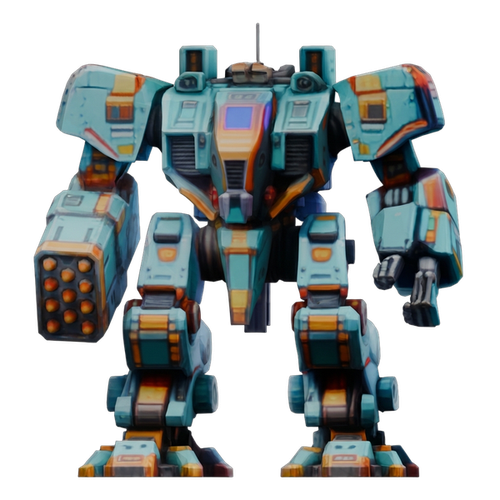}} &
    \includegraphics[width=0.12\linewidth, trim={17.86 17.86 17.86 17.86}, clip]{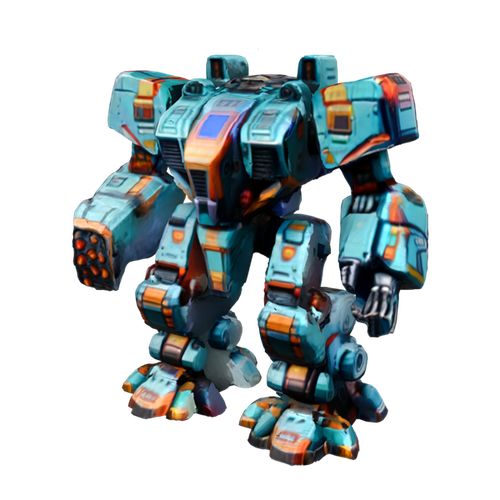} &
    \includegraphics[width=0.12\linewidth, trim={17.86 17.86 17.86 17.86}, clip]{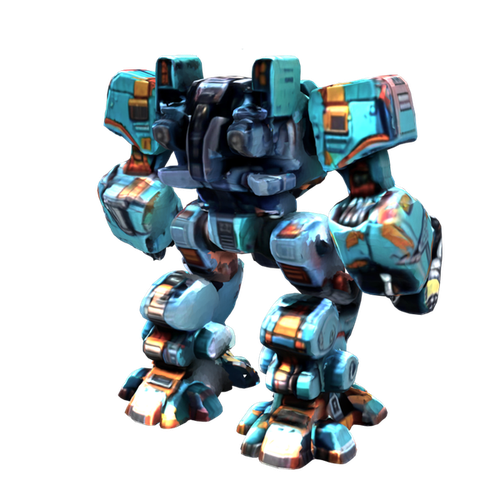} &
    \raisebox{3pt}{\includegraphics[width=0.115\linewidth, trim={0 0 0 0}, clip]{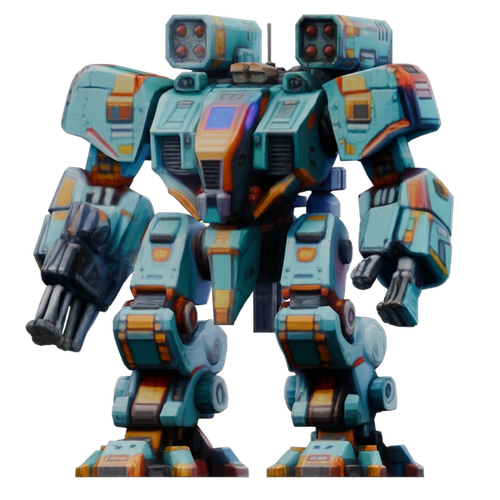}} &
    \includegraphics[width=0.12\linewidth, trim={17.86 17.86 17.86 17.86}, clip]{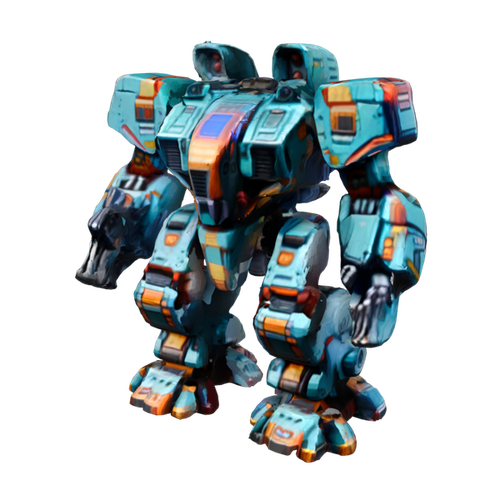} &
    \includegraphics[width=0.12\linewidth, trim={17.86 17.86 17.86 17.86}, clip]{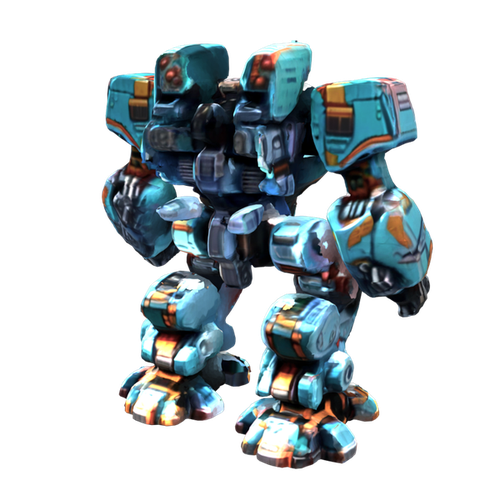} \\

    \includegraphics[width=0.12\linewidth, trim={17.86 17.86 17.86 17.86}, clip]{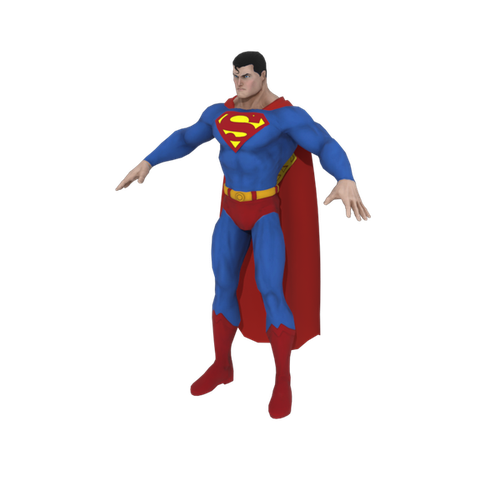} & \includegraphics[width=0.12\linewidth, trim={17.86 17.86 17.86 17.86}, clip]{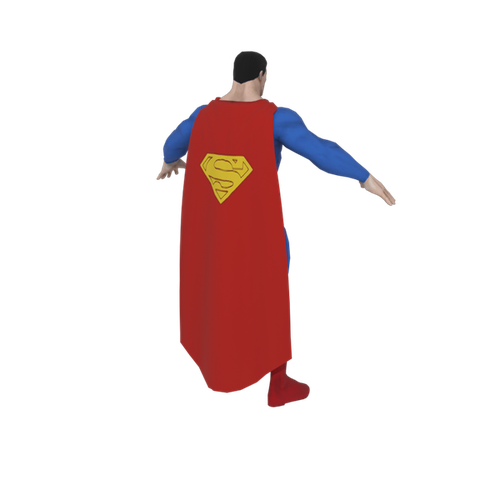} &
    \includegraphics[width=0.12\linewidth, trim={14.65 14.65 14.65 14.65}, clip]{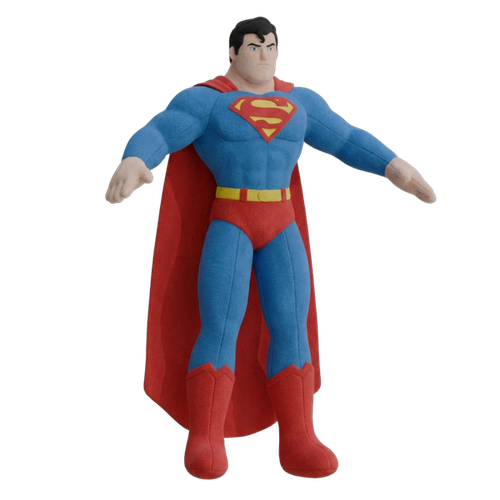} &
    \includegraphics[width=0.12\linewidth, trim={17.86 17.86 17.86 17.86}, clip]{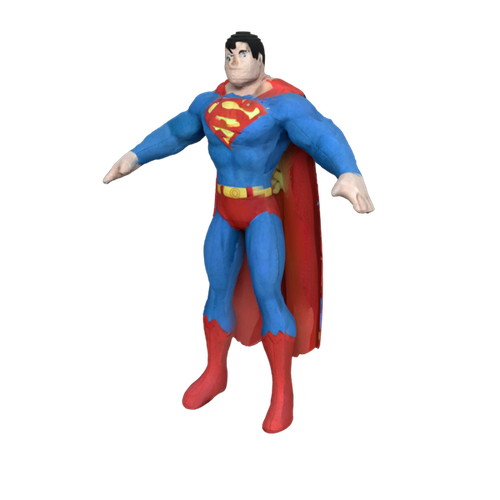} & \includegraphics[width=0.12\linewidth, trim={17.86 17.86 17.86 17.86}, clip]{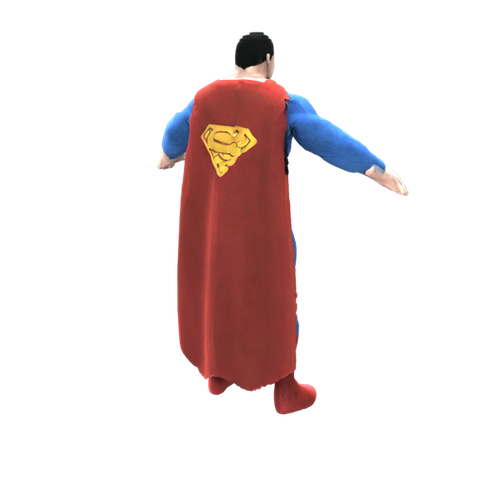} &
    \includegraphics[width=0.12\linewidth, trim={14.65 14.65 14.65 14.65}, clip]{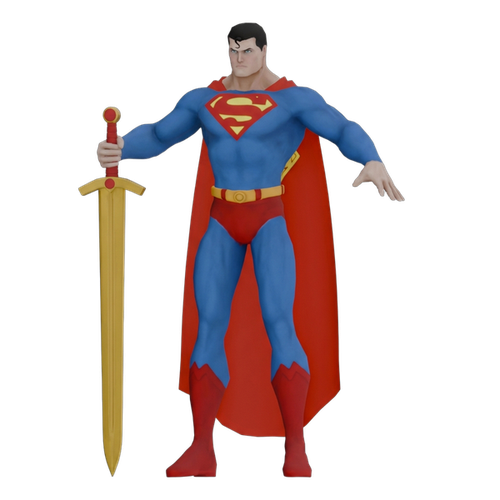} &
    \includegraphics[width=0.12\linewidth, trim={17.86 17.86 17.86 17.86}, clip]{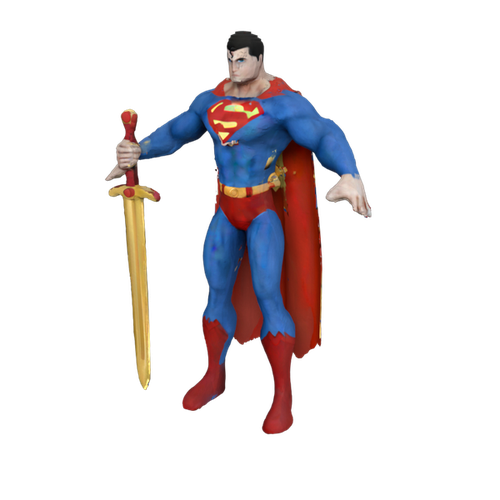} & \includegraphics[width=0.12\linewidth, trim={17.86 17.86 17.86 17.86}, clip]{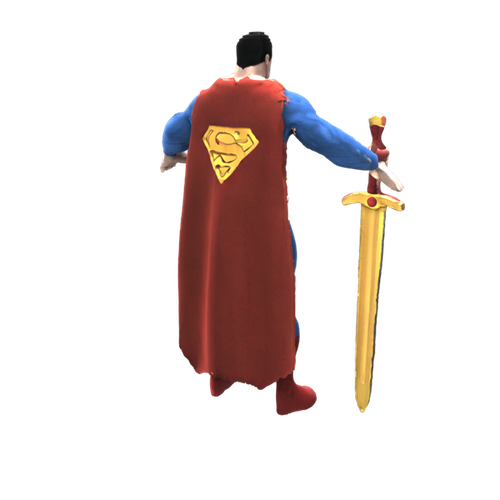}  \\

    \includegraphics[width=0.12\linewidth, trim={17.86 17.86 17.86 17.86}, clip]{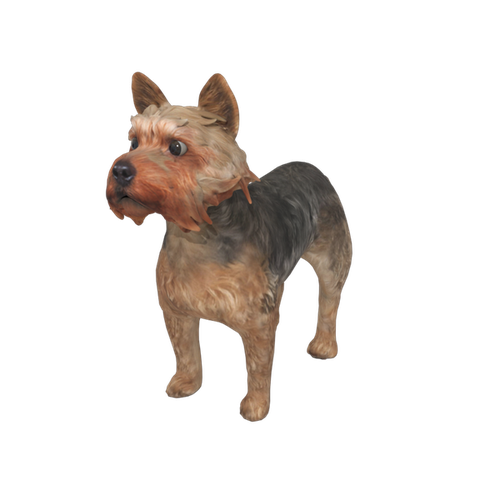} & \includegraphics[width=0.12\linewidth, trim={17.86 17.86 17.86 17.86}, clip]{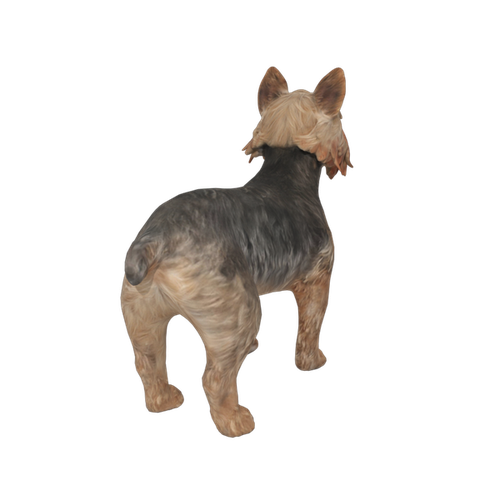} &
    \includegraphics[width=0.12\linewidth, trim={14.65 14.65 14.65 14.65}, clip]{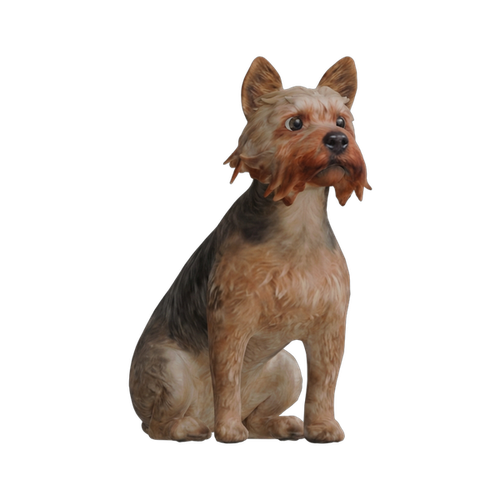} &
    \includegraphics[width=0.12\linewidth, trim={17.86 17.86 17.86 17.86}, clip]{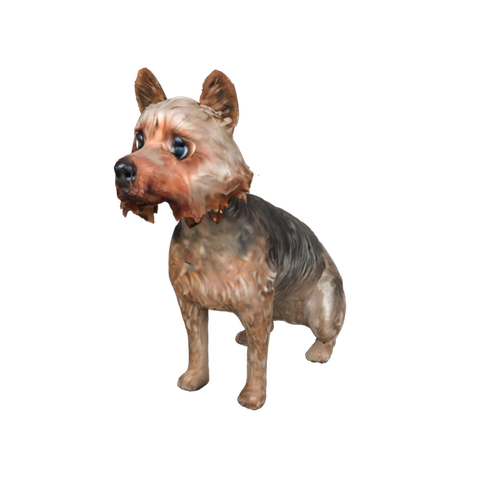} & \includegraphics[width=0.12\linewidth, trim={17.86 17.86 17.86 17.86}, clip]{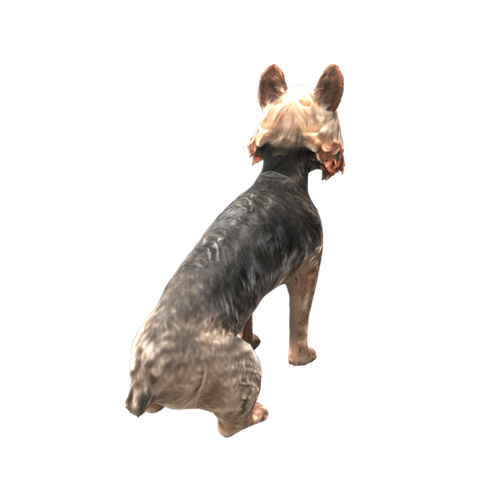} &
    \includegraphics[width=0.12\linewidth, trim={14.65 14.65 14.65 14.65}, clip]{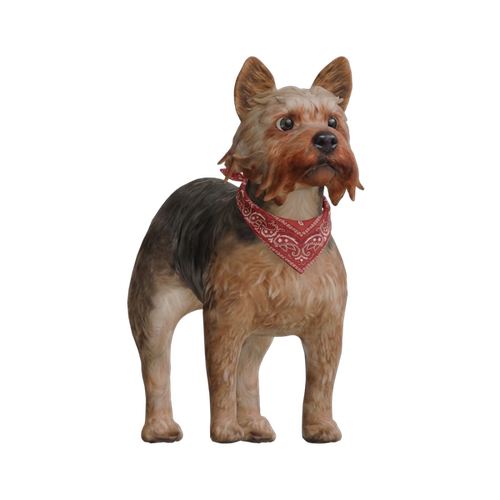} &
    \includegraphics[width=0.12\linewidth, trim={17.86 17.86 17.86 17.86}, clip]{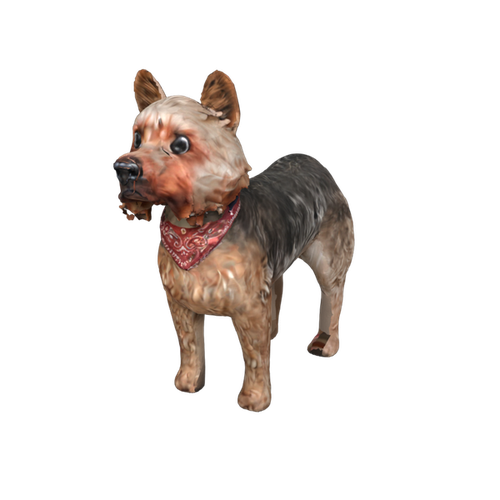} & \includegraphics[width=0.12\linewidth, trim={17.86 17.86 17.86 17.86}, clip]{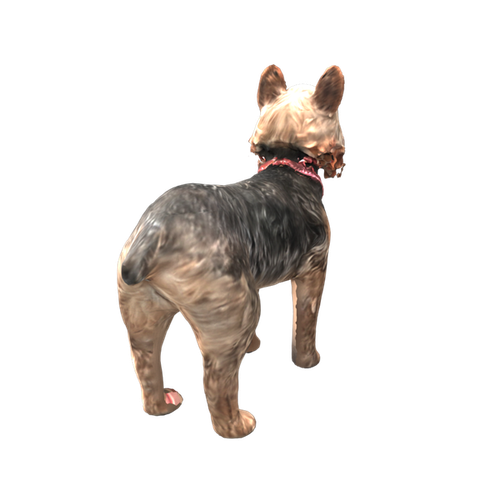} \\ [-5pt]

    \includegraphics[width=0.12\linewidth, trim={17.86 17.86 17.86 17.86}, clip]{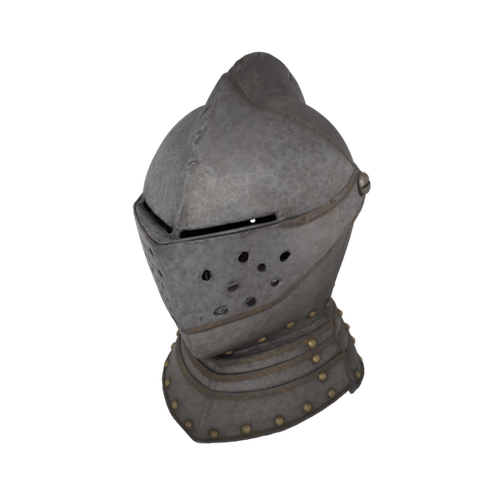} & \includegraphics[width=0.12\linewidth, trim={17.86 17.86 17.86 17.86}, clip]{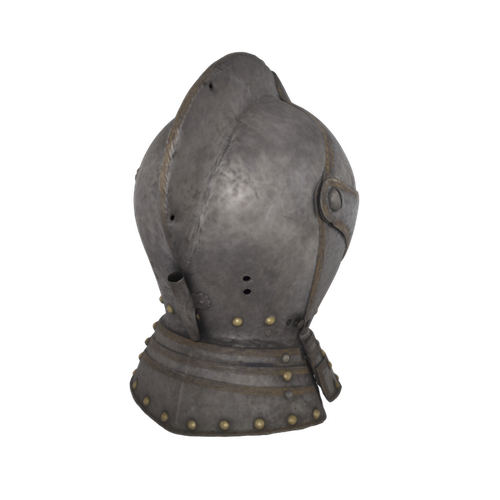} &
    \includegraphics[width=0.115\linewidth, trim={0 0 0 0}, clip]{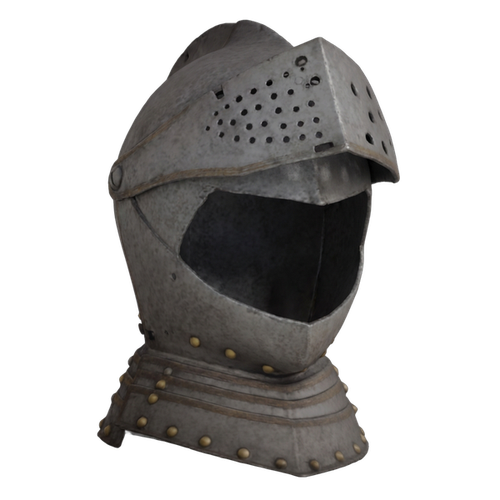} &
    \includegraphics[width=0.12\linewidth, trim={17.86 17.86 17.86 17.86}, clip]{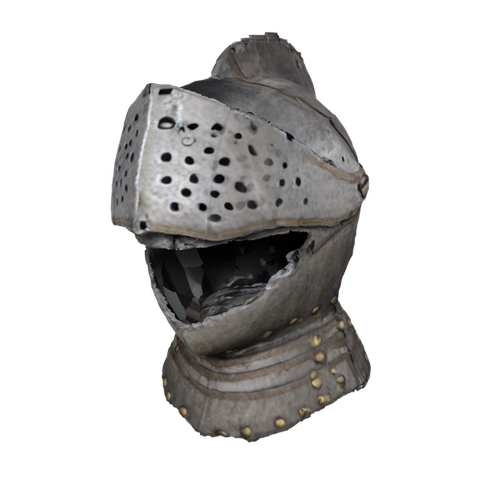} & \includegraphics[width=0.12\linewidth, trim={17.86 17.86 17.86 17.86}, clip]{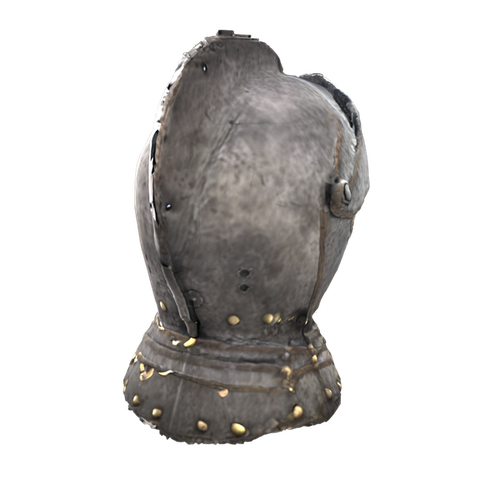} &
    \includegraphics[width=0.115\linewidth, trim={0 0 0 0}, clip]{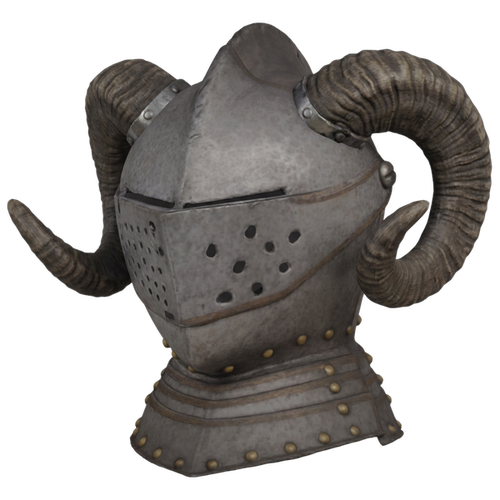} &
    \includegraphics[width=0.12\linewidth, trim={17.86 17.86 17.86 17.86}, clip]{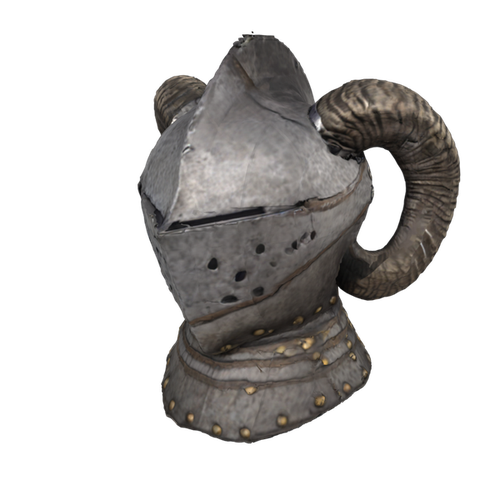} & \includegraphics[width=0.12\linewidth, trim={17.86 17.86 17.86 17.86}, clip]{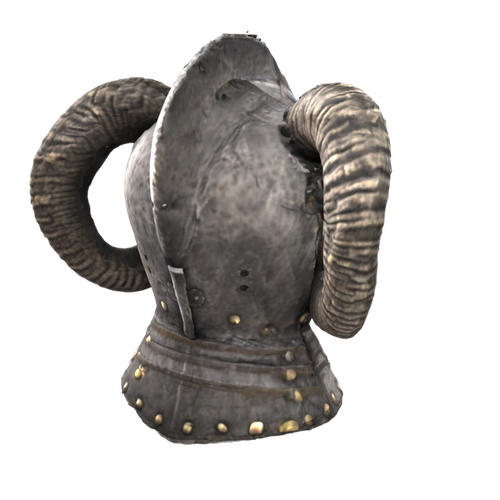} \\

    \includegraphics[width=0.12\linewidth, trim={17.86 17.86 17.86 17.86}, clip]{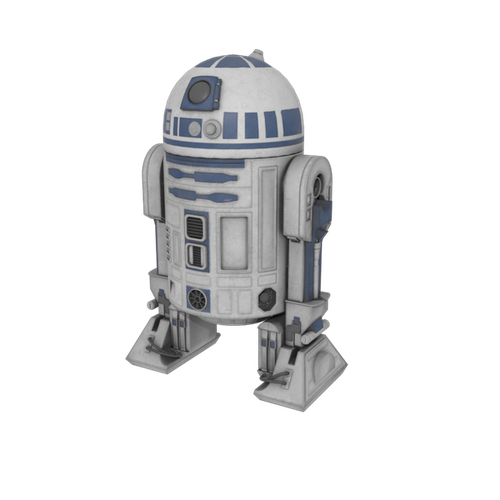} & \includegraphics[width=0.12\linewidth, trim={17.86 17.86 17.86 17.86}, clip]{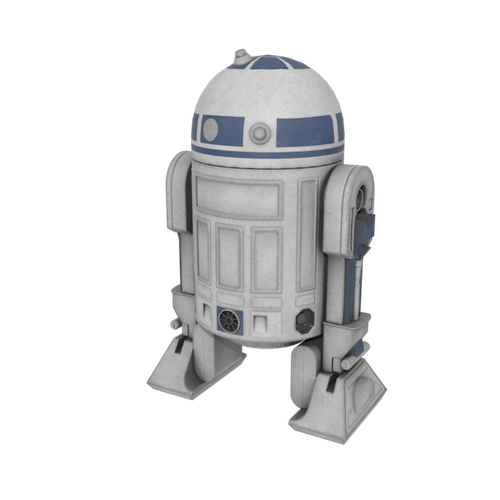} &
    \raisebox{3pt}{\includegraphics[width=0.115\linewidth, trim={0 0 0 0}, clip]{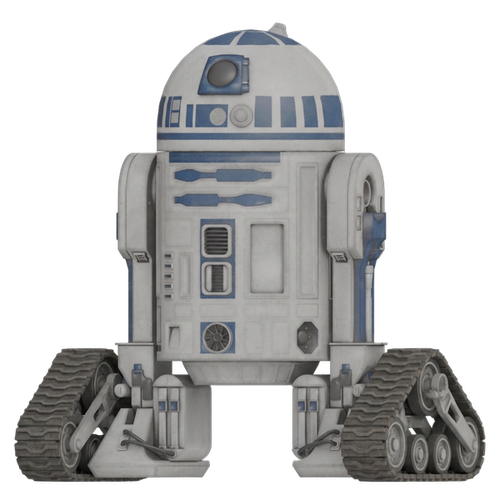}} &
    \includegraphics[width=0.12\linewidth, trim={17.86 5.95 17.86 17.86}, clip]{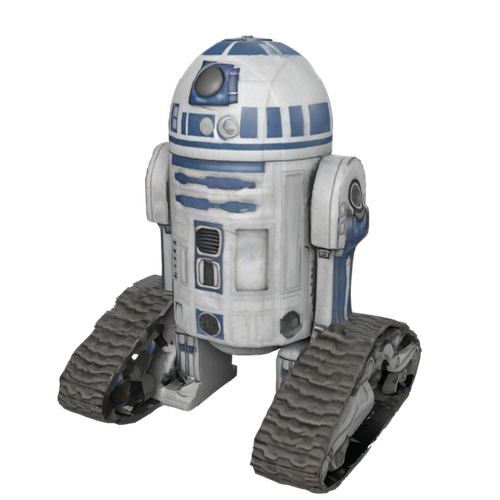} &
    \includegraphics[width=0.12\linewidth, trim={17.86 5.95 17.86 17.86}, clip]{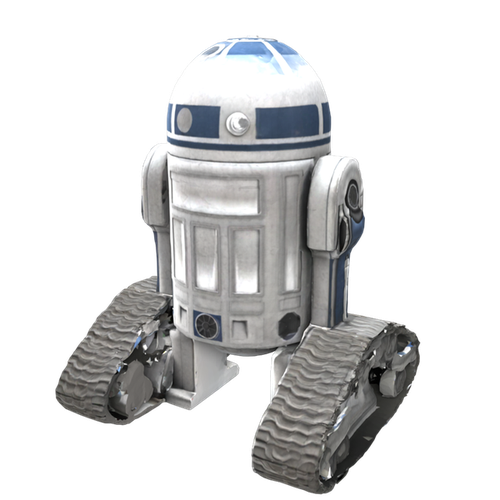} &
    \raisebox{3pt}{\includegraphics[width=0.115\linewidth, trim={0 0 0 0}, clip]{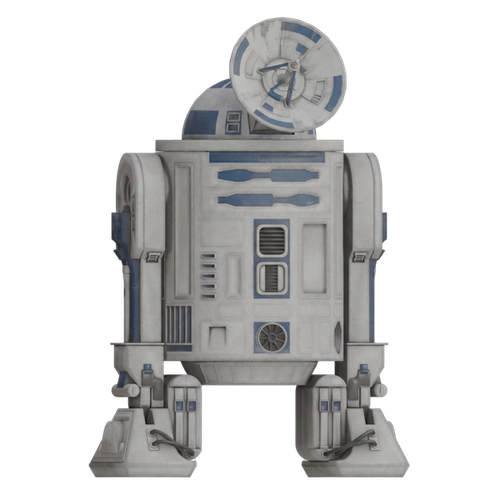}} &
    \includegraphics[width=0.12\linewidth, trim={17.86 17.86 17.86 17.86}, clip]{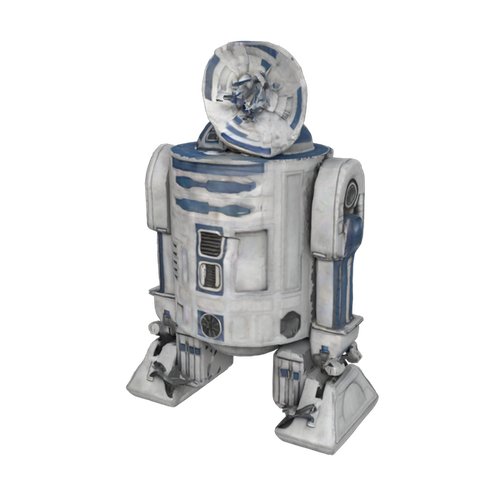} &
    \includegraphics[width=0.12\linewidth, trim={17.86 17.86 17.86 17.86}, clip]{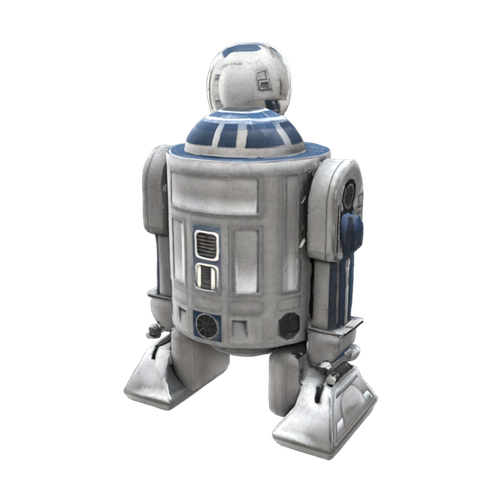} \\ [4pt]

    \includegraphics[width=0.12\linewidth, trim={47.62 47.62 47.62 23.81}, clip]{images/qualitative_results/Source_meshes/stormtrooper_60deg.png} & \includegraphics[width=0.12\linewidth, trim={47.62 47.62 47.62 23.81}, clip]{images/qualitative_results/Source_meshes/stormtrooper_240deg.png} &
    \raisebox{3pt}{\includegraphics[width=0.115\linewidth, trim={0 0 0 0}, clip]{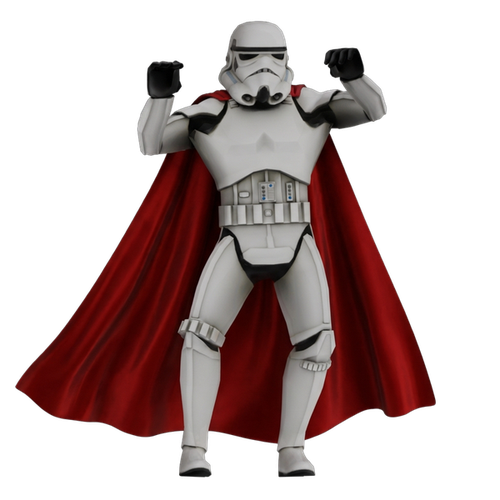}} &
    \includegraphics[width=0.12\linewidth, trim={17.86 17.86 17.86 17.86}, clip]{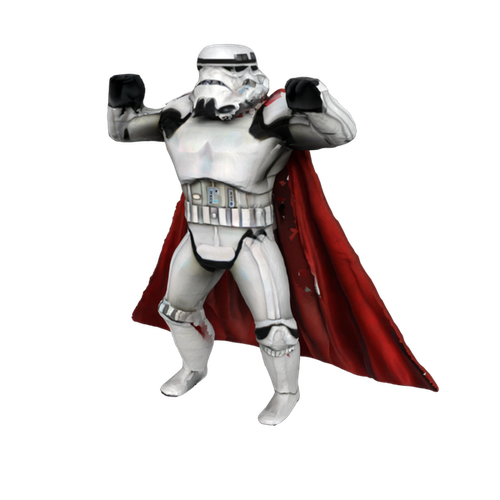} & \includegraphics[width=0.12\linewidth, trim={17.86 17.86 17.86 17.86}, clip]{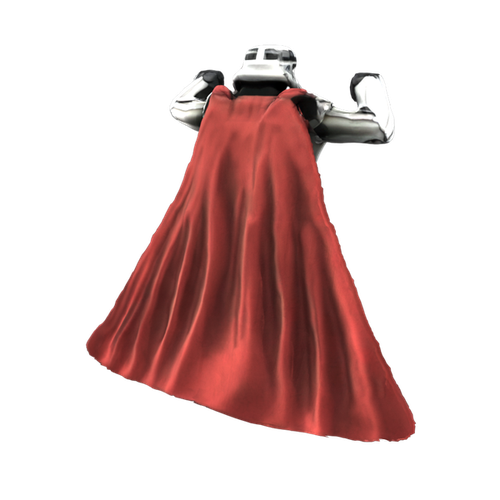} &
    \raisebox{3pt}{\includegraphics[width=0.115\linewidth, trim={0 0 0 0}, clip]{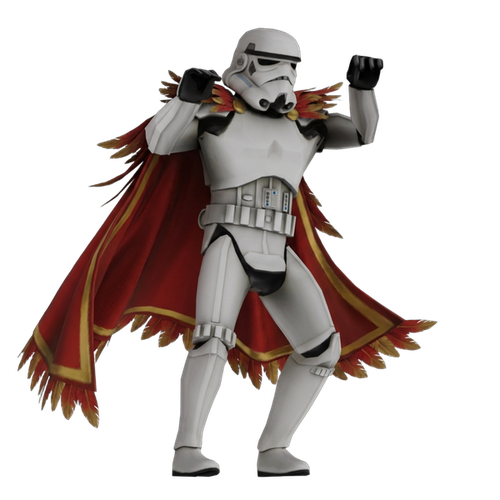}} &
    \includegraphics[width=0.12\linewidth, trim={47.62 47.62 47.62 23.81}, clip]{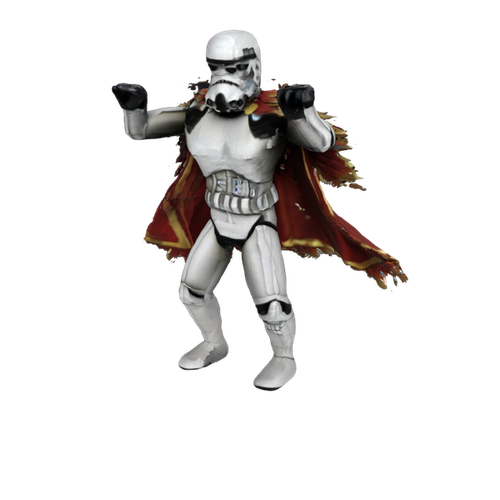} & \includegraphics[width=0.12\linewidth, trim={47.62 47.62 47.62 23.81}, clip]{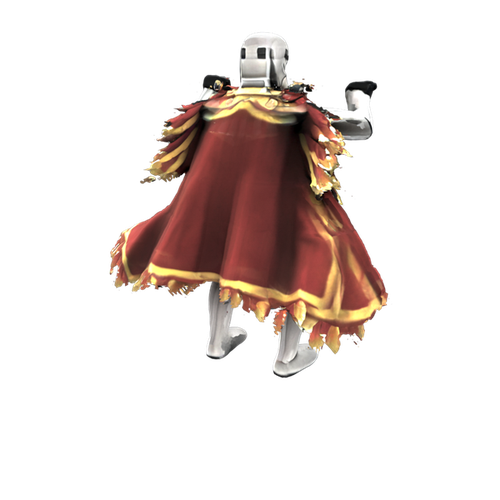} \\ [-8pt]

    \includegraphics[width=0.12\linewidth, trim={17.86 17.86 17.86 17.86}, clip]{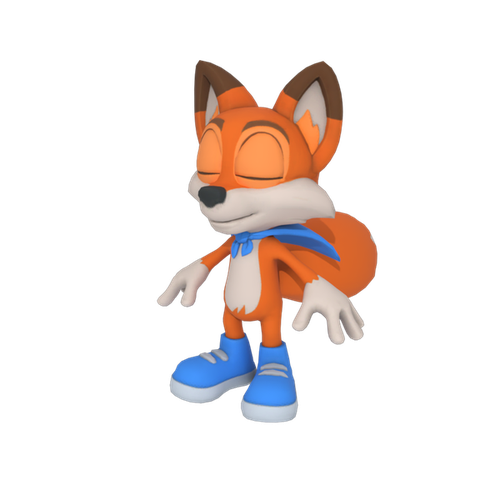} & \includegraphics[width=0.12\linewidth, trim={17.86 17.86 17.86 17.86}, clip]{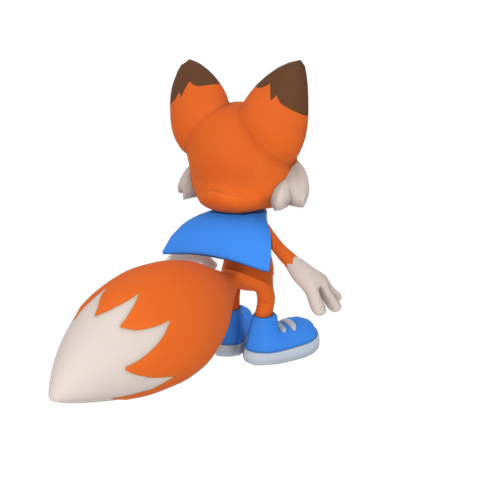} &
    \includegraphics[width=0.12\linewidth, trim={14.65 14.65 14.65 14.65}, clip]{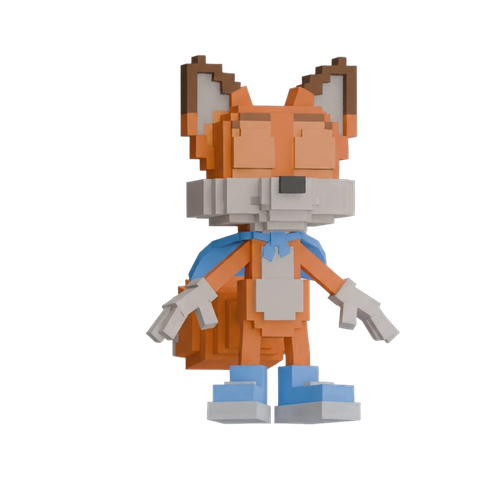} &
    \raisebox{3pt}{\hspace{6pt}{\includegraphics[width=0.12\linewidth, trim={0 17.86 47.62 17.86}, clip]{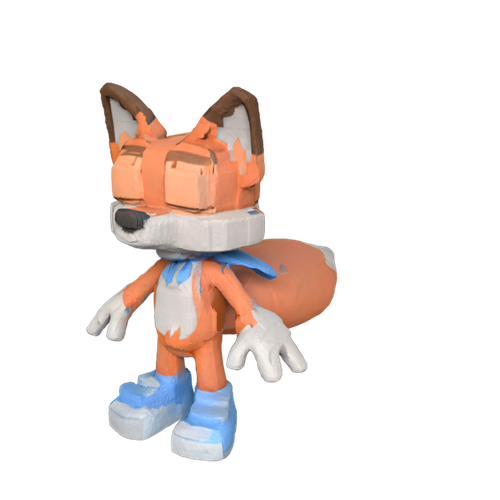}}
     } & \raisebox{-7pt}{\hspace{-11pt}{\includegraphics[width=0.142\linewidth, trim={17.86 17.86 17.86 17.86}, clip]{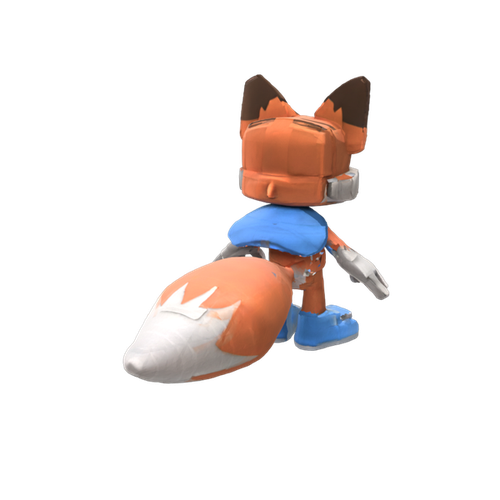}}
     } &
    \includegraphics[width=0.12\linewidth, trim={14.65 14.65 14.65 14.65}, clip]{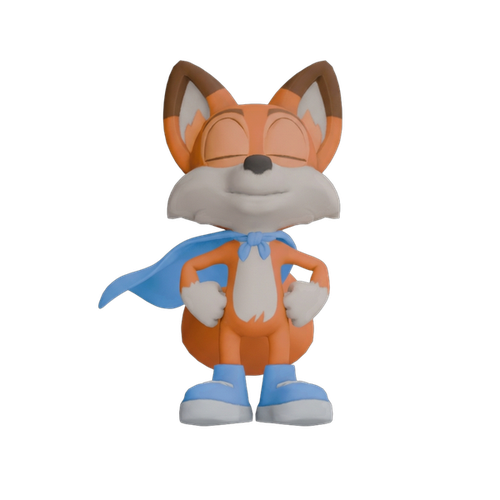} &
    \raisebox{3pt}{\hspace{6pt}{\includegraphics[width=0.12\linewidth, trim={0 17.86 47.62 17.86}, clip]{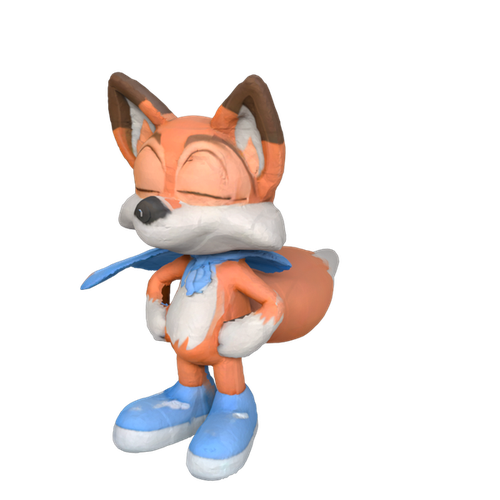}}
     } & \raisebox{-7pt}{\hspace{-11pt}{\includegraphics[width=0.142\linewidth, trim={17.86 17.86 17.86 17.86}, clip]{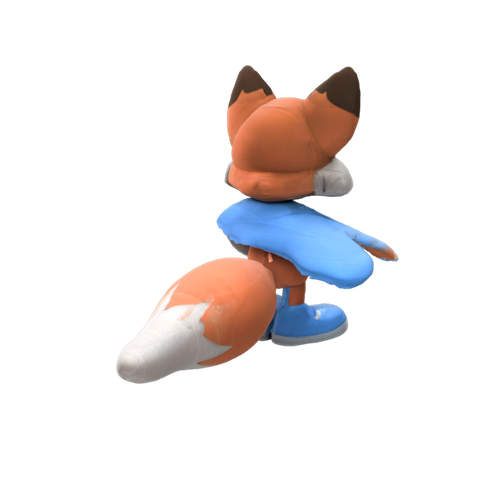}}
     } \\ [-3pt]

    \includegraphics[width=0.12\linewidth, trim={17.86 17.86 17.86 17.86}, clip]{images/qualitative_results/Source_meshes/camera_60deg.png} & \includegraphics[width=0.12\linewidth, trim={17.86 17.86 17.86 17.86}, clip]{images/qualitative_results/Source_meshes/camera_240deg.png} &
    \includegraphics[width=0.12\linewidth, trim={14.65 14.65 14.65 14.65}, clip]{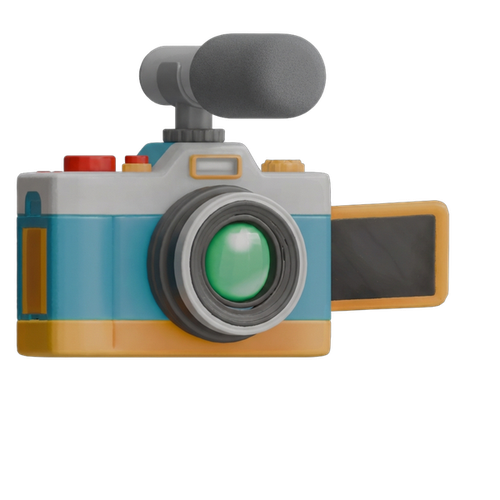} &
    \includegraphics[width=0.12\linewidth, trim={17.86 17.86 17.86 17.86}, clip]{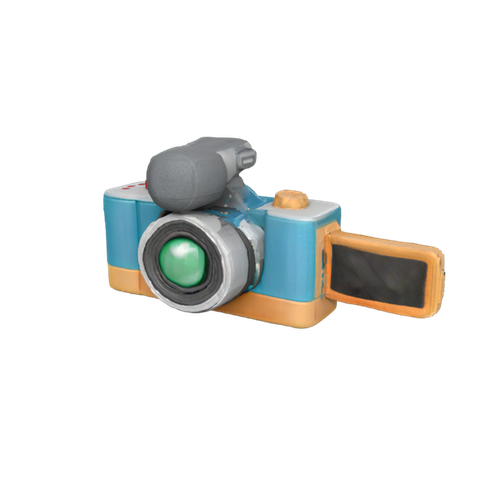} & \includegraphics[width=0.12\linewidth, trim={17.86 17.86 17.86 17.86}, clip]{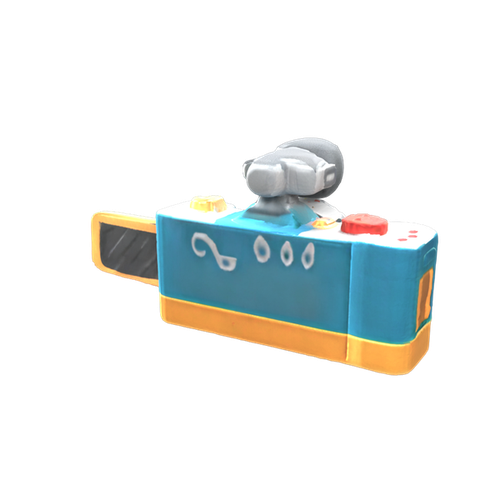} &
    \includegraphics[width=0.12\linewidth, trim={14.65 14.65 14.65 14.65}, clip]{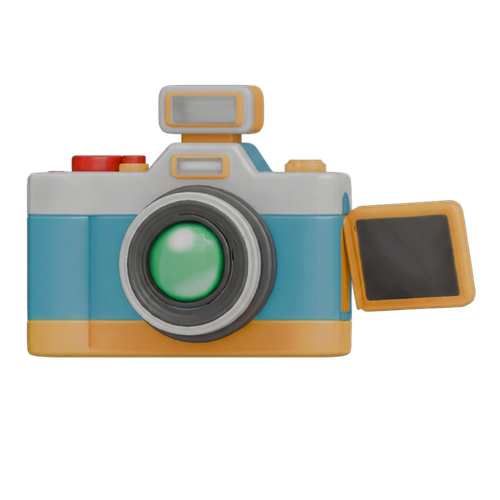} &
    \includegraphics[width=0.12\linewidth, trim={17.86 17.86 17.86 17.86}, clip]{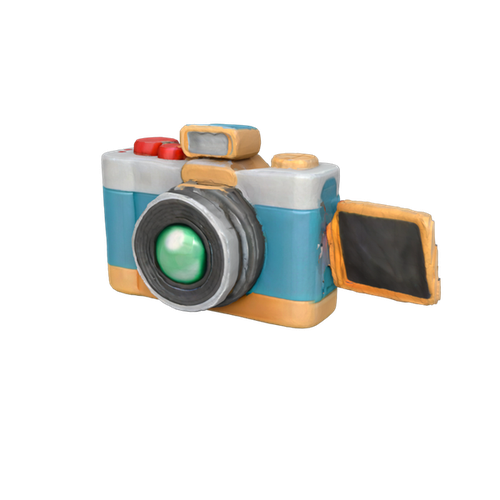} & \includegraphics[width=0.12\linewidth, trim={17.86 17.86 17.86 17.86}, clip]{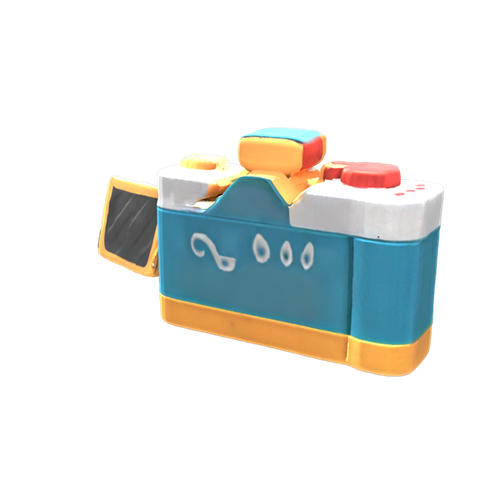} \\ [-3pt]

    \bottomrule
  \end{tabular}
  }%

  \caption{\textbf{Qualitative Results.} Columns 1--2 show the \textbf{Source Textured Mesh} (front/back). Columns 3--5 and 6--8 present \textbf{Edit A} and \textbf{Edit B}: each includes the \textbf{Edit Condition} and the edited mesh (front/back), demonstrating diverse semantic changes on one object.}
  \label{fig:more_results2}
\end{figure*}

    \FloatBarrier
    \bibliographystyle{ACM-Reference-Format}
    \bibliography{main}
    \clearpage
    \appendix
    \section*{Appendix}
    
\begin{figure*}[t]
  \centering
  \setlength{\tabcolsep}{0.5pt}
  \renewcommand{\arraystretch}{0.3}

  \resizebox{\linewidth}{!}{%
  \begin{tabular}{@{}ccccccccc@{}}
    \toprule
    \multicolumn{2}{c}{\scriptsize \textbf{Source Textured Mesh}} &
    \multicolumn{1}{c}{\scriptsize \textbf{Input View}} &
    \multicolumn{2}{c}{\scriptsize \textbf{Ours}} &
    \multicolumn{2}{c}{\scriptsize \textbf{3DEditFormer}} &
    \multicolumn{2}{c}{\scriptsize \textbf{EditP23}} \\
    \cmidrule(r){1-2} \cmidrule(lr){3-3} \cmidrule(lr){4-5} \cmidrule(lr){6-7} \cmidrule(l){8-9}

    \includegraphics[width=0.109\linewidth, trim={113.1 53.57 136.9 101.19}, clip]{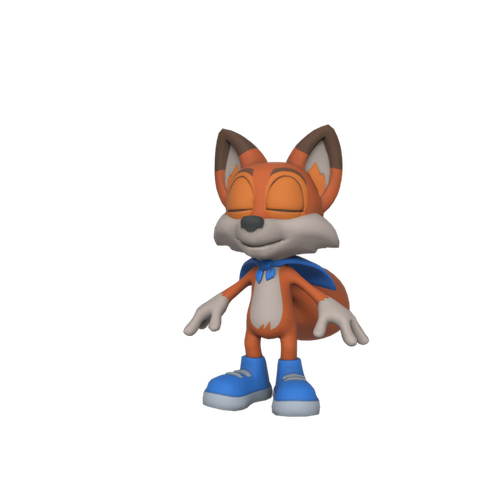} &
    \includegraphics[width=0.109\linewidth, trim={178.57 65.48 77.38 101.19}, clip]{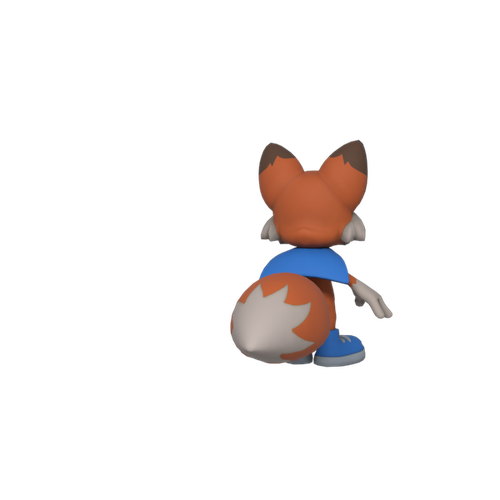} &
    \includegraphics[width=0.109\linewidth, trim={68.36 0 68.36 39.06}, clip]{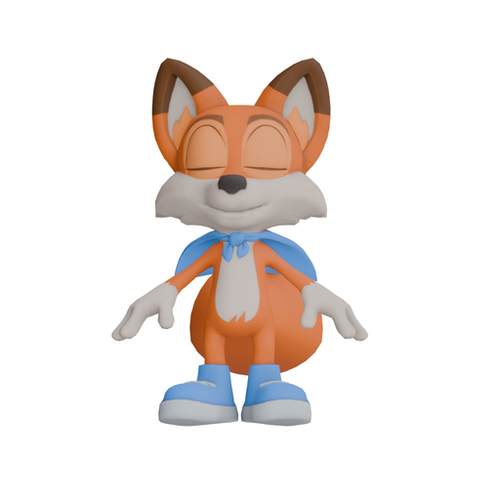} &
    \includegraphics[width=0.109\linewidth, trim={113.1 53.57 136.9 101.19}, clip]{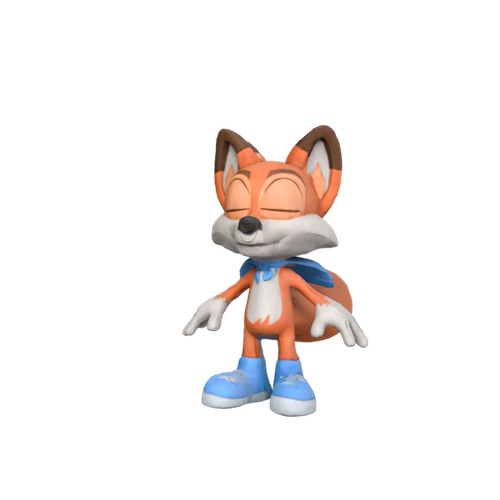} &
    \includegraphics[width=0.109\linewidth, trim={178.57 65.48 77.38 101.19}, clip]{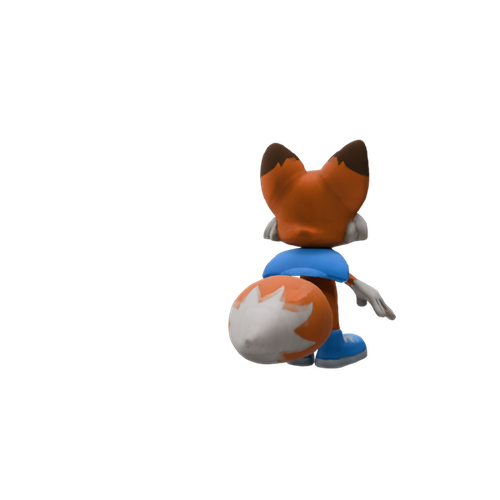} &
    \includegraphics[width=0.109\linewidth, trim={113.1 53.57 136.9 101.19}, clip]{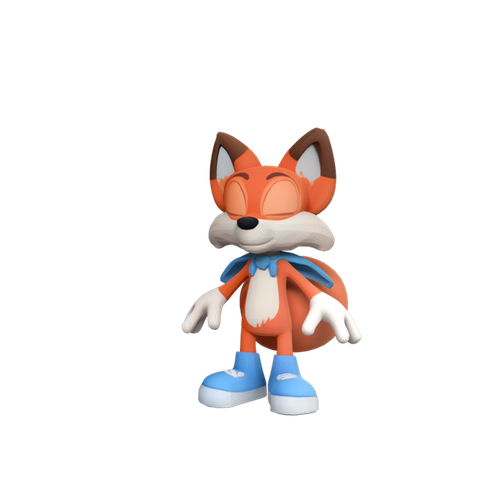} &
    \includegraphics[width=0.109\linewidth, trim={178.57 65.48 77.38 101.19}, clip]{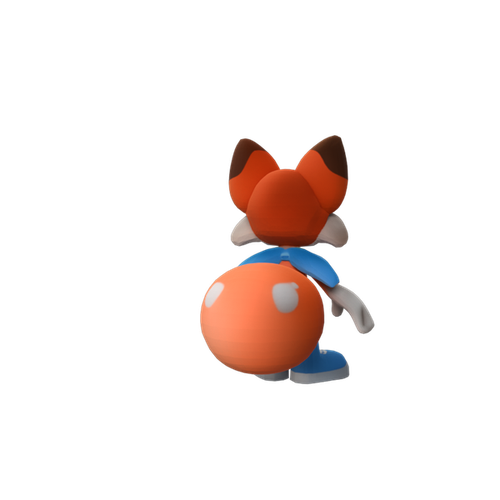} &
    \includegraphics[width=0.109\linewidth, trim={113.1 53.57 136.9 101.19}, clip]{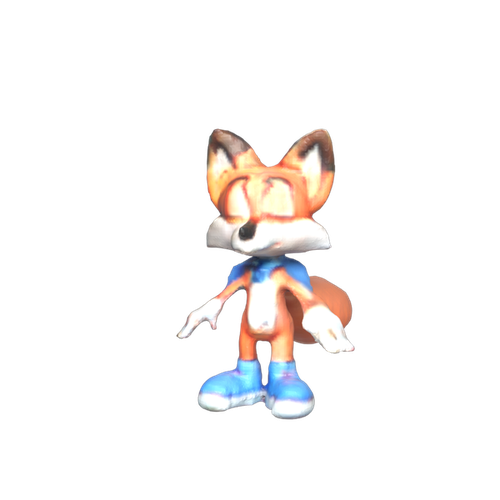} &
    \includegraphics[width=0.109\linewidth, trim={178.57 65.48 77.38 101.19}, clip]{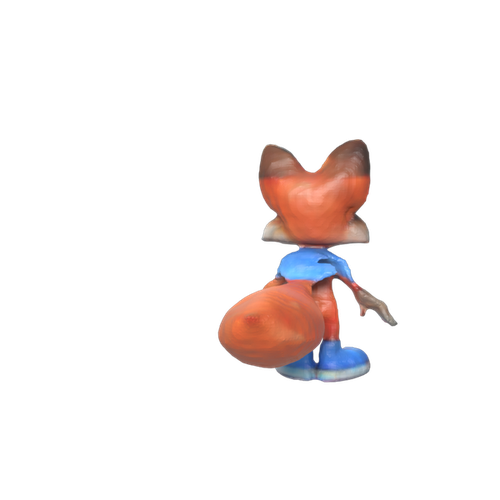} \\

    \includegraphics[width=0.109\linewidth, trim={113.1 53.57 136.9 101.19}, clip]{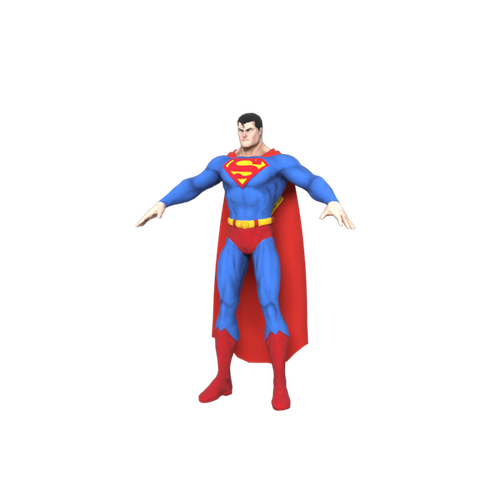} &
    \includegraphics[width=0.109\linewidth, trim={154.76 65.48 101.19 101.19}, clip]{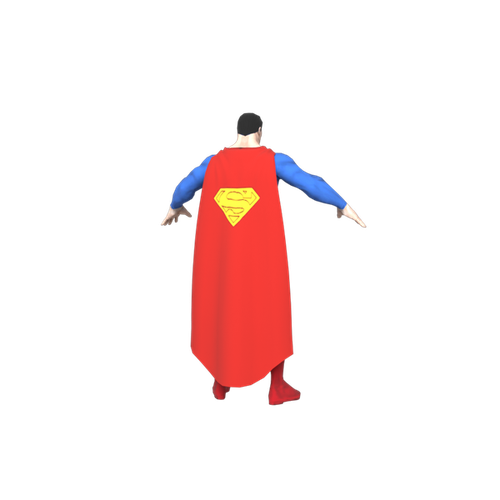} &
    \includegraphics[width=0.109\linewidth, trim={68.36 0 68.36 9.77}, clip]{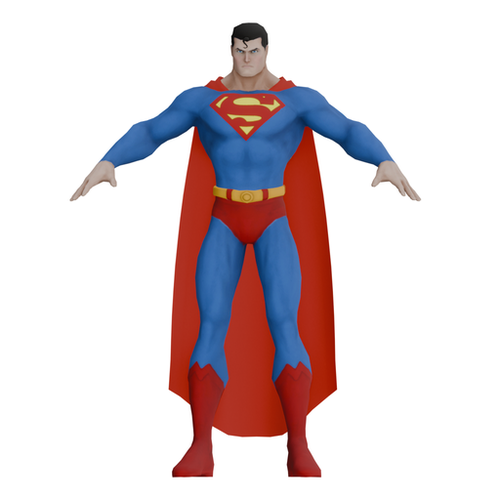} &
    \includegraphics[width=0.109\linewidth, trim={113.1 53.57 136.9 101.19}, clip]{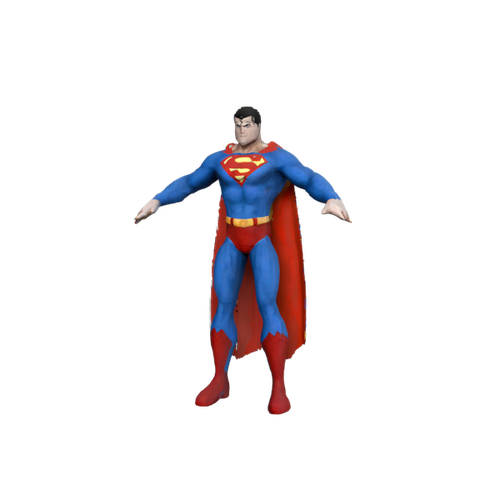} &
    \includegraphics[width=0.109\linewidth, trim={154.76 65.48 101.19 101.19}, clip]{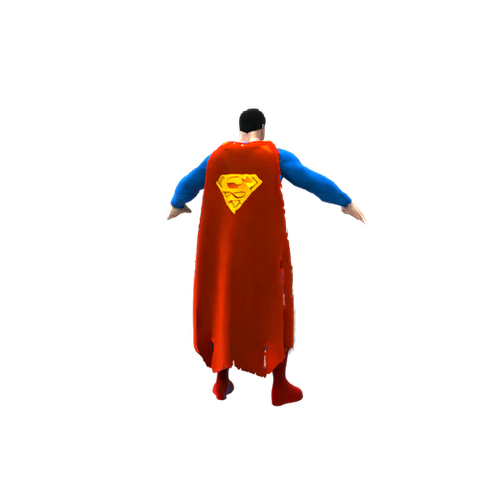} &
    \includegraphics[width=0.109\linewidth, trim={113.1 53.57 136.9 101.19}, clip]{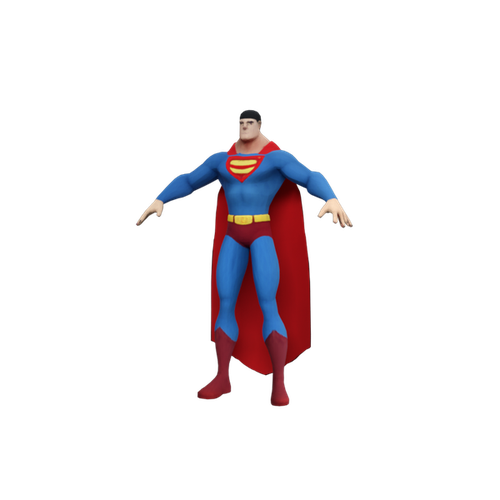} &
    \includegraphics[width=0.109\linewidth, trim={154.76 65.48 101.19 101.19}, clip]{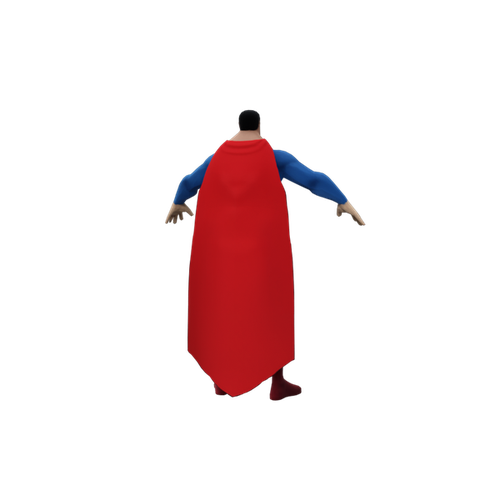} &
    \includegraphics[width=0.109\linewidth, trim={113.1 53.57 136.9 101.19}, clip]{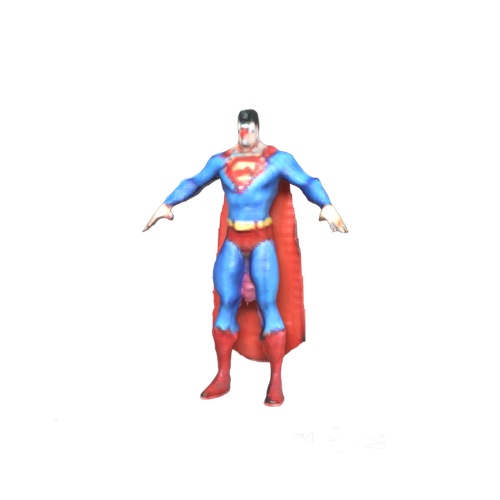} &
    \includegraphics[width=0.109\linewidth, trim={154.76 65.48 101.19 101.19}, clip]{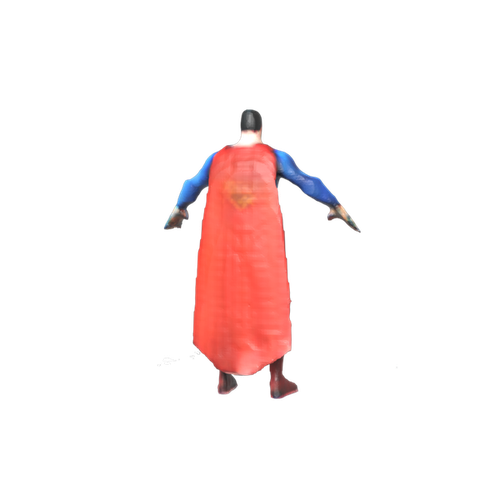} \\

    \includegraphics[width=0.109\linewidth, trim={113.1 65.48 136.9 101.19}, clip]{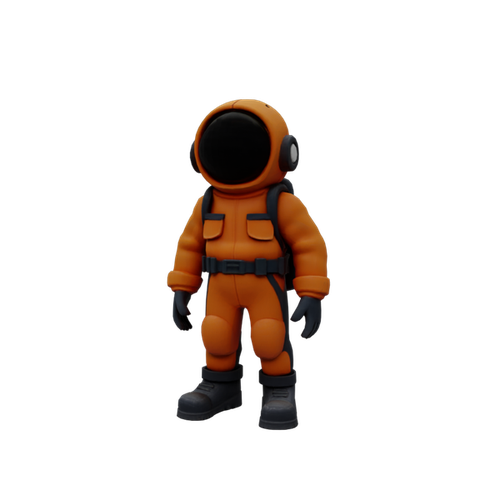} &
    \includegraphics[width=0.109\linewidth, trim={154.76 77.38 101.19 101.19}, clip]{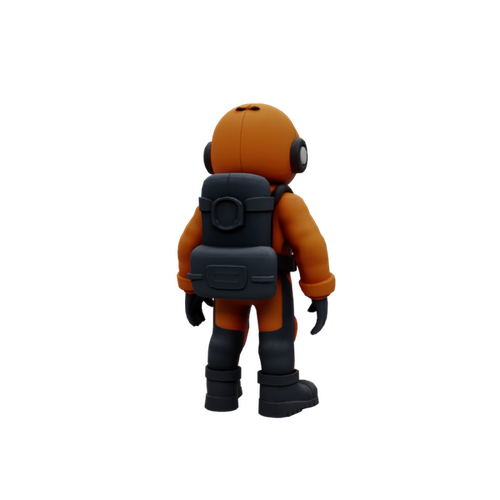} &
    \includegraphics[width=0.109\linewidth, trim={68.36 0 68.36 9.77}, clip]{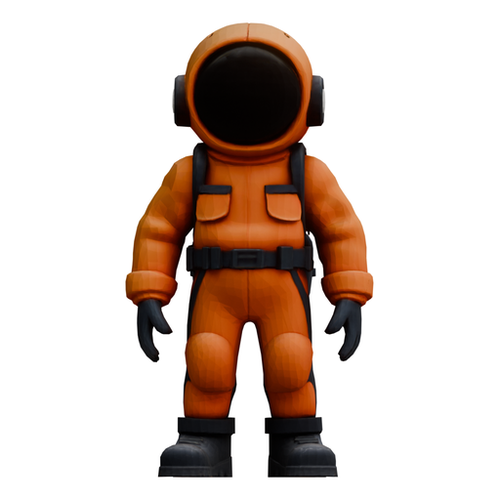} &
    \includegraphics[width=0.109\linewidth, trim={113.1 65.48 136.9 101.19}, clip]{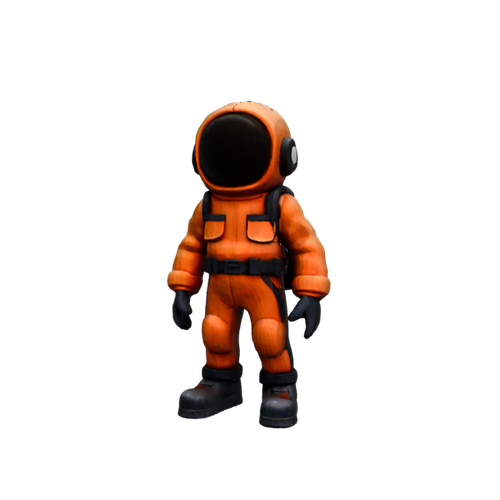} &
    \includegraphics[width=0.109\linewidth, trim={154.76 77.38 101.19 101.19}, clip]{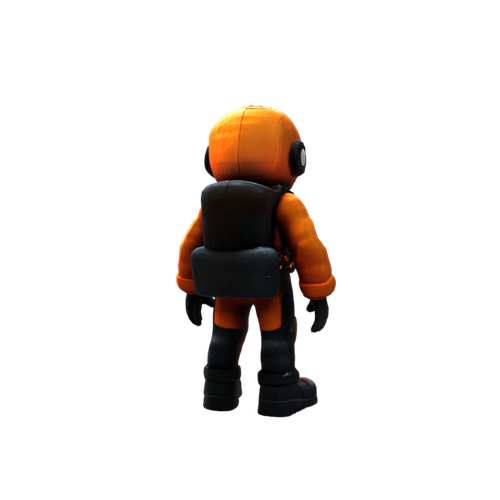} &
    \includegraphics[width=0.109\linewidth, trim={113.1 65.48 136.9 101.19}, clip]{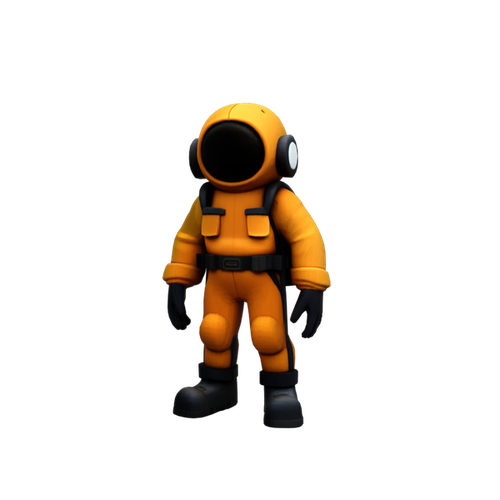} &
    \includegraphics[width=0.109\linewidth, trim={154.76 77.38 101.19 101.19}, clip]{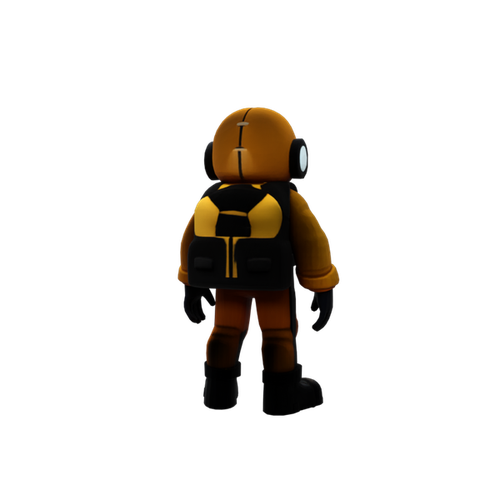} &
    \includegraphics[width=0.109\linewidth, trim={113.1 65.48 136.9 101.19}, clip]{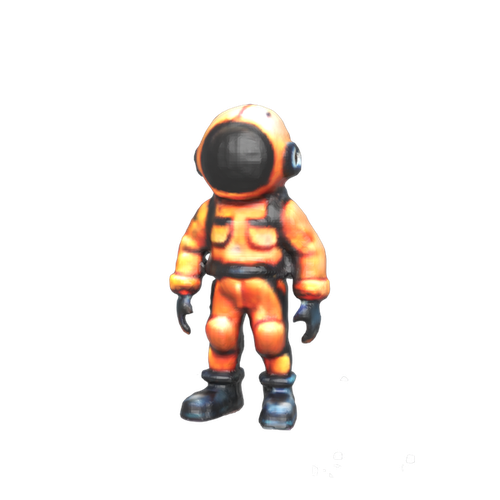} &
    \includegraphics[width=0.109\linewidth, trim={154.76 77.38 101.19 101.19}, clip]{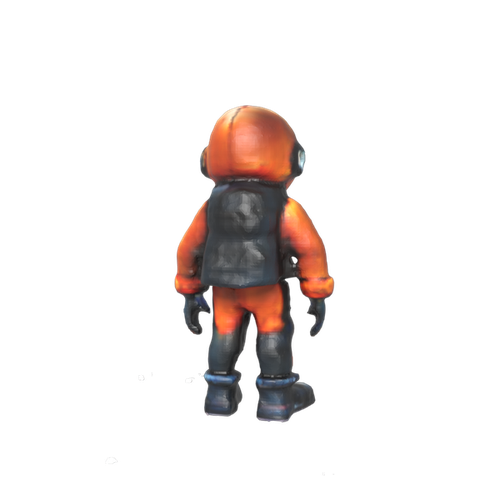} \\

    \includegraphics[width=0.109\linewidth, trim={113.1 53.57 136.9 101.19}, clip]{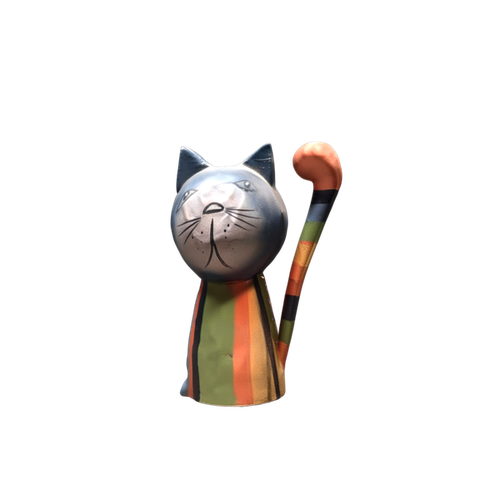} &
    \includegraphics[width=0.109\linewidth, trim={154.76 65.48 101.19 101.19}, clip]{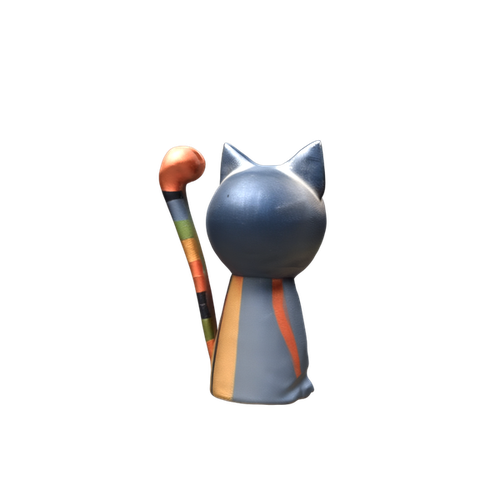} &
    \raisebox{6pt}{\includegraphics[width=0.109\linewidth, trim={48.83 0 48.83 9.77}, clip]{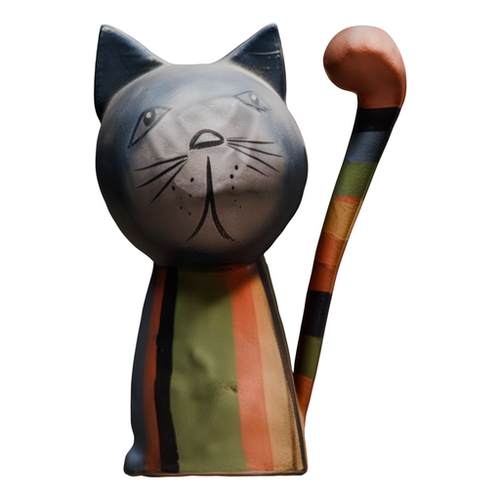}} &
    \includegraphics[width=0.109\linewidth, trim={113.1 53.57 136.9 101.19}, clip]{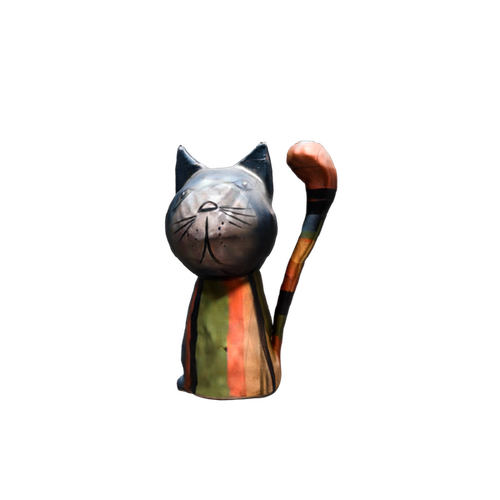} &
    \includegraphics[width=0.109\linewidth, trim={154.76 65.48 101.19 101.19}, clip]{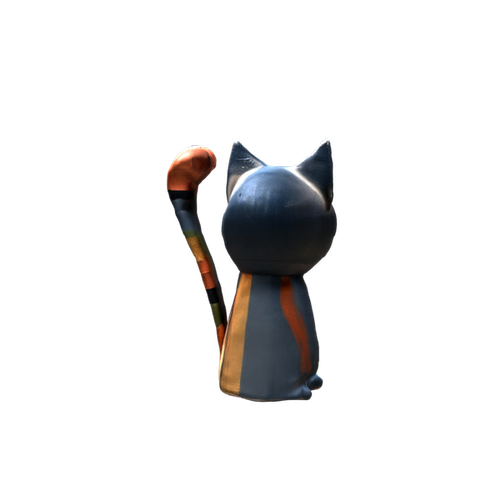} &
    \includegraphics[width=0.109\linewidth, trim={113.1 53.57 136.9 101.19}, clip]{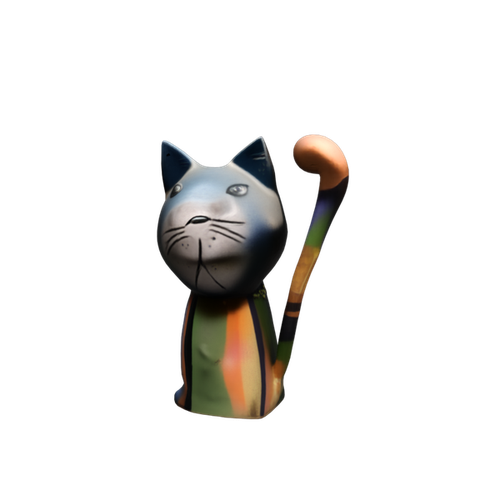} &
    \includegraphics[width=0.109\linewidth, trim={154.76 65.48 101.19 101.19}, clip]{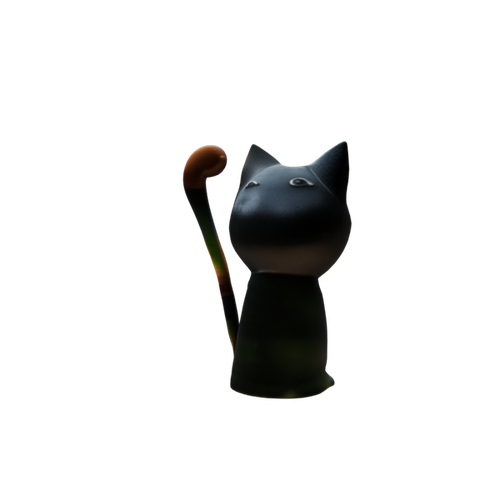} &
    \includegraphics[width=0.109\linewidth, trim={113.1 53.57 136.9 101.19}, clip]{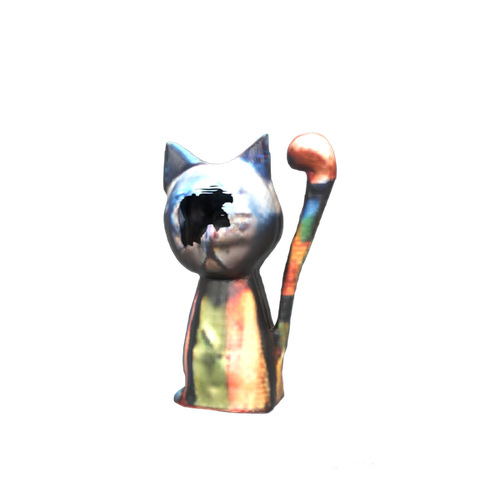} &
    \includegraphics[width=0.109\linewidth, trim={154.76 65.48 101.19 101.19}, clip]{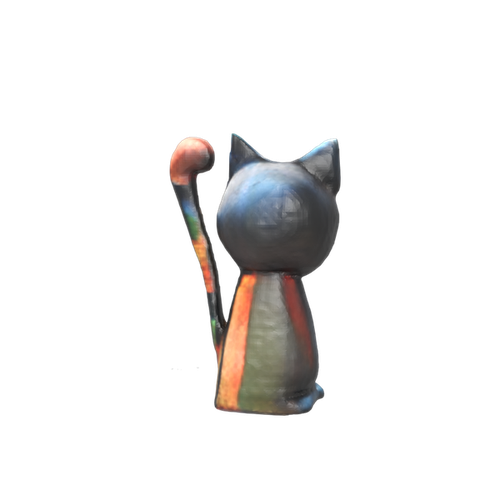} \\

    \includegraphics[width=0.109\linewidth, trim={113.1 53.57 136.9 101.19}, clip]{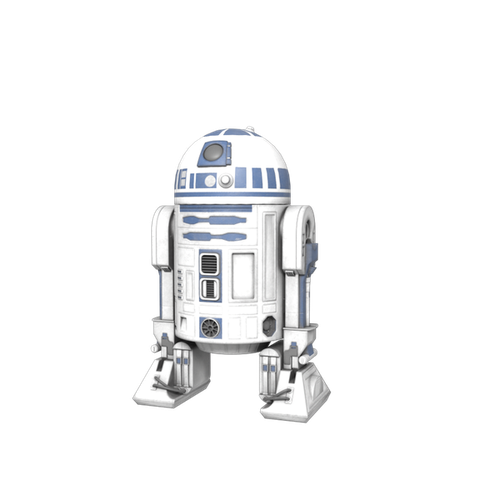} &
    \includegraphics[width=0.109\linewidth, trim={154.76 65.48 101.19 101.19}, clip]{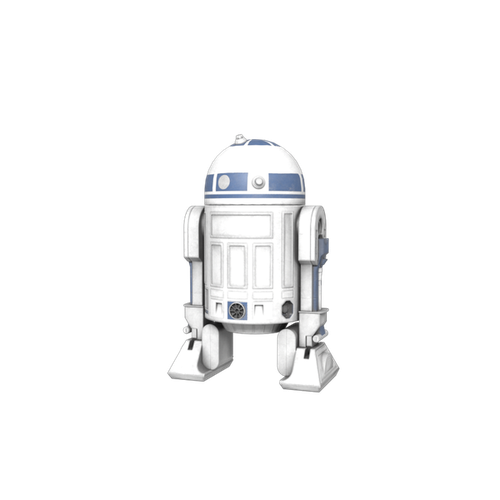} &
    \raisebox{6pt}{\includegraphics[width=0.109\linewidth, trim={48.83 0 48.83 9.77}, clip]{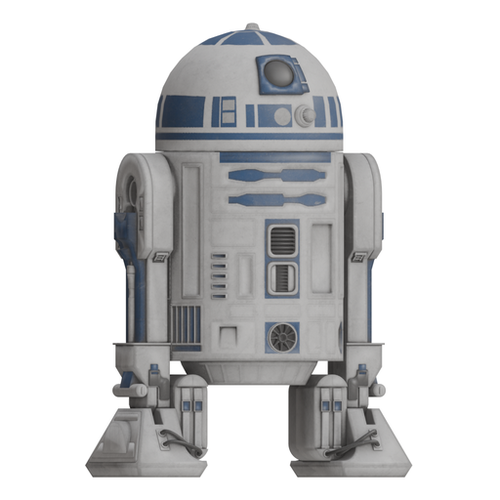}} &
    \includegraphics[width=0.109\linewidth, trim={113.1 53.57 136.9 101.19}, clip]{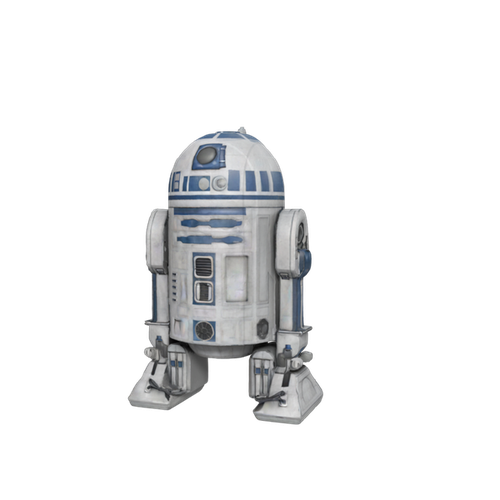} &
    \includegraphics[width=0.109\linewidth, trim={154.76 65.48 101.19 101.19}, clip]{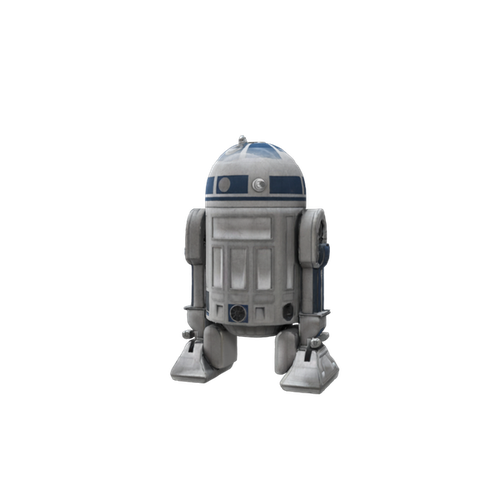} &
    \includegraphics[width=0.109\linewidth, trim={113.1 53.57 136.9 101.19}, clip]{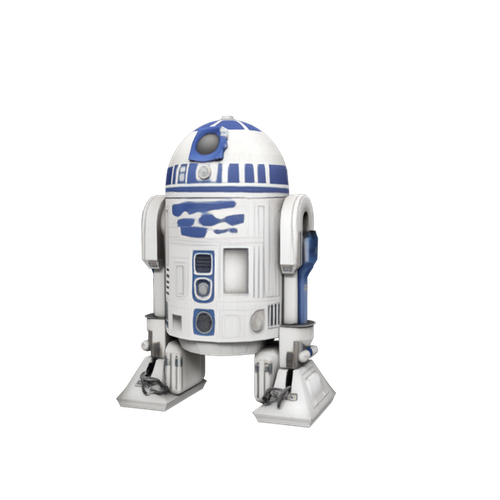} &
    \includegraphics[width=0.109\linewidth, trim={154.76 65.48 101.19 101.19}, clip]{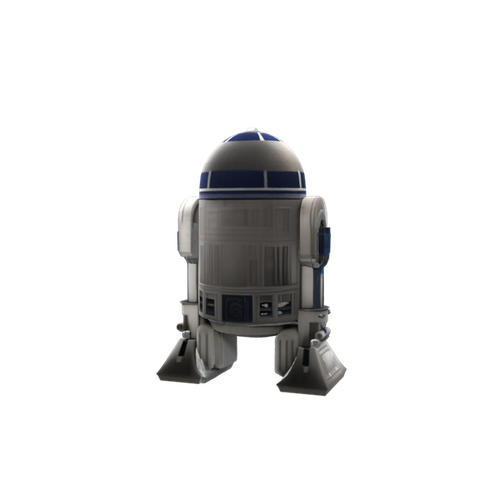} &
    \includegraphics[width=0.109\linewidth, trim={113.1 53.57 136.9 101.19}, clip]{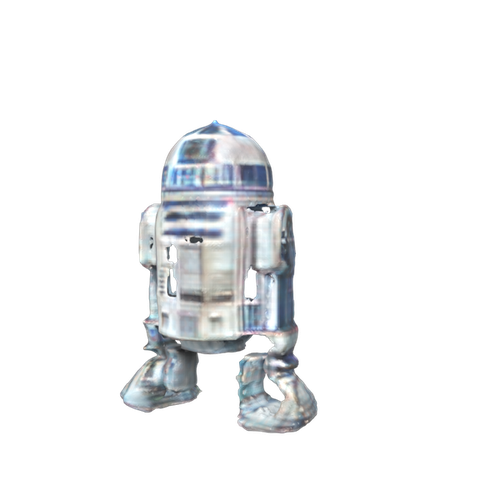} &
    \includegraphics[width=0.109\linewidth, trim={154.76 65.48 101.19 101.19}, clip]{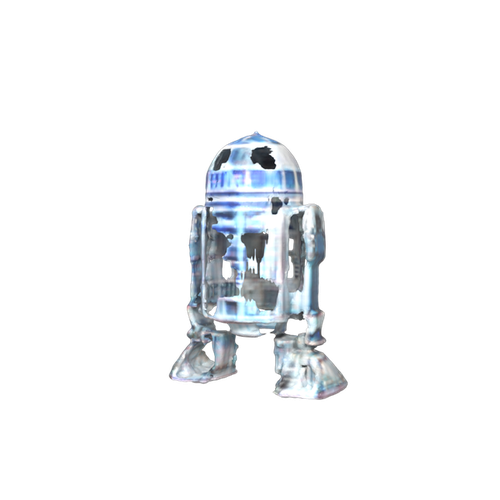} \\

    \bottomrule
  \end{tabular}%
  }%

  \caption{\textbf{Qualitative Reconstruction Comparisons on BenchUp.} We compare reconstruction quality of the \textbf{Zero-Edit} experiment against 3DEditFormer and EditP23. The first two columns show the source mesh (Front and Back views), followed by the Input View of the object provided as the editing condition. Our method (Cols 4--5) offers better reconstruction quality for both shape and texture compared to the baselines.}
  \label{fig:zero_edit_results}
\end{figure*}

This supplementary complements the main paper with additional implementation details, experiments, and results that further support our claims. The content here provides the specifics required to reproduce our results and a deeper look into the method.
\section{Method - Additional details}\label{appendix:method}
\subsection{Data Generation Pipeline}\label{appendix:data}

In this section, we describe our data curation process in detail.
\paragraph{Parts Data Samples}
To construct the Parts samples, we started from the filtered mesh list used to train the Step1X-3D \cite{li2025step1x} texture model. This list contains 30k filtered assets from Objaverse \cite{deitke2022objaverseuniverseannotated3d}, with high-quality geometry and non-uniform textures. We selected 6,870 distinct assets, each containing at least two geometry nodes (i.e., already segmented into parts). For every mesh that passed these filter, we generated multiple variants by progressively removing parts; each intermediate version can serve as either a source or a target mesh in our pipeline.
\paragraph{DFM Data Samples}
For the DFM samples, we downloaded assets from the Objaverse Animation subset introduced by \cite{li2025puppetmasterscalinginteractivevideo}. We restricted this set to assets that also appear in the Step1X-3D Texture filtered asset list, resulting in 560 animated sequences. For each asset, we then heuristically selected the three most significant keyframes based on their bounding boxes, yielding three mesh variants per asset. Each variant can serve as either a source or a target during training.
\paragraph{Rendering}
Following the Step1X-3D rendering protocol, we rendered each mesh variant from six orthogonal viewpoints for texture-model fine-tuning, and additionally generated 20 views with random azimuth in $[-70,70]$ and elevation in $[-15,30]$. During training of both the geometry and texture models, we sample either one of these random renders or the front orthogonal view to serve as the image prompt specifying the desired edit. To improve robustness to occluded regions, we sample the front view with higher probability (20\%) than the other renders.
\paragraph{Sampling}
We follow the backbone sampling strategy, originally proposed by \cite{chen2025dorasamplingbenchmarking3d}. Specifically, we sample 64,834 points from each mesh variant: the first half are sampled uniformly over the surface, while the second half are "sharp" samples taken near edges whose adjacent face angles are below a chosen threshold, capturing delicate structures and fine-grained details. 

\paragraph{Producing Shape Latent Representations}
As part of data preparation, we run each asset variant through the pretrained Step1X-3D shape encoder and store the posterior distributions. During training, we sample latent codes from these cached posteriors, eliminating the need to repeatedly run VAE inference on the source shape at every training step.

\subsection{Geometry Editing Pipeline Data Encoders}

This section provides additional details on the Shape and Image encoders used in the geometry editing pipeline.

\paragraph{Image Encoder}
The image encoder processes the input image using both CLIP and DINOv2. The tokens produced by the two encoders are projected to a shared dimensionality using a linear layer, and the resulting embeddings are concatenated into a single sequence that serves as the image conditioning signal. For additional details, please refer to the Step1X-3D technical report.

\paragraph{Shape Encoder}
The shape encoder corresponds to the pretrained Shape VAE encoder, which is trained prior to the backbone DiT model. It is based on the Dora VAE architecture proposed in~\cite{chen2025dorasamplingbenchmarking3d}. The encoder takes a sampled mesh represented as a point cloud, where half of the points are sampled uniformly and the remaining half are sampled along sharp edges (see the sampling procedure in \cref{appendix:data}). To better capture geometric details from edge-aware sampling, the encoder employs a dual cross-attention architecture that enhances the encoding of points sampled near sharp edges. Additional implementation details can be found in the original paper.
\section{Experiments - Additional details}\label{appendix:eval}
\subsection{Quantitative Results}

\paragraph{Reconstruction Fidelity}
To evaluate the 3D reconstruction fidelity of \algoname, we conducted a Zero-Edit experiment. For each object in our benchmark, we rendered 20 isomorphic views of the source shape, with elevations randomly sampled from ($-$15,30) and azimuths from ($-$30,30). Each rendered view was used as the edit conditioning image and paired with the corresponding source shape in the evaluation set. We then evaluated our model and the baselines on this setup, reporting SSIM, LPIPS, CLIP-I, and DINO-I for both visible and occluded regions. The quantitative results are presented in \cref{tab:zero_edit}. The results demonstrate that \algoname achieves high-fidelity reconstruction compared to the baselines, with minimal drift in both visible and occluded regions. We also provide qualitative comparison to the baselines in \cref{fig:zero_edit_results}.
\begin{table}[H]
\small
\centering
\caption{\textbf{Reconstruction Fidelity Comparison on BenchUp.} We evaluate reconstruction fidelity by conducting a \textbf{Zero-Edit} experiment and measuring \textbf{Visible Regions} and \textbf{Occluded Regions} Fidelity. Bold indicates best performance.}
\label{tab:zero_edit}
\setlength{\tabcolsep}{1.2mm} %
\resizebox{\linewidth}{!}{
\begin{tabular}{lcccccccc}
    \toprule
    \multirow{2}{*}{Method} & \multicolumn{4}{c}{Visible Region Fid.} & \multicolumn{4}{c}{Occluded Region Fid.} \\
    \cmidrule(lr){2-5} \cmidrule(lr){6-9}
     & SSIM$\uparrow$ & LPIPS$\downarrow$ & CLIP-I$\uparrow$ & DINO-I$\uparrow$ & SSIM$\uparrow$ & LPIPS$\downarrow$ & CLIP-I$\uparrow$ & DINO-I$\uparrow$ \\
    \midrule
    3DEditFormer & 0.780 & 0.221 & 0.929 & 0.890 & 0.784 & 0.221 & 0.904 & 0.798 \\
    EditP23 & 0.800 & 0.216 & 0.921 & 0.867 & \textbf{0.824} & 0.213 & 0.894 & 0.802 \\
    Ours & \textbf{0.821} & \textbf{0.124} & \textbf{0.951} & \textbf{0.928} & 0.793 & \textbf{0.135} & \textbf{0.934} & \textbf{0.874} \\
    \bottomrule
\end{tabular}
}
\end{table}

\subsection{Qualitative Results}

\ifarxiv\else
\paragraph{Failure Cases Analysis}
As noted in \ifappendix \cref{sec:experiments} \else the main paper\fi, \algoname exhibits degraded performance on assets with fine-grained geometric details or non-object-centric compositions.
Since all shapes are decoded through the pretrained Step1X-3D VAE, the output quality is bounded by its representational capacity; for assets with thin structures or intricate detail, the VAE encoder--decoder itself introduces smoothing and detail loss.
In the original image-to-3D setting the backbone can partially compensate by leveraging the input image as a strong appearance prior, re-synthesizing details that the latent code fails to preserve.
Our editing formulation, however, is conditioned on the source shape latent and trained to preserve the input geometry except where the edit condition specifies a change, making it more susceptible to representational drift in the latent space.
\cref{fig:failure_Cases} illustrates this with several representative examples: comparing each source shape with its VAE-only reconstruction confirms that much of the observed degradation is attributable to the limited expressiveness of the shape latent space rather than to the editing process itself.

\fi
\ifarxiv
    \ifarxiv
\begin{figure*}[!htbp]
\else
\begin{figure*}[t]
\fi
  \centering
  \setlength{\tabcolsep}{1pt} 
  \renewcommand{\arraystretch}{0.6}

\resizebox{0.80\linewidth}{!}{%
\begin{tabular}{@{}cc c cc@{}}
    \toprule
    \multicolumn{2}{c}{\scriptsize \textbf{Source Textured Mesh}} &
    \multicolumn{1}{c}{\scriptsize \textbf{Edit Condition}} &
    \multicolumn{2}{c}{\scriptsize \textbf{Edited}} \\
    
    \cmidrule(r){1-2} \cmidrule(lr){3-3} \cmidrule(lr){4-5}

    \includegraphics[width=0.16\linewidth, trim={17.86 17.86 17.86 17.86}, clip]{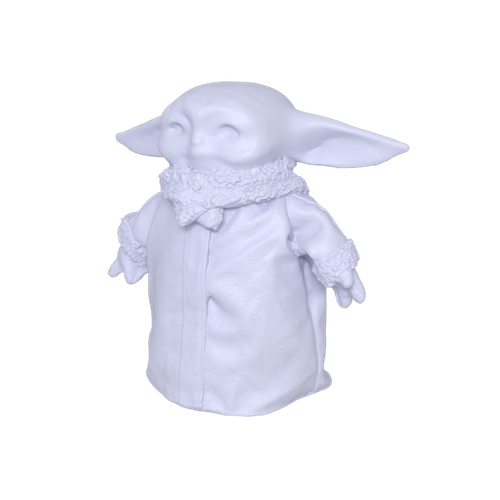} & \includegraphics[width=0.16\linewidth, trim={17.86 17.86 17.86 17.86}, clip]{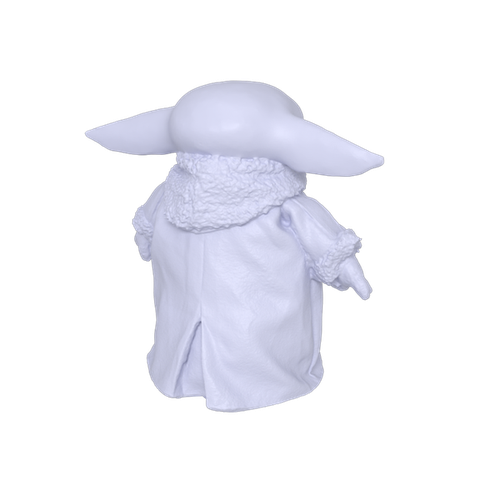} &
    \includegraphics[width=0.16\linewidth, trim={14.65 14.65 14.65 14.65}, clip]{images/qualitative_results/Edit_images/grogu_0008.png} &
    \includegraphics[width=0.16\linewidth, trim={17.86 17.86 17.86 17.86}, clip]{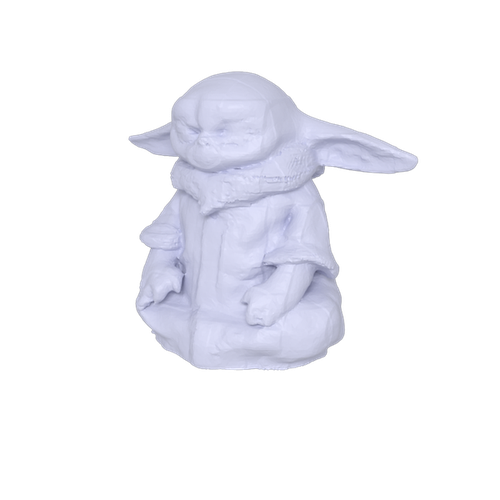} & \includegraphics[width=0.16\linewidth, trim={17.86 17.86 17.86 17.86}, clip]{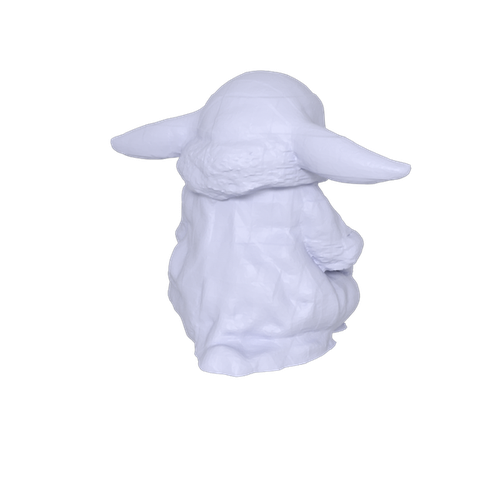} \\ [-5pt]

    \includegraphics[width=0.16\linewidth, trim={17.86 17.86 17.86 17.86}, clip]{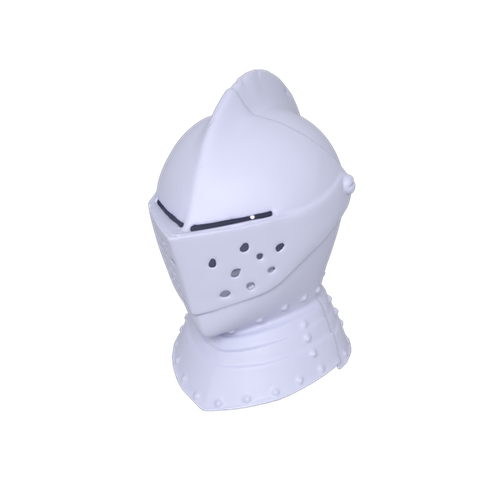} & \includegraphics[width=0.16\linewidth, trim={17.86 17.86 17.86 17.86}, clip]{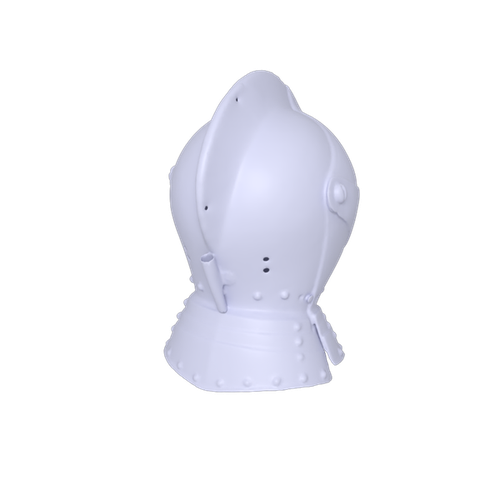} &
    \raisebox{6pt}{\includegraphics[width=0.14\linewidth, trim={14.65 14.65 14.65 14.65}, clip]{images/qualitative_results/Edit_images/helmet_0024.png}} &
    \includegraphics[width=0.16\linewidth, trim={17.86 17.86 17.86 17.86}, clip]{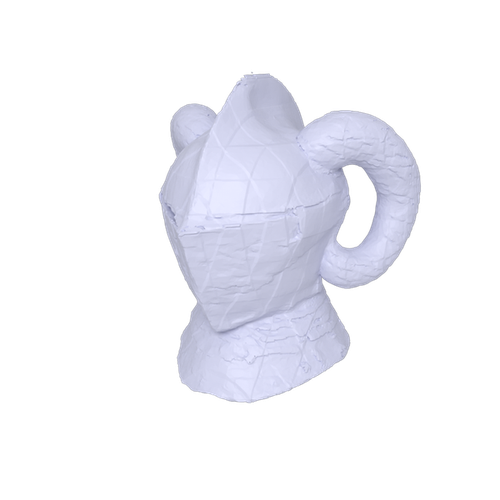} & \includegraphics[width=0.16\linewidth, trim={17.86 17.86 17.86 17.86}, clip]{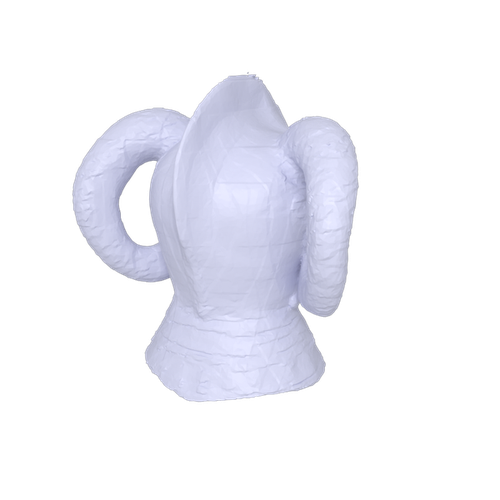} \\ [-5pt]

    \includegraphics[width=0.16\linewidth, trim={17.86 17.86 17.86 17.86}, clip]{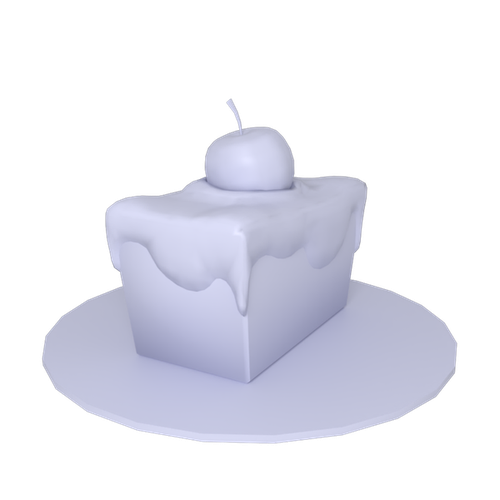} & \includegraphics[width=0.16\linewidth, trim={17.86 17.86 17.86 17.86}, clip]{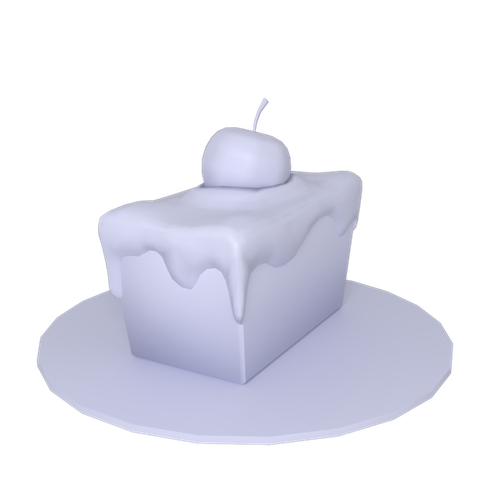} &
    \includegraphics[width=0.16\linewidth, trim={14.65 14.65 14.65 14.65}, clip]{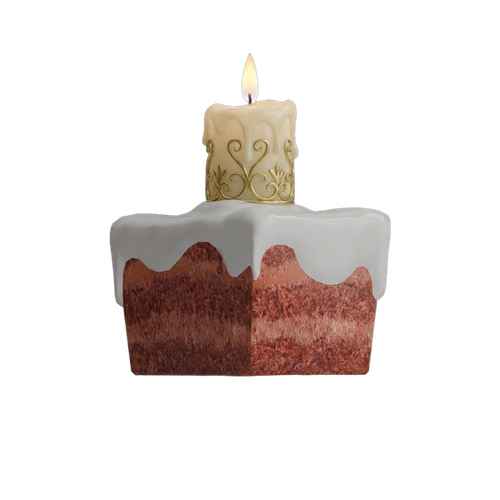} &
    \includegraphics[width=0.16\linewidth, trim={17.86 17.86 17.86 17.86}, clip]{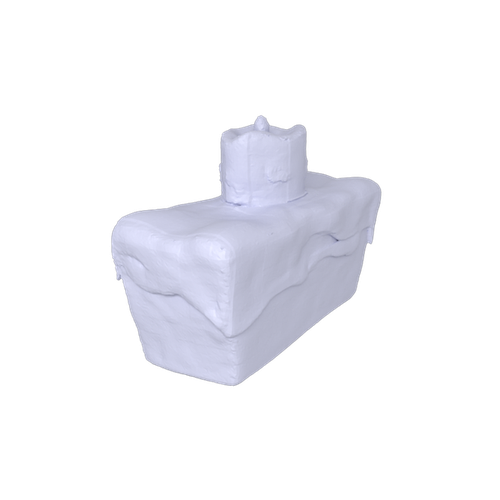} & \includegraphics[width=0.16\linewidth, trim={17.86 17.86 17.86 17.86}, clip]{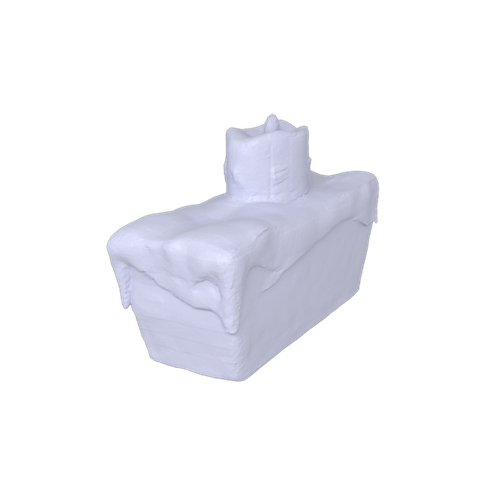} \\ [-5pt]

    \includegraphics[width=0.16\linewidth, trim={17.86 17.86 17.86 17.86}, clip]{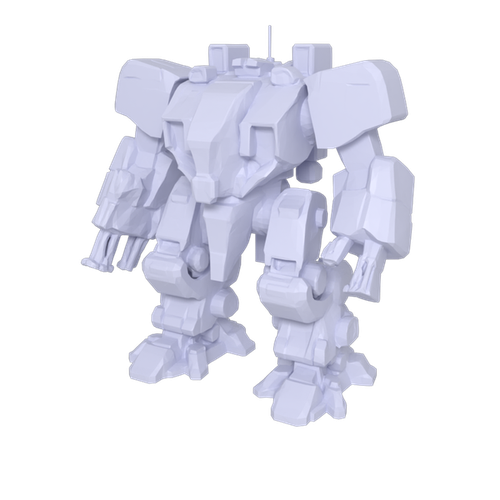} & \includegraphics[width=0.16\linewidth, trim={17.86 17.86 17.86 17.86}, clip]{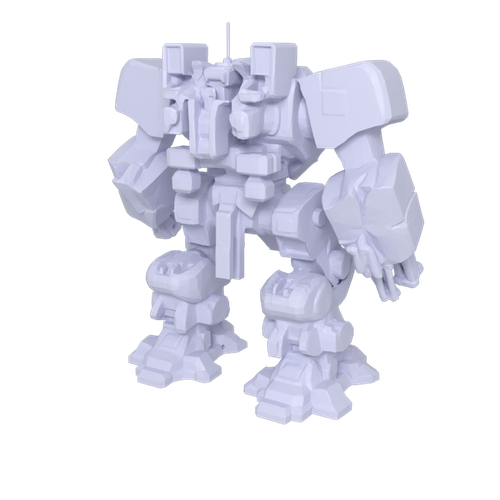} &
    \raisebox{9pt}{\includegraphics[width=0.14\linewidth, trim={14.65 14.65 14.65 14.65}, clip]{images/qualitative_results/Edit_images/mech_0022.png}} &
    \includegraphics[width=0.16\linewidth, trim={17.86 17.86 17.86 17.86}, clip]{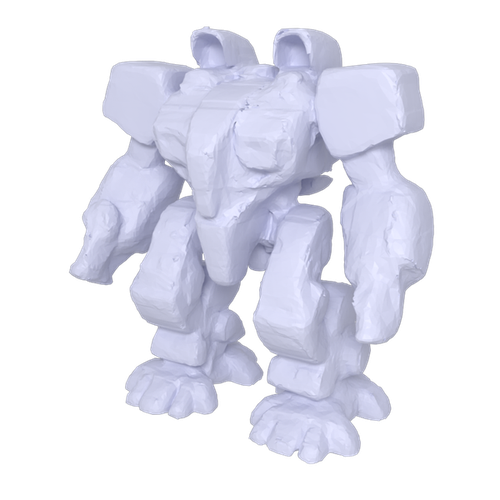} & \includegraphics[width=0.16\linewidth, trim={17.86 17.86 17.86 17.86}, clip]{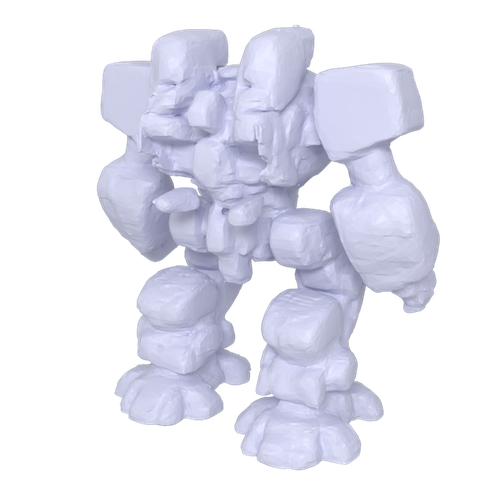} \\ [-5pt]

    \includegraphics[width=0.16\linewidth, trim={17.86 17.86 17.86 17.86}, clip]{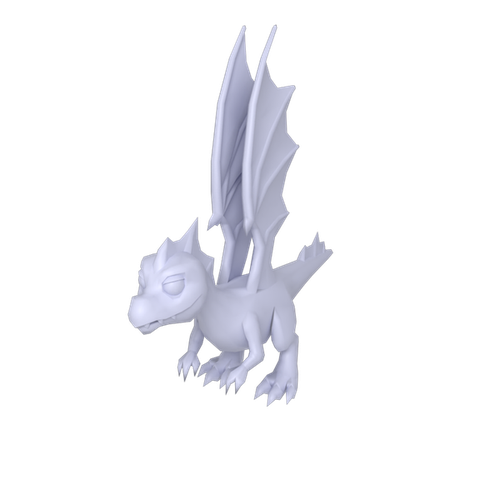} & \includegraphics[width=0.16\linewidth, trim={17.86 17.86 17.86 17.86}, clip]{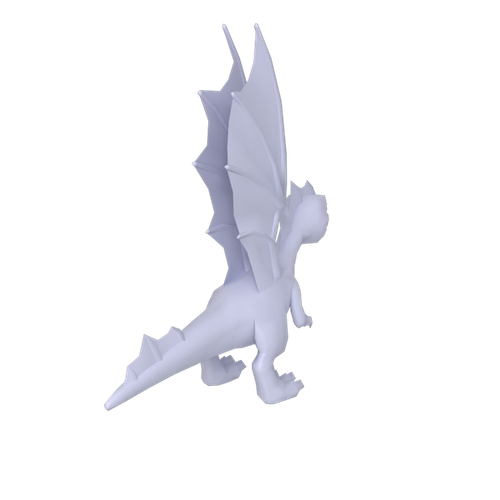} &
    \includegraphics[width=0.16\linewidth, trim={14.65 14.65 14.65 14.65}, clip]{images/qualitative_results/Edit_images/red_dragon_0012.png} &
    \includegraphics[width=0.16\linewidth, trim={17.86 17.86 17.86 17.86}, clip]{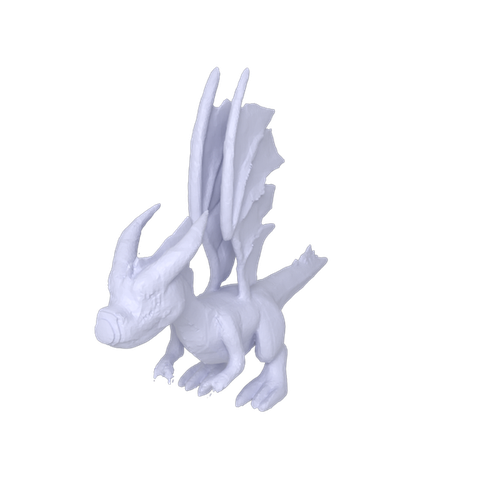} & \includegraphics[width=0.16\linewidth, trim={17.86 17.86 17.86 17.86}, clip]{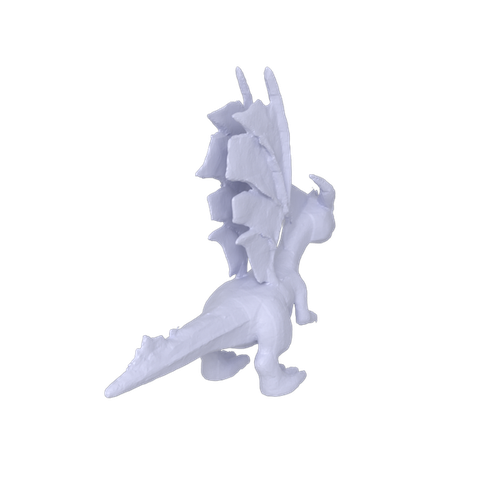} \\ [-5pt]

    \includegraphics[width=0.16\linewidth, trim={17.86 17.86 17.86 17.86}, clip]{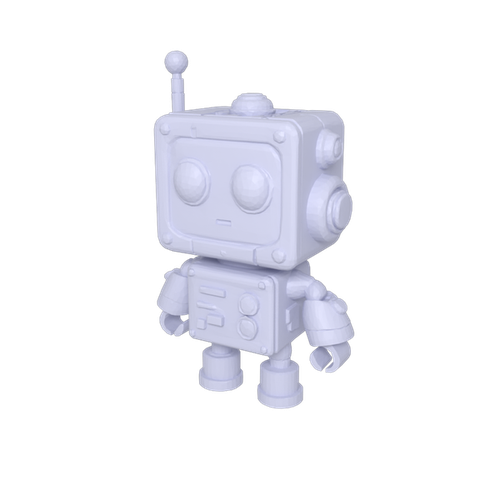} & \includegraphics[width=0.16\linewidth, trim={17.86 17.86 17.86 17.86}, clip]{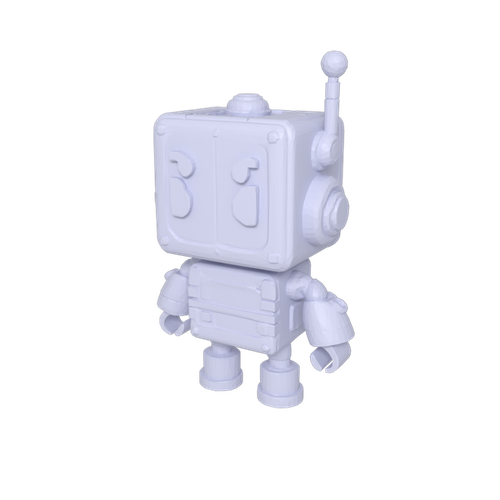} &
    \raisebox{6pt}{\includegraphics[width=0.14\linewidth, trim={14.65 14.65 14.65 14.65}, clip]{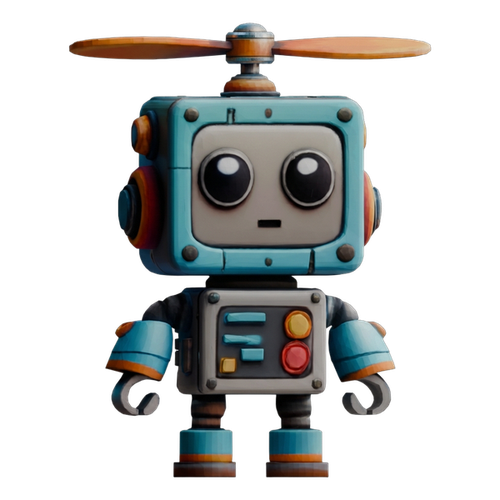}} &
    \includegraphics[width=0.16\linewidth, trim={17.86 17.86 17.86 17.86}, clip]{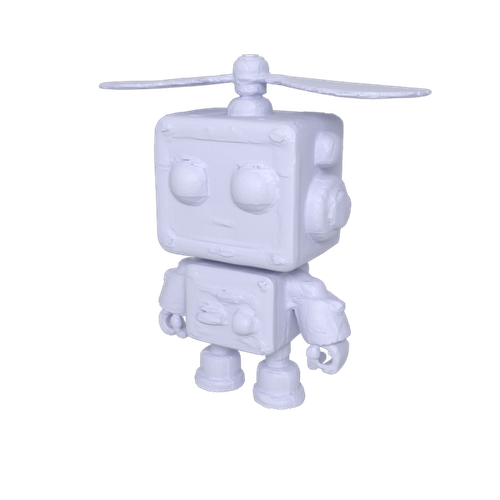} & \includegraphics[width=0.16\linewidth, trim={17.86 17.86 17.86 17.86}, clip]{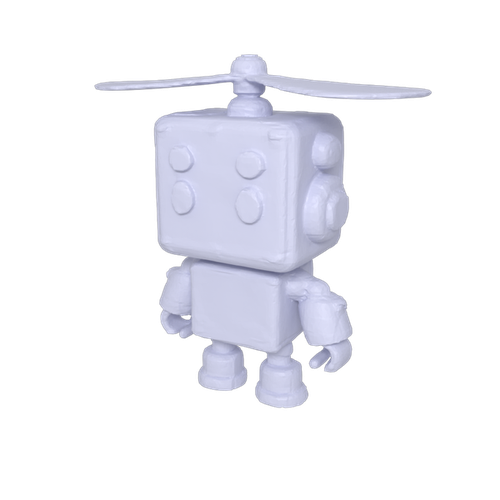} \\

    \includegraphics[width=0.16\linewidth, trim={17.86 17.86 17.86 17.86}, clip]{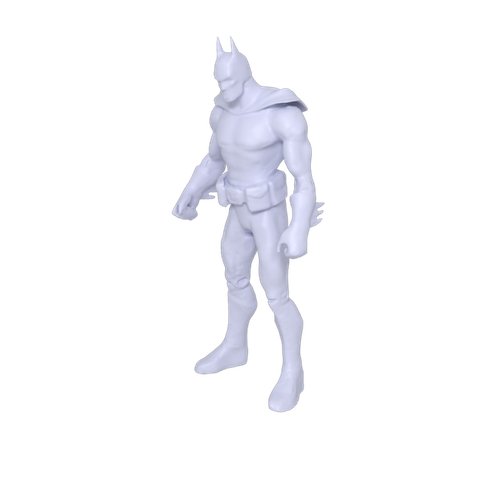} & \includegraphics[width=0.16\linewidth, trim={17.86 17.86 17.86 17.86}, clip]{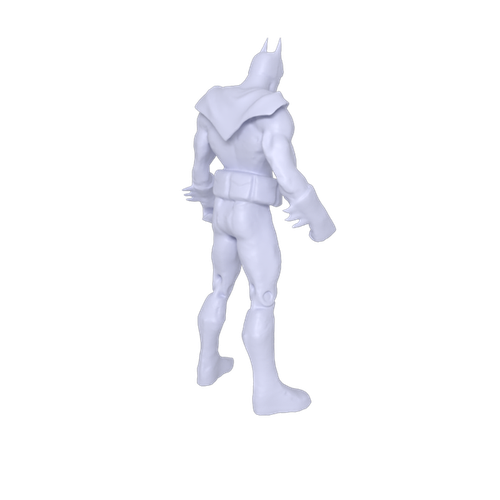} &
    \raisebox{6pt}{\includegraphics[width=0.14\linewidth, trim={14.65 14.65 14.65 14.65}, clip]{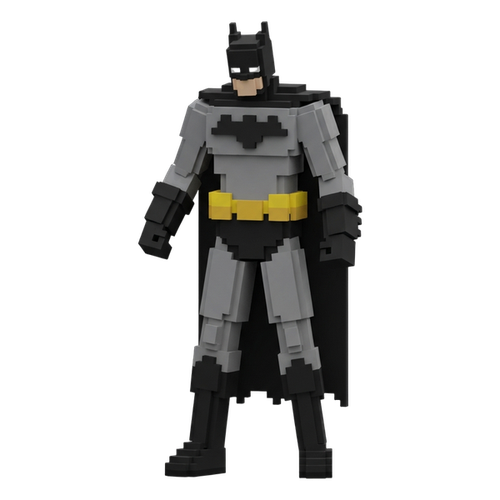}} &
    \includegraphics[width=0.16\linewidth, trim={17.86 17.86 17.86 17.86}, clip]{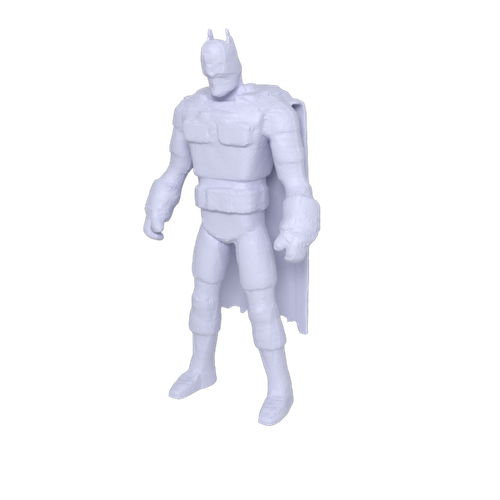} & \includegraphics[width=0.16\linewidth, trim={17.86 17.86 17.86 17.86}, clip]{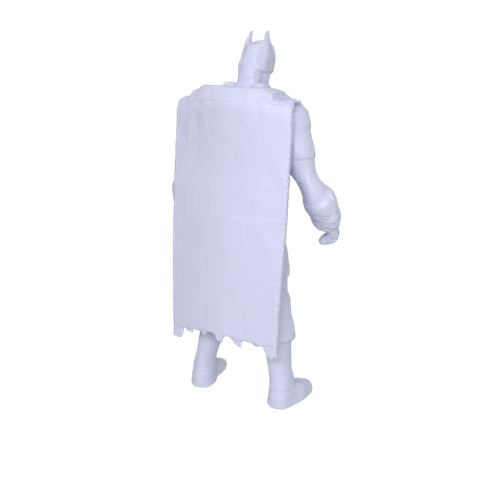} \\ [-17pt]

    \includegraphics[width=0.16\linewidth, trim={17.86 17.86 17.86 17.86}, clip]{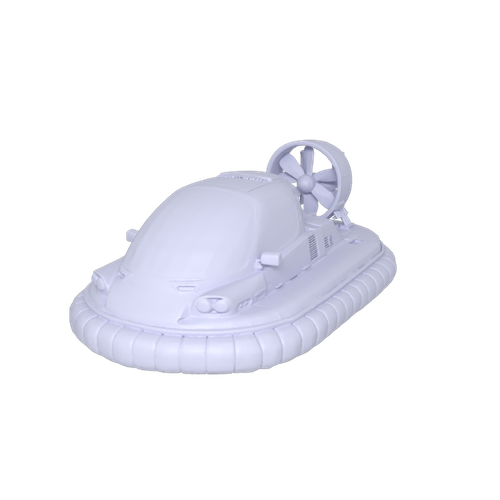} & \includegraphics[width=0.16\linewidth, trim={17.86 17.86 17.86 17.86}, clip]{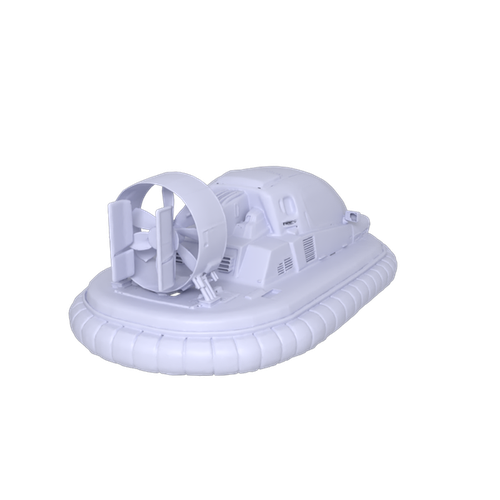} &
    \raisebox{-11pt}{\includegraphics[width=0.19\linewidth, trim={14.65 14.65 14.65 14.65}, clip]{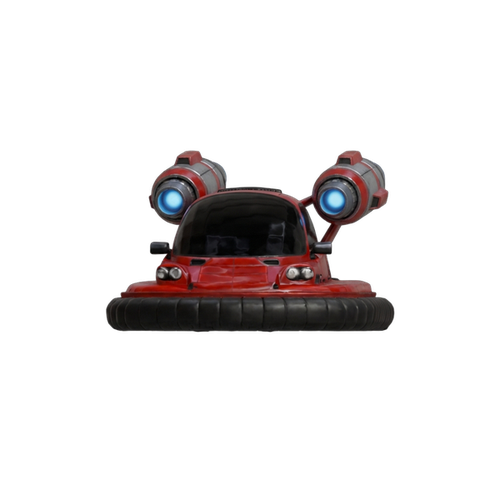}} &
    \includegraphics[width=0.16\linewidth, trim={17.86 17.86 17.86 17.86}, clip]{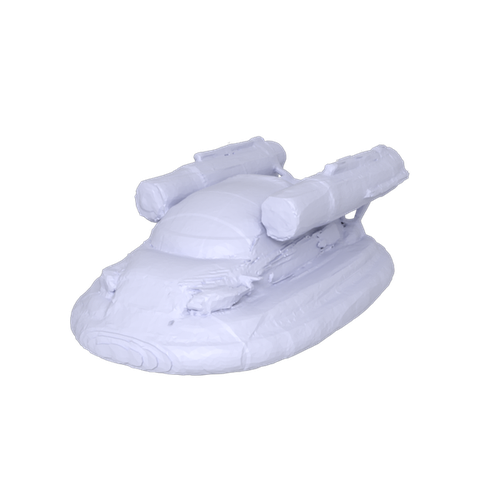} & \includegraphics[width=0.16\linewidth, trim={17.86 17.86 17.86 17.86}, clip]{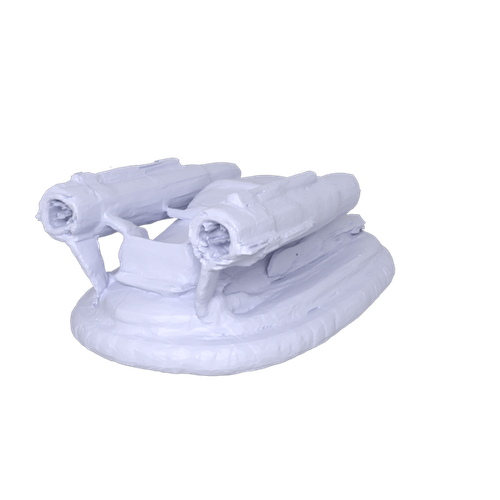} \\ [-8pt]

    \bottomrule
  \end{tabular}
  }%

  \caption{\textbf{Untextured Qualitative Results.} Columns 1--2 show the \textbf{Source Mesh} (front/back). Columns 3--5 present \textbf{Edit Condition} and the \textbf{edited mesh }(front/back).}
  \label{fig:untextured_results}
\end{figure*}

\fi
\paragraph{Multi-step Editing}
To emulate a real-world 3D editing workflow, we perform multi-step editing, where image-based edits are applied sequentially, gradually applying local and global edits to the output of the previous step.
At editing step $i$, we provide the geometry pipeline with the output shape of the previous step and a desired edit image describing the intended transition from step $i-1$ to step $i$. 
To avoid color drift caused by multiple iterations of the texture editing pipeline, we feed each stage with the original shape's multiview renders instead of those from the previous stage. The remaining control signals (normal and position renders) are derived from the previous step's output shape.
Results are shown in \cref{fig:multistep}.
\ifarxiv
\begin{figure*}[!htbp]
\else
\begin{figure*}[t]
\fi
  \centering
  \setlength{\tabcolsep}{2pt}
  \renewcommand{\arraystretch}{0.6}

  \setlength{\arrayrulewidth}{0.6pt}%
  \resizebox{0.80\linewidth}{!}{%
  \begin{tabular}[t]{@{}c@{}}
    \hline
    \scriptsize \shortstack{\textbf{Input Textured Mesh}} \\
    \hline
    \includegraphics[width=0.19\linewidth, trim={17.86 35.71 17.86 17.86}, clip]{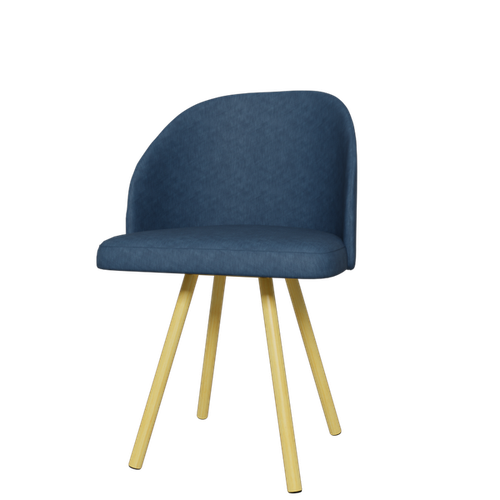} \\
    \includegraphics[width=0.19\linewidth, trim={17.86 17.86 17.86 17.86}, clip]{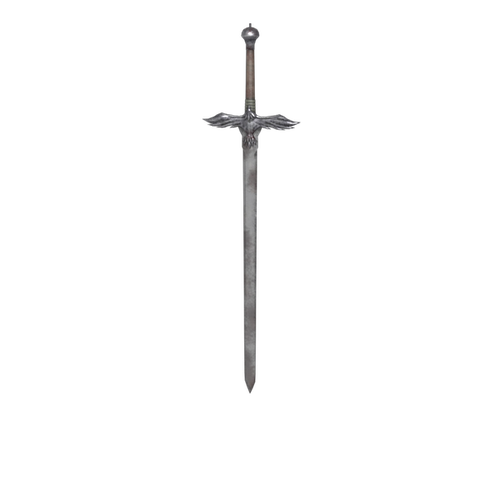} \\
    \includegraphics[width=0.19\linewidth, trim={17.86 17.86 17.86 29.76}, clip]{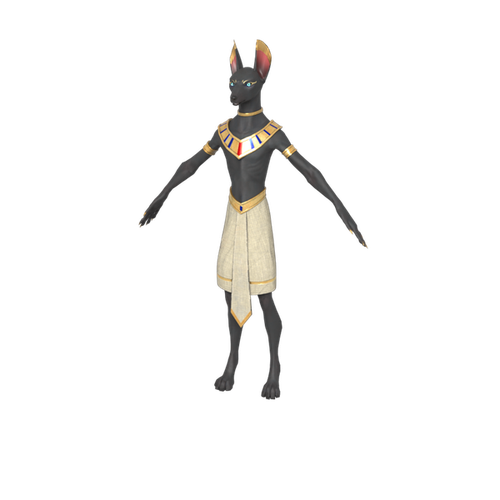} \\
    \hline
  \end{tabular}%
  \vrule width 0.6pt%
  \begin{tabular}[t]{@{}cc@{}}
    \hline
    \scriptsize \shortstack{\textbf{Edit Condition \#1}} &
    \scriptsize \shortstack{\textbf{Output Textured Mesh}} \\
    \hline
    \includegraphics[width=0.16\linewidth, trim={24.41 39.06 24.41 24.41}, clip]{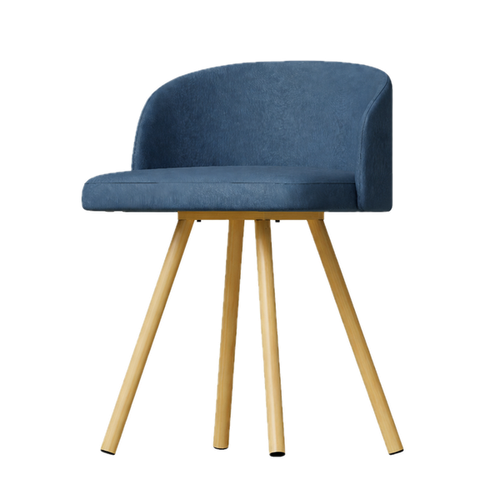} &
    \includegraphics[width=0.19\linewidth, trim={17.86 35.71 17.86 17.86}, clip]{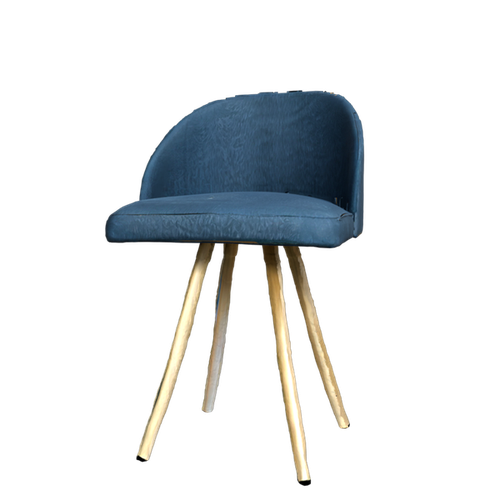} \\
    \raisebox{7pt}{\includegraphics[width=0.16\linewidth, trim={24.41 43.95 24.41 24.41}, clip]{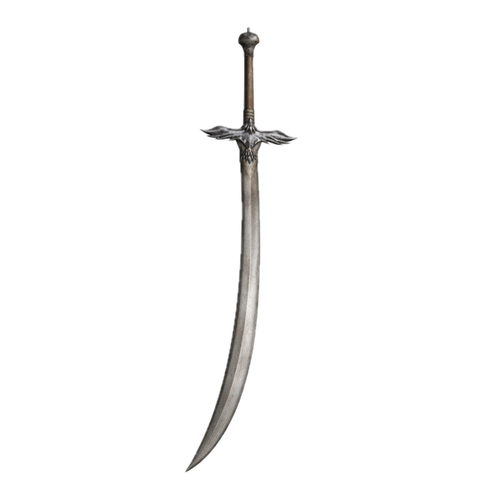}} &
    \includegraphics[width=0.19\linewidth, trim={17.86 17.86 17.86 17.86}, clip]{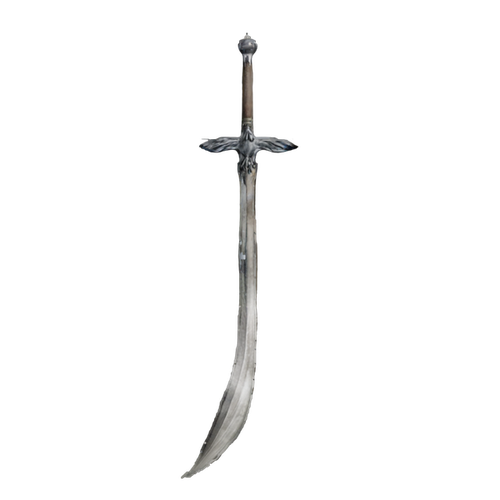} \\
    \raisebox{5pt}{\includegraphics[width=0.15\linewidth, trim={24.41 14.65 24.41 26.86}, clip]{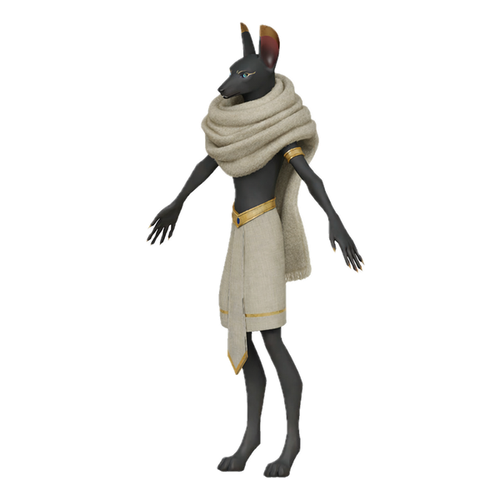}} &
    \includegraphics[width=0.19\linewidth, trim={17.86 17.86 17.86 29.76}, clip]{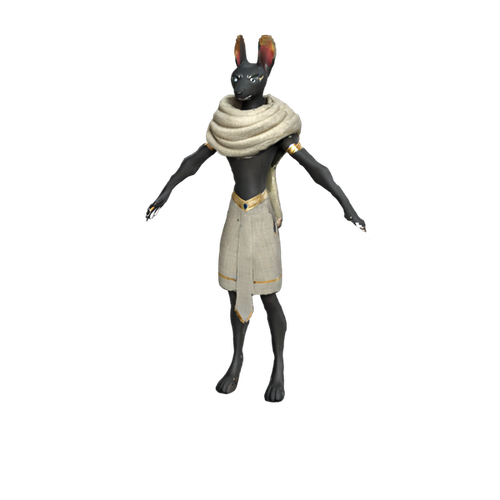} \\
    \hline
  \end{tabular}%
  \vrule width 0.6pt%
  \begin{tabular}[t]{@{}cc@{}}
    \hline
    \scriptsize \shortstack{\textbf{Edit Condition \#2}} &
    \scriptsize \shortstack{\textbf{Output Textured Mesh}} \\
    \hline
    \includegraphics[width=0.16\linewidth, trim={24.41 39.06 24.41 24.41}, clip]{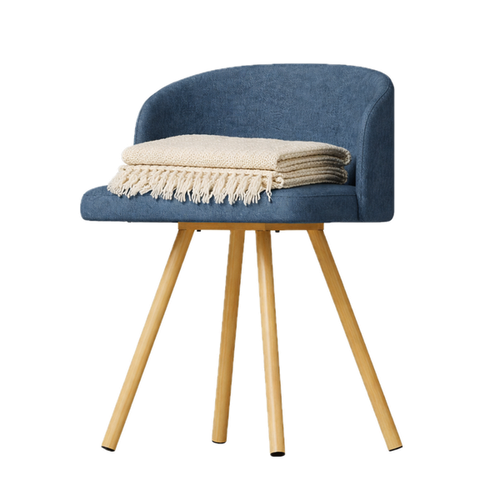} &
    \includegraphics[width=0.19\linewidth, trim={17.86 35.71 17.86 17.86}, clip]{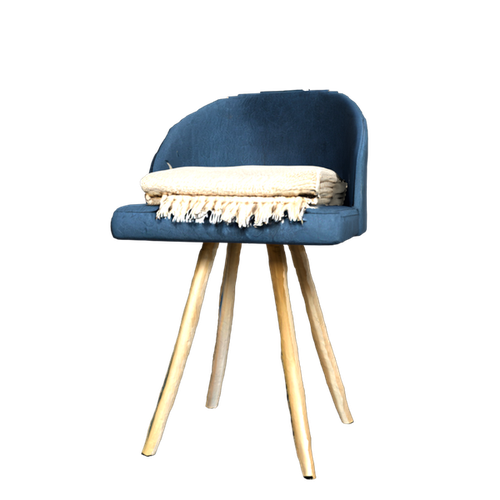} \\
    \raisebox{7pt}{\includegraphics[width=0.16\linewidth, trim={24.41 43.95 24.41 24.41}, clip]{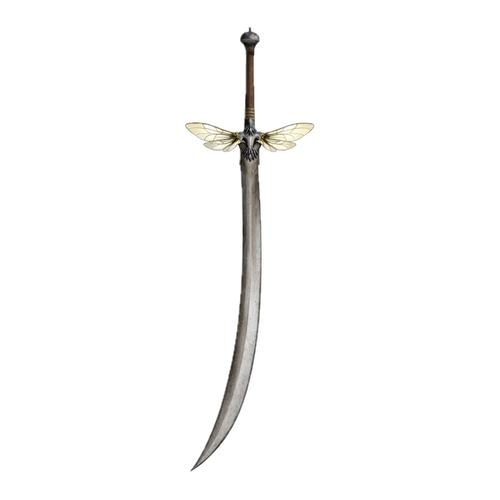}} &
    \includegraphics[width=0.19\linewidth, trim={17.86 17.86 17.86 17.86}, clip]{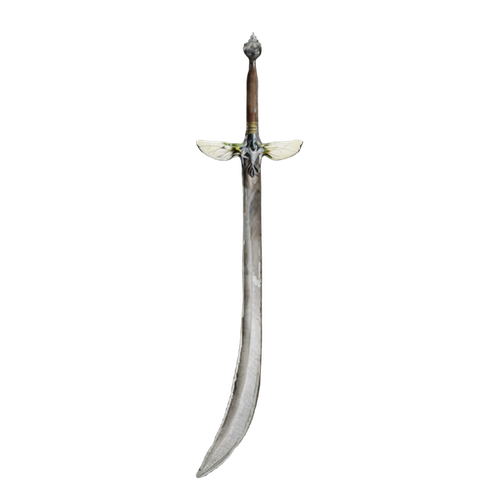} \\
    \raisebox{5pt}{\includegraphics[width=0.15\linewidth, trim={24.41 14.65 24.41 26.86}, clip]{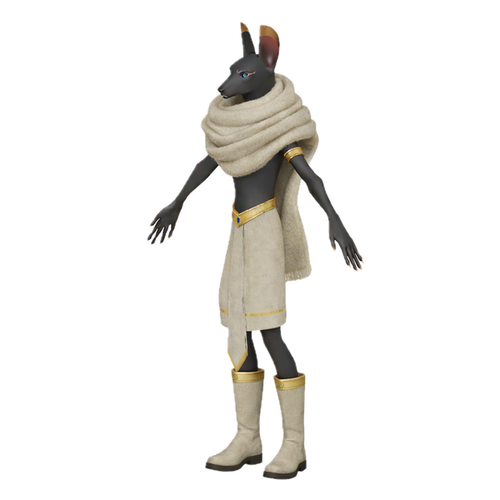}} &
    \includegraphics[width=0.19\linewidth, trim={17.86 17.86 17.86 29.76}, clip]{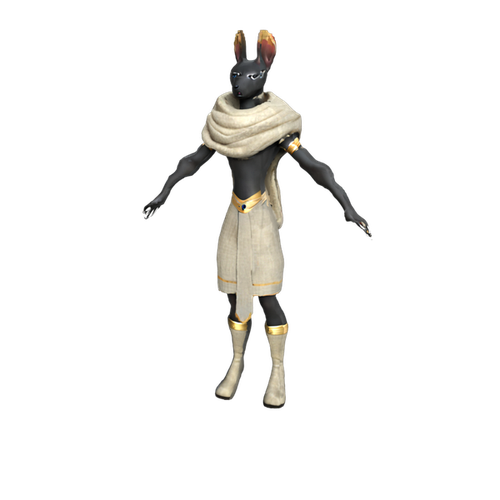} \\
    \hline
  \end{tabular}%
  }%

  \caption{\textbf{Multi-step Editing.} Sequential edits are applied iteratively, where each step takes the previous output as input. For each step, an edit condition image specifies the desired modification, producing a progressively refined result.}
  \label{fig:multistep}
\end{figure*}[t]

\paragraph{Untextured Results}
In addition to the textured qualitative results presented in the main paper, we also include results from the \algoname geometry editing pipeline (prior to texture baking), enabling independent evaluation of its performance. These additional results are shown in \cref{fig:untextured_results}.
    
\subsection{Benchmark Generation Pipeline}

\paragraph{Edit Category Taxonomy}
Our benchmark categorizes edits into four semantically distinct types, each designed to probe different aspects of a model's editing capabilities:

\paragraph{Parts (30\%)}
Adding, removing, or replacing discrete components of an object. This category tests the model's ability to perform localized structural modifications while preserving overall object coherence. Examples include adding accessories to characters (e.g., hats, bionic limbs), modifying vehicle components (e.g., spoilers), or altering furniture elements (e.g., cushions, handles). We specifically instruct the model to focus on medium to large-scale part modifications rather than small accessory changes, ensuring edits are visually significant.

\paragraph{Global -- Deformation (30\%)}
Modifying the overall shape, proportions, or stylistic representation of an object while maintaining its semantic identity. This category evaluates the model's capacity for holistic geometric transformations. Examples include stylization (e.g., converting to a blocky/Minecraft aesthetic), or natural proportion changes (e.g., enlarging a character's body part).

\paragraph{Global -- Pose Change (30\%)}
Altering the articulated pose or configuration of objects with movable parts. This category assesses the model's understanding of object kinematics and plausible configurations. Examples include character poses (e.g., raising arms, kneeling, sitting), mechanical state changes (e.g., opening a laptop, folding car mirrors), and equipment articulation (e.g., tilting a bulldozer blade).

\paragraph{Global -- Texture / Material (10\%)}
Modifying surface appearance, material properties, or color schemes without altering geometry. This category is assigned a lower sampling ratio as it primarily tests appearance transfer rather than geometric editing capabilities. Examples include material substitution (e.g., wood to metal), surface weathering (e.g., rust, wear), and color changes.

\paragraph{Structured Meta-Prompt Design}

To ensure reproducibility and consistency across the benchmark, we employ a structured meta-prompt (\cref{fig:supp_edit_prompt}) that provides Gemini 3 Pro with explicit instructions and curated examples for each edit category.

\newpage
\paragraph{View Sampling Strategy}
For each source mesh, we render 20 views uniformly with azimuth angles in the range $[-30^\circ, 30^\circ]$, corresponding to frontal and near-frontal viewpoints. This focused azimuth range ensures that generated edits depict the most semantically informative object views while maintaining sufficient viewpoint diversity.

\paragraph{Post-Processing}
Generated images undergo automated background removal using BiRefNet~\cite{zheng2024birefnet} to produce clean RGBA outputs. Each edit is stored with metadata recording the source view, sampled edit category, and generation parameters, enabling full reproducibility.
\ifarxiv
\else

\fi

\begin{figure*}[t]
\centering
\begin{lstlisting}[style=promptstyle, basicstyle=\ttfamily\fontsize{7.5pt}{9pt}\selectfont]
PROMPT = """Please edit the given image while keeping everything other than my edit request exactly the same.
This includes the viewing angle, camera perspective, lighting conditions, background, overall composition, and the core identity of the object.

The edit types are: "parts", "global - deformation", "global - pose change", and "global - texture / material".
Below are descriptions and examples for each type.
These examples are meant to help you understand the kinds of edits expected.
Please do not limit yourself to these specific examples; instead, create novel edits in the same spirit that are appropriate for the object in the image.
Edits should be simple, clear, and should not be so drastic as to lose the object's identity.

Parts - Add, remove, or replace a part of the object.
Examples:

Add a hat to a character.
Change an arm of a character to be a bionic arm.
Add a spoiler to a car.
Make the character hold something
Remove the handle from a mug.
Replace the wheels on a skateboard with different objects.
Remove a chimney from a house.
Add a cushion to a chair.

For this type of edit - please avoid small item changes (like adding a flashlight) and focus on medium to medium-large item additions / subtraction / replacements (like an entire arm, holding a large object, replacing a head).
Also - do not avoid mixing in different colors for the new parts.

Global - Deformation - Modify the overall shape or proportions of the object while maintaining its identity.
Examples:

Change the style to boxy / minecraft.
Change the style to be plush / doll.
Make the character or an object taller or shorter (naturally, not just stretched).
Modify a character's body proportions (naturally, not just stretched).
Change the style to toy or low-poly.
Enlarge the head of an animal.

Global - Pose Change - Change the pose of the object while maintaining its identity.
Examples:

Make a character raise their arms.
Make an animal sit or lie down.
Open a laptop that was closed.
Fold down the mirrors on a car.
Make a character kneel.
Tilt the blade of a bulldozer.

Global - Texture / Material - Change the surface appearance, material, color, or style of the object without altering its shape.
Examples:

Change a wooden table to a metal one.
Change the surface to rough stone or concrete.
Make the object look rusty and weathered.
Change the color of a sofa from brown to blue.

If the type of edit does is not relevant for the kind of object you are given, for example - pose change for a vehicle - you can choose a different edit type.

I want you to be creative, I am happy with very significant edits, I just want the unedited areas and the overall identity to remain generally the same.
I want the edits to pop but still be very clear.

Remember: aside from the requested edit, everything else must remain identical.
The resulting image should be geometrically plausible and visually coherent with the original, but it can obviously look like a significant edit was made / or a different version of the original.
Please use a solid background color that allows for easy background removal.
For the current image, please make the following edit: {edit}"""
\end{lstlisting}
\caption{The full meta-prompt used for the editing pipeline.}
\label{fig:supp_edit_prompt}
\end{figure*}

\clearpage

\else
    \bibliographystyle{ACM-Reference-Format}
    \bibliography{main}

\fi

\end{document}